\documentclass[conference]{IEEEtran}

\usepackage{cite}
\usepackage{amsmath,amssymb,amsfonts}
\usepackage{algorithmic}
\usepackage{graphicx}
\usepackage{textcomp}
\usepackage{booktabs} 
\usepackage{caption}
\usepackage{subcaption}
\usepackage[utf8]{inputenc}
\usepackage{threeparttable}
\usepackage{longtable}
\usepackage{url}
\usepackage{multirow}
\usepackage{xcolor}
\usepackage{bm}

\def\BibTeX{{\rm B\kern-.05em{\sc i\kern-.025em b}\kern-.08em
    T\kern-.1667em\lower.7ex\hbox{E}\kern-.125emX}}
\begin{document}

\title{Maximizing Relation Extraction Potential: A Data-Centric Study to Unveil Challenges and Opportunities
}

\author{\IEEEauthorblockN{Anushka Swarup\IEEEauthorrefmark{1},
Avanti Bhandarkar, Olivia P. Dizon-Paradis,
Ronald Wilson, and Damon L. Woodard}
\IEEEauthorblockA{Florida Institute for National Security (FINS),
University of Florida\\
Gainesville FL 32611, USA.\\
\IEEEauthorrefmark{1}Corresponding author\\
}}


\maketitle

\begin{abstract}
Relation extraction is a Natural Language Processing task that aims to extract relationships from textual data. It is a critical step for information extraction. Due to its wide-scale applicability, research in relation extraction has rapidly scaled to using highly advanced neural networks. Despite their computational superiority, modern relation extractors fail to handle complicated extraction scenarios. However, a comprehensive performance analysis of the state-of-the-art extractors that compile these challenges has been missing from the literature, and this paper aims to bridge this gap. The goal has been to investigate the possible data-centric characteristics that impede neural relation extraction. Based on extensive experiments conducted using 15 state-of-the-art relation extraction algorithms ranging from recurrent architectures to large language models and seven large-scale datasets, this research suggests that modern relation extractors are not robust to complex data and relation characteristics. It emphasizes pivotal issues, such as contextual ambiguity, correlating relations, long-tail data, and fine-grained relation distributions. In addition, it sets a marker for future directions to alleviate these issues, thereby proving to be a critical resource for novice and advanced researchers. Efficient handling of the challenges described can have significant implications for the field of information extraction, which is a critical part of popular systems such as search engines and chatbots. Data and relevant code can be found at \url{https://aaig.ece.ufl.edu/projects/relation-extraction}.
\end{abstract}

\begin{IEEEkeywords}
Information Extraction, Joint Relation Extraction, Large Language Models, Natural Language Processing, Relation Classification, Relation Extraction
\end{IEEEkeywords}

\newcommand{\overlap}{
\begin{figure}[h]
  \centering
  \includegraphics[width=\columnwidth]{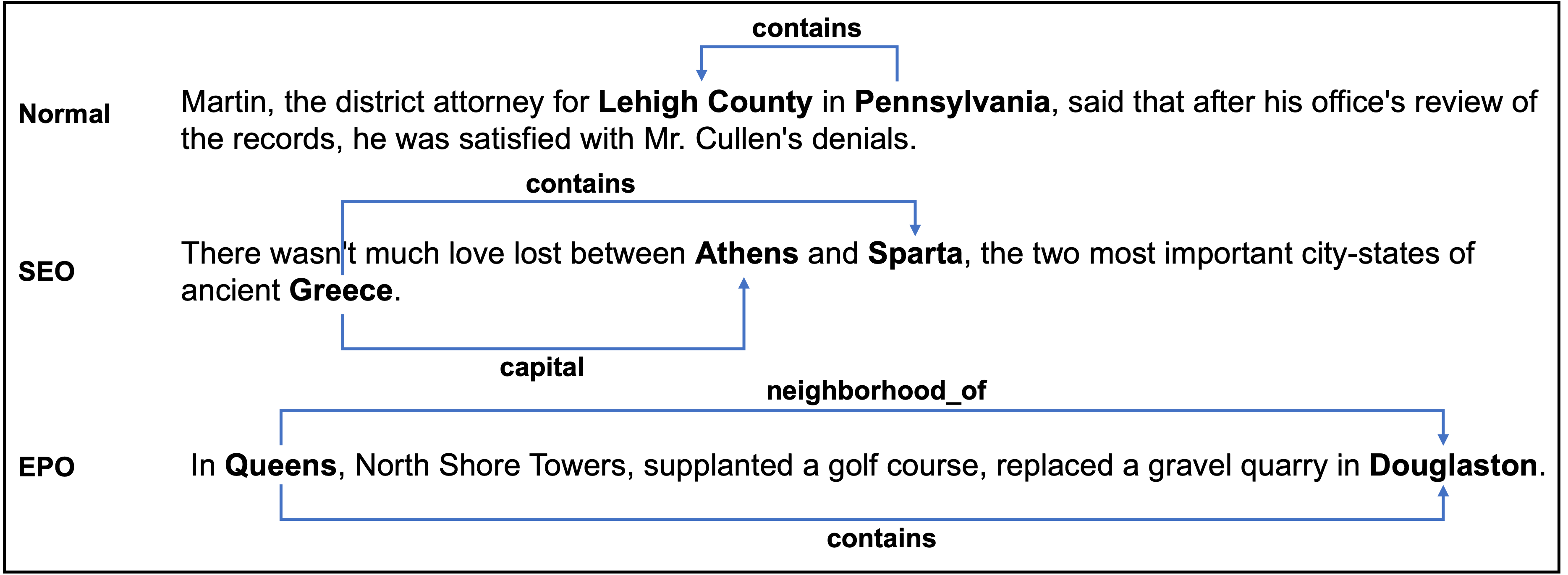}
  \caption{Different categorizations of the multiple relations and overlapping entity problem}
  \label{fig:overlap}
\end{figure}
}

\newcommand{\pipe}{
\begin{figure}[h]
  \centering
  \includegraphics[width=\columnwidth]{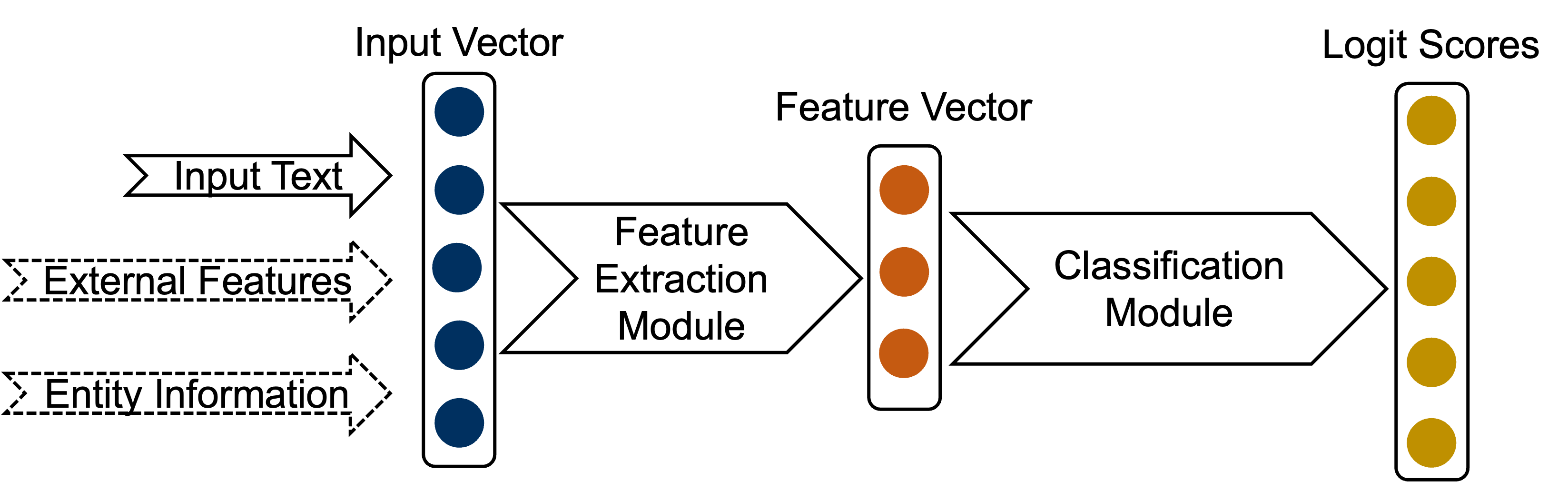}
  \caption{A basic deep learning pipeline for relation extraction}
  \label{fig:pipe}
\end{figure}
}

\newcommand{\taxo}{
\begin{figure*}[h]
  \centering
\includegraphics[width=0.6\textwidth]{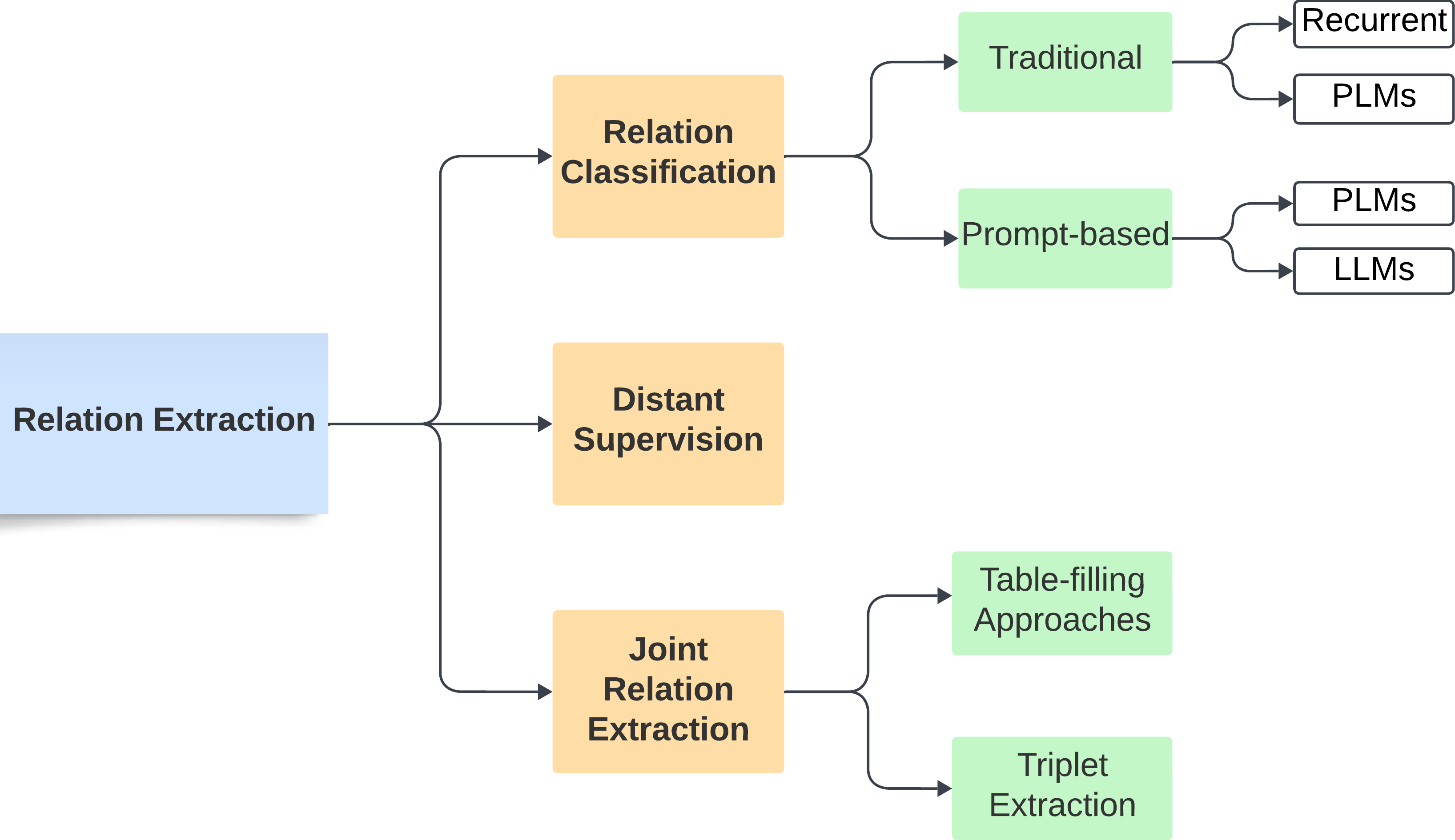}
  \caption{Taxonomy of relation extraction algorithms}
  \label{fig:taxo}
\end{figure*}
}

\newcommand{\longtail}{
\begin{figure}[h]
  \centering
  \includegraphics[width=0.75\columnwidth]{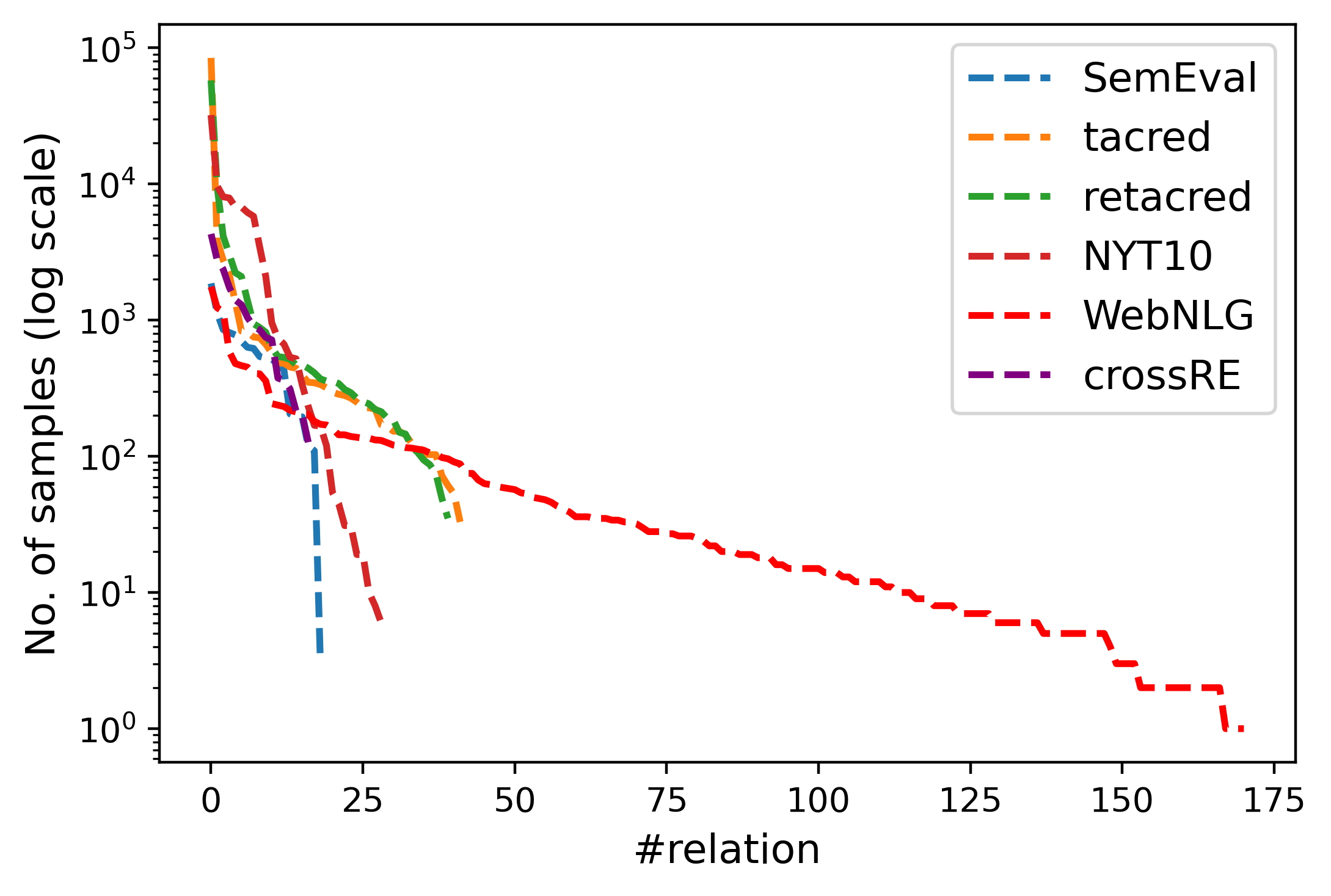}
  \caption{Distribution of samples per relation category.}
  \label{fig:longtail}
\end{figure}
}

\newcommand{\superAll}{
\begin{figure}[h]
  \centering
  \includegraphics[width=\columnwidth]{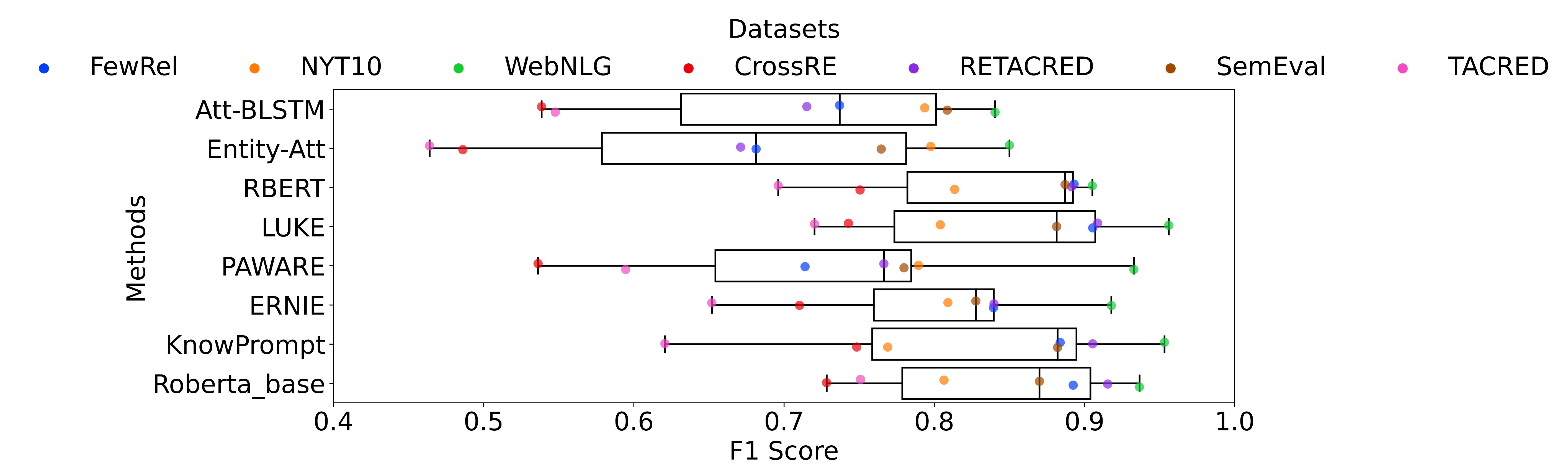}
  \caption{Average micro-F1 scores for all datasets and relation classifier combinations in the fully supervised setting. The x-axis has been magnified according to the range of scores achieved.}
  \label{fig:sup_all}
\end{figure}
}

\newcommand{\jointAll}{
\begin{figure}[h]
  \centering
  \includegraphics[width=\columnwidth]{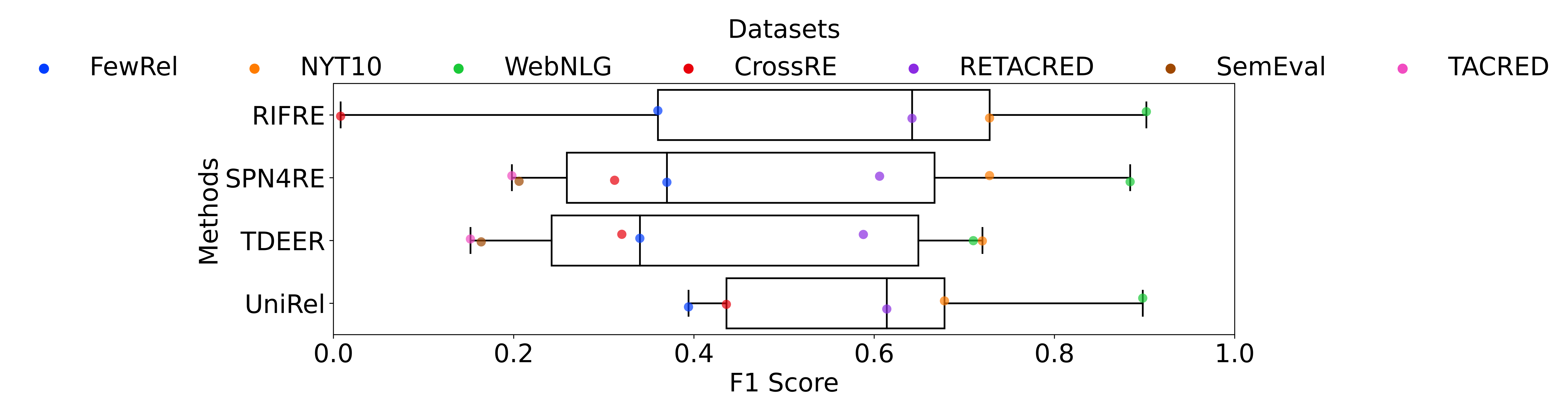}
  \caption{Average micro-F1 scores for all datasets and joint relation extractor combinations in the fully supervised setting.}
  \label{fig:joint_all}
\end{figure}
}

\newcommand{\fewAll}{
\begin{figure*}[h]
  \centering
  \includegraphics[width=\linewidth]{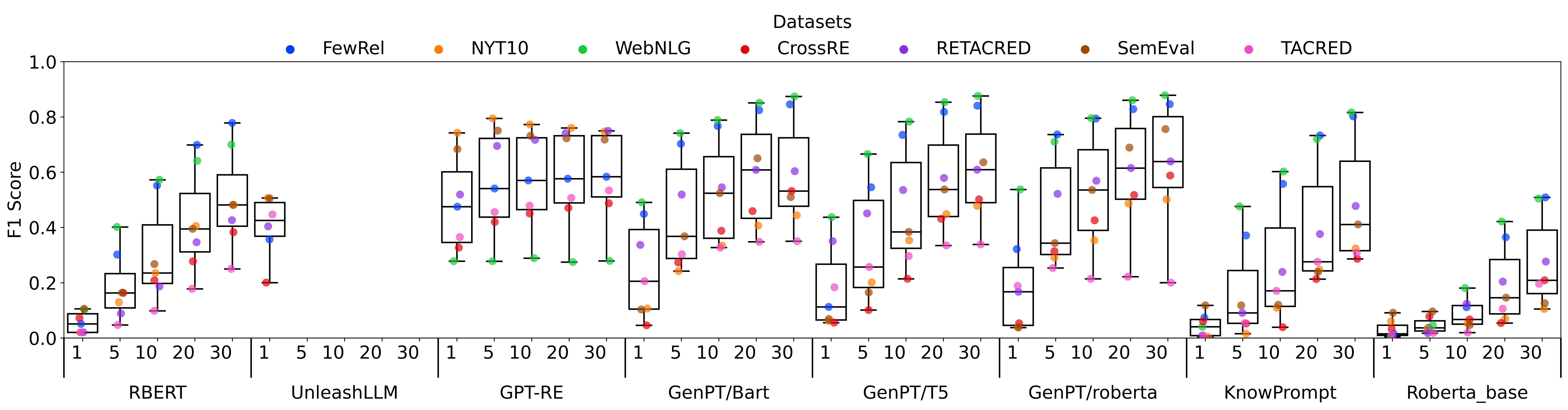}
  \caption{Average micro-F1 scores for all datasets and relation classifier combinations in the few-shot setting. Note that the UnleashLLM algorithm was only compatible with the k=1 setup.}
  \label{fig:few_all}
\end{figure*}
}

\newcommand{\superComplex}{
\begin{figure}[h]
     \centering
     \begin{subfigure}[b]{\columnwidth}
         \centering
         \includegraphics[width=\columnwidth]{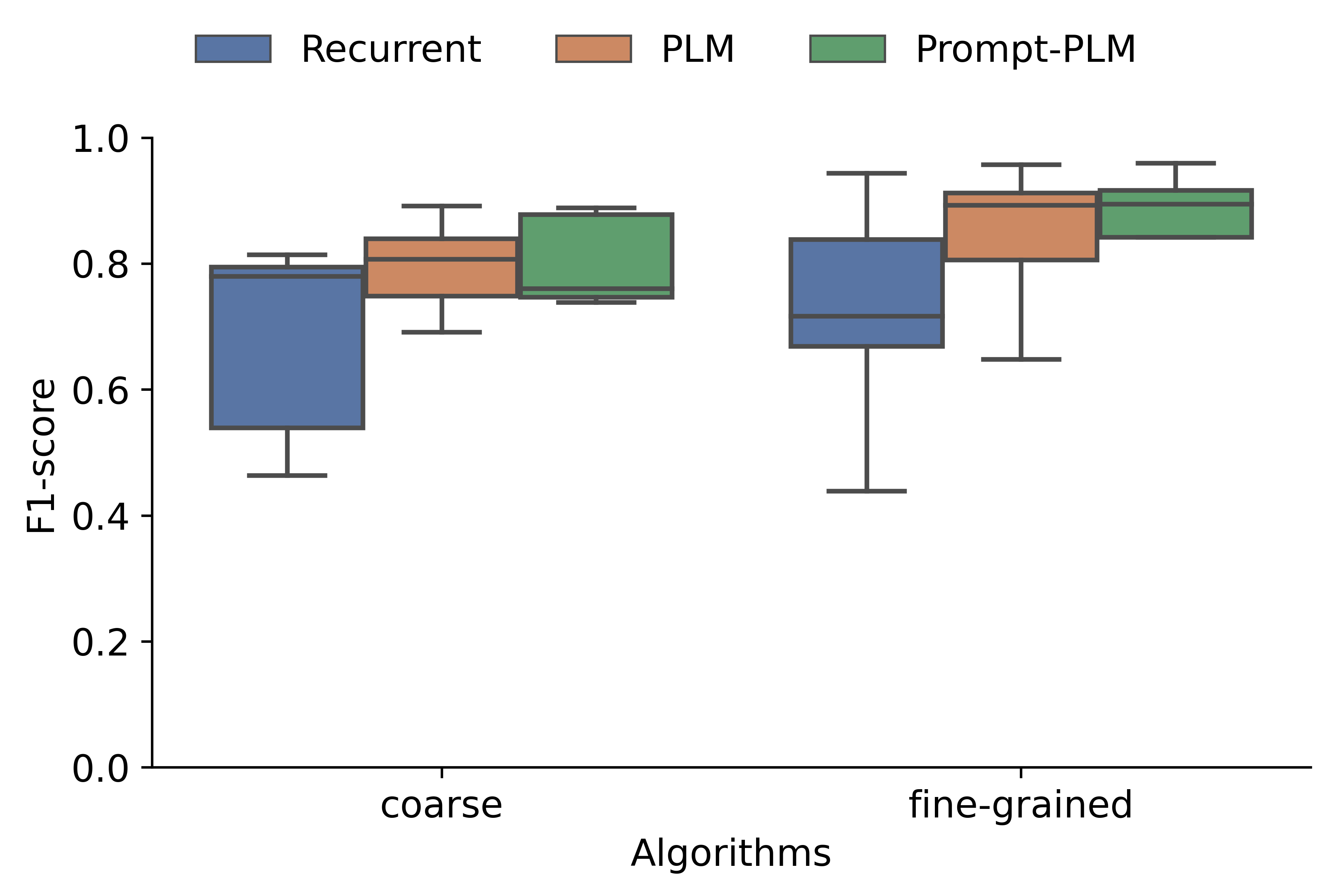}
         \caption{Fine-grained vs. Coarse}
         \label{subfig:sup_fine}
     \end{subfigure}
     \hfill
     \begin{subfigure}[b]{\columnwidth}
         \centering
         \includegraphics[width=\columnwidth]{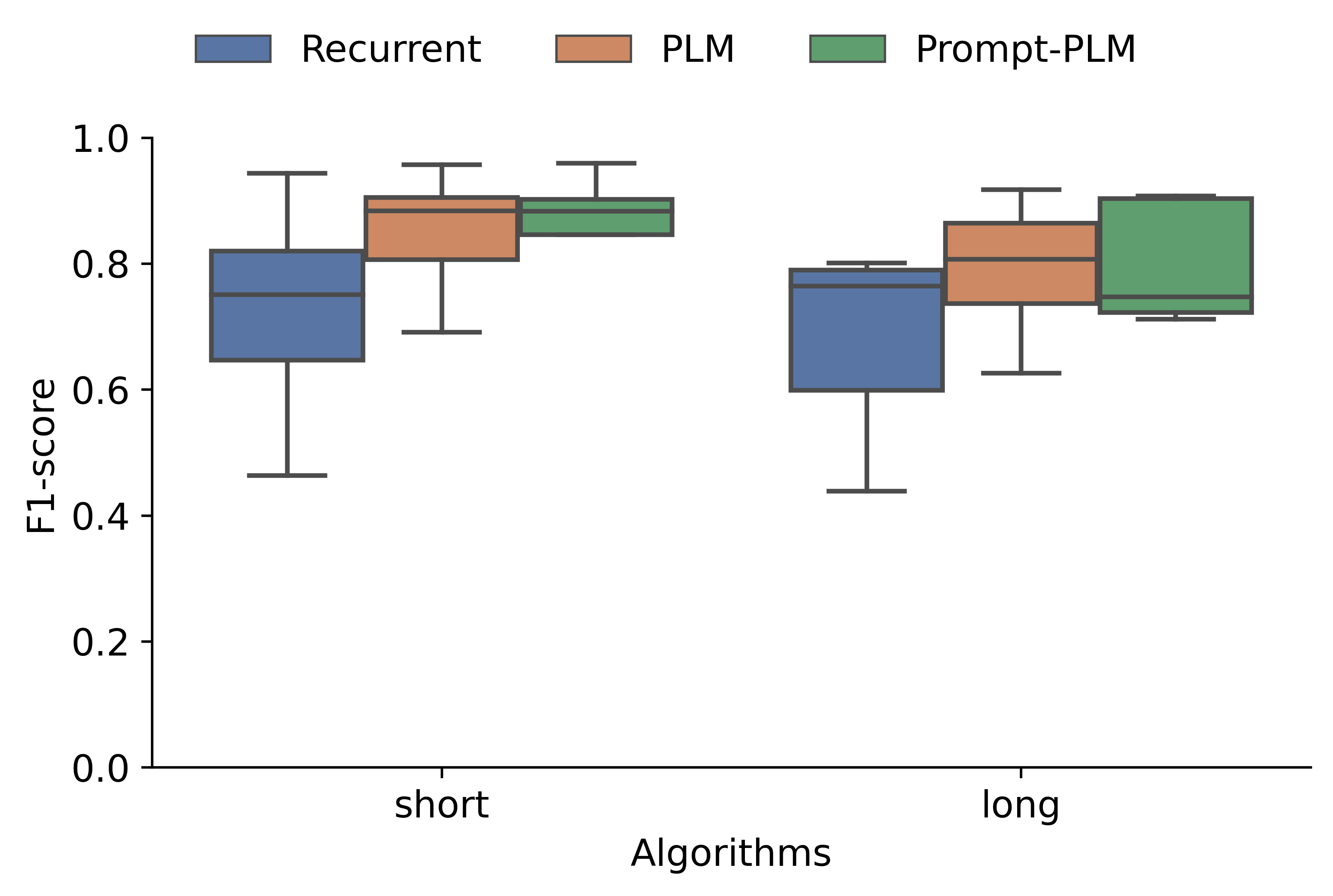}
         \caption{Long vs. Short sampled}
         \label{subfig:sup_len}
     \end{subfigure}
     \caption{Micro F1-score distribution for supervised relation classification}
     \label{fig:superComplex}
\end{figure}
}

\newcommand{\fewComplex}{
\begin{figure}[h]
     \centering
     \begin{subfigure}[b]{\columnwidth}
         \centering
         \includegraphics[width=\columnwidth]{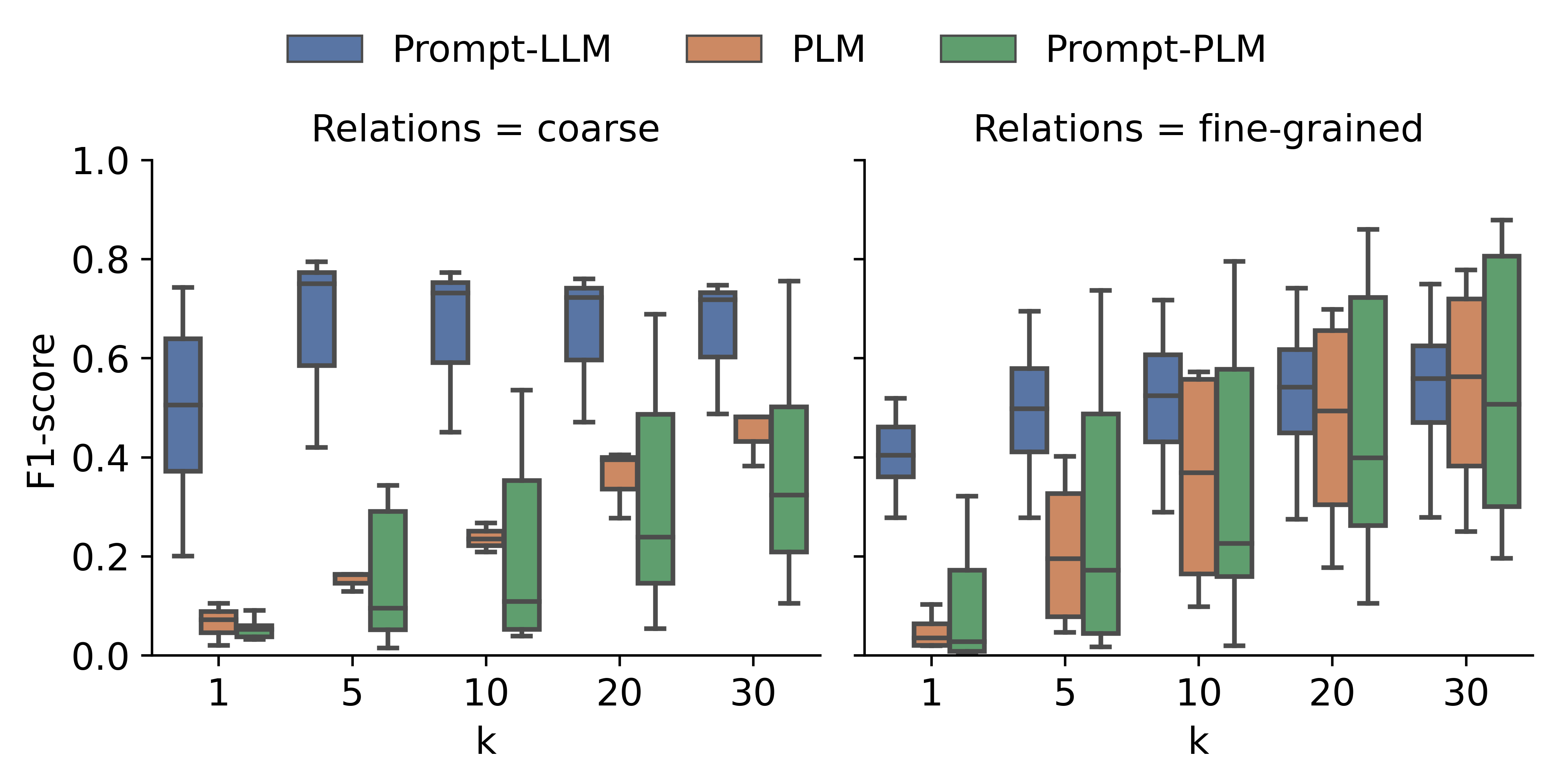}
         \caption{Fine-grained vs. Coarse relation distribution}
         \label{subfig:few_fine}
     \end{subfigure}
     \hfill
     \begin{subfigure}[b]{\columnwidth}
         \centering
         \includegraphics[width=\columnwidth]{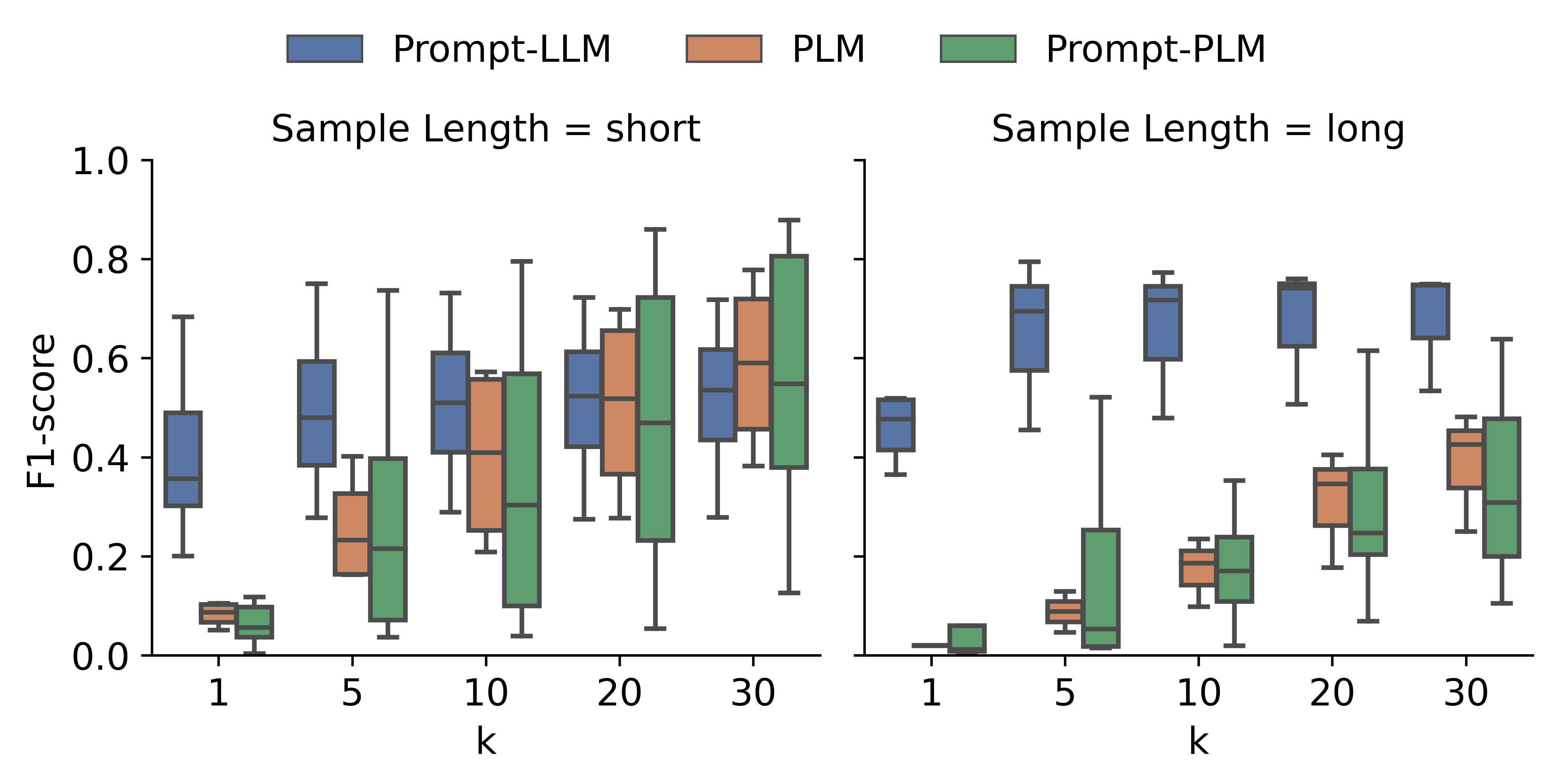}
         \caption{Long vs. Short sampled}
         \label{subfig:few_len}
     \end{subfigure}
     \caption{Micro F1-score distribution for few-shot relation classification}
     \label{fig:fewComplex}
\end{figure}
}

\newcommand{\superOverlap}{
\begin{figure}[h]
     \centering
     \begin{subfigure}[b]{\columnwidth}
         \centering
         \includegraphics[width=\textwidth]{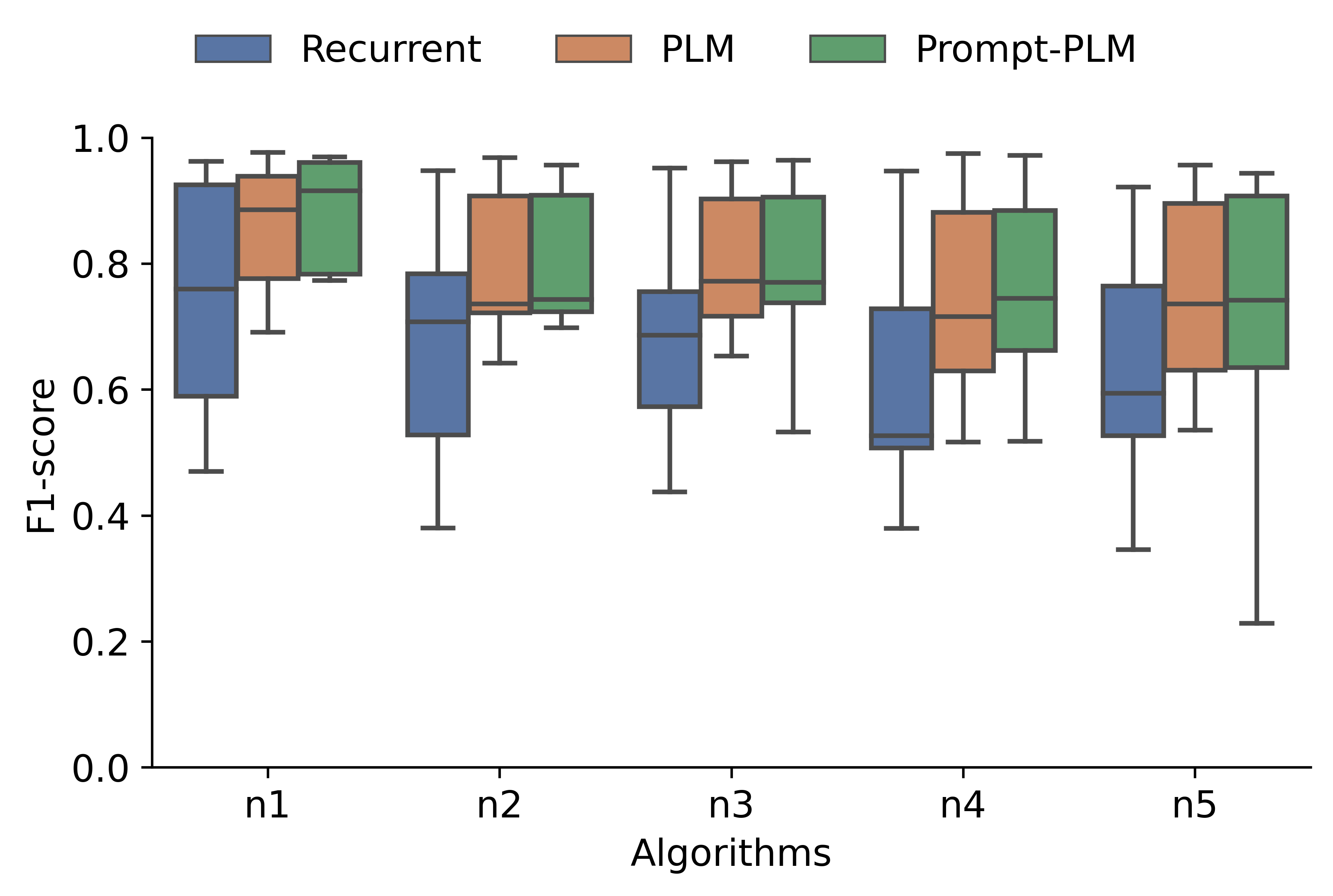}
         \caption{Multiple Relations}
         \label{subfig:sup_multi}
     \end{subfigure}
     \hfill
     \begin{subfigure}[b]{\columnwidth}
         \centering
         \includegraphics[width=\columnwidth]{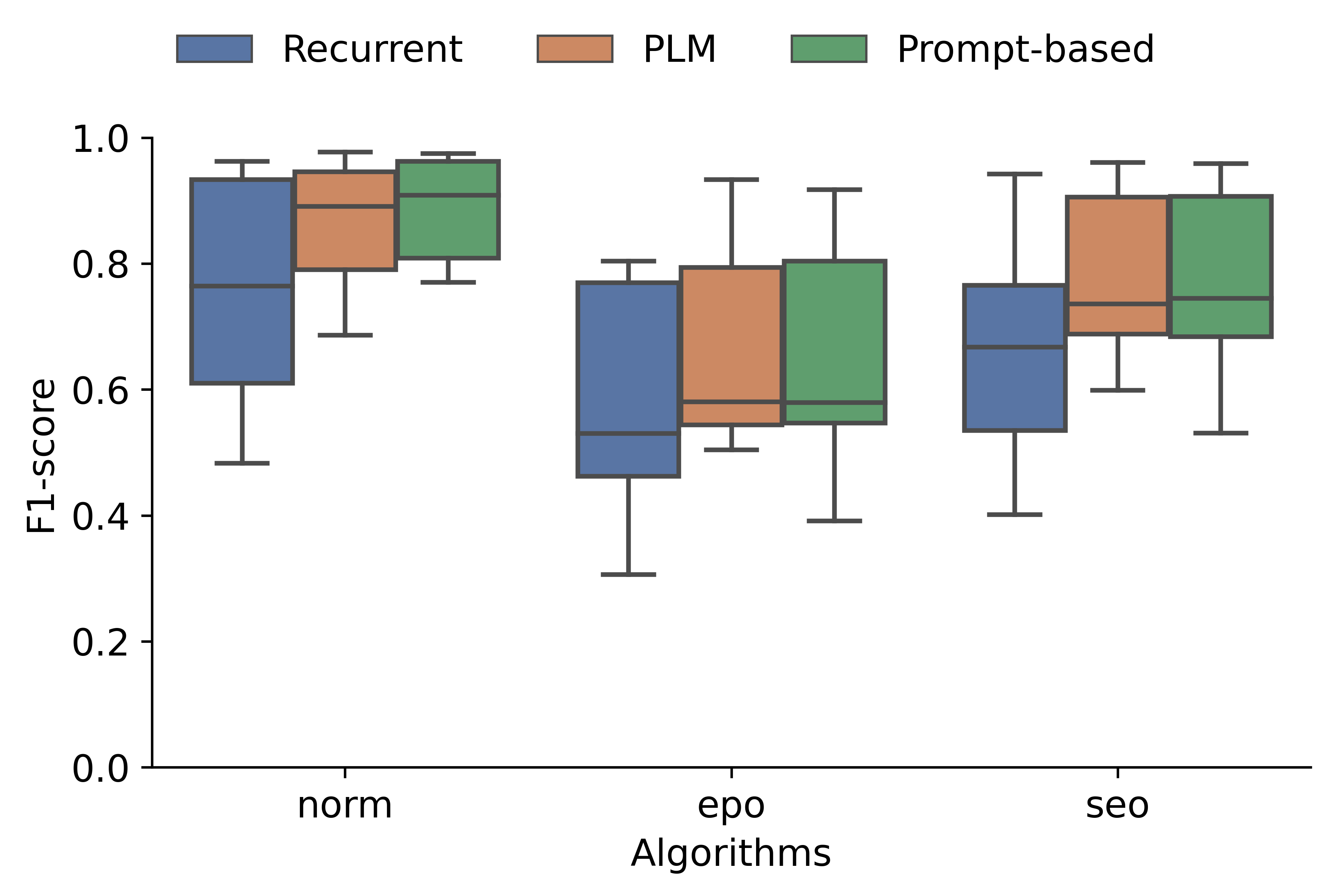}
         \caption{Overlapping Entities}
         \label{subfig:sup_over}
     \end{subfigure}
     \caption{Micro F1-score distribution for supervised relation classification}
     \label{fig:superOverlap}
\end{figure}
}

\newcommand{\fewOverlap}{
\begin{figure*}[h]
     \centering
     \begin{subfigure}[b]{\textwidth}
         \centering
         \includegraphics[width=\textwidth]{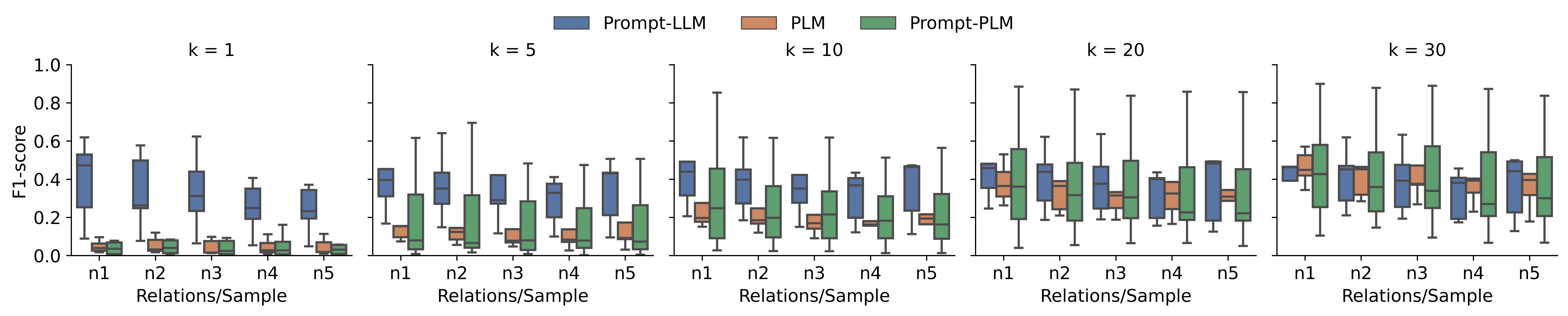}
         \caption{Multiple Relations}
         \label{subfig:few_multi}
     \end{subfigure}
     \hfill
     \begin{subfigure}[b]{\textwidth}
         \centering
         \includegraphics[width=\textwidth]{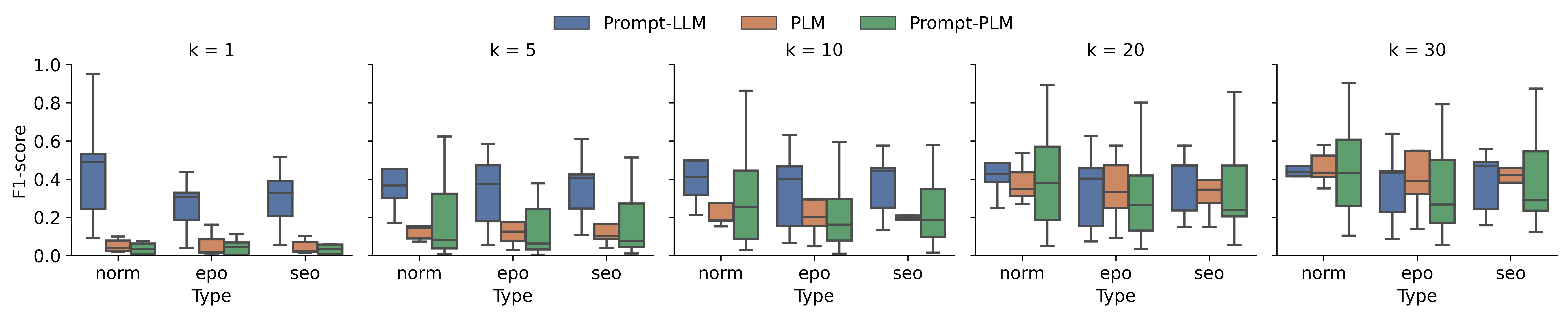}
         \caption{Overlapping Entities}
         \label{subfig:few_over}
     \end{subfigure}
     \caption{Micro F1-score distribution for few-shot relation classification}
     \label{fig:fewOverlap}
\end{figure*}
}

\newcommand{\jreComplex}{
\begin{figure}[h]
     \centering
     \begin{subfigure}[b]{\columnwidth}
         \centering
         \includegraphics[width=\columnwidth]{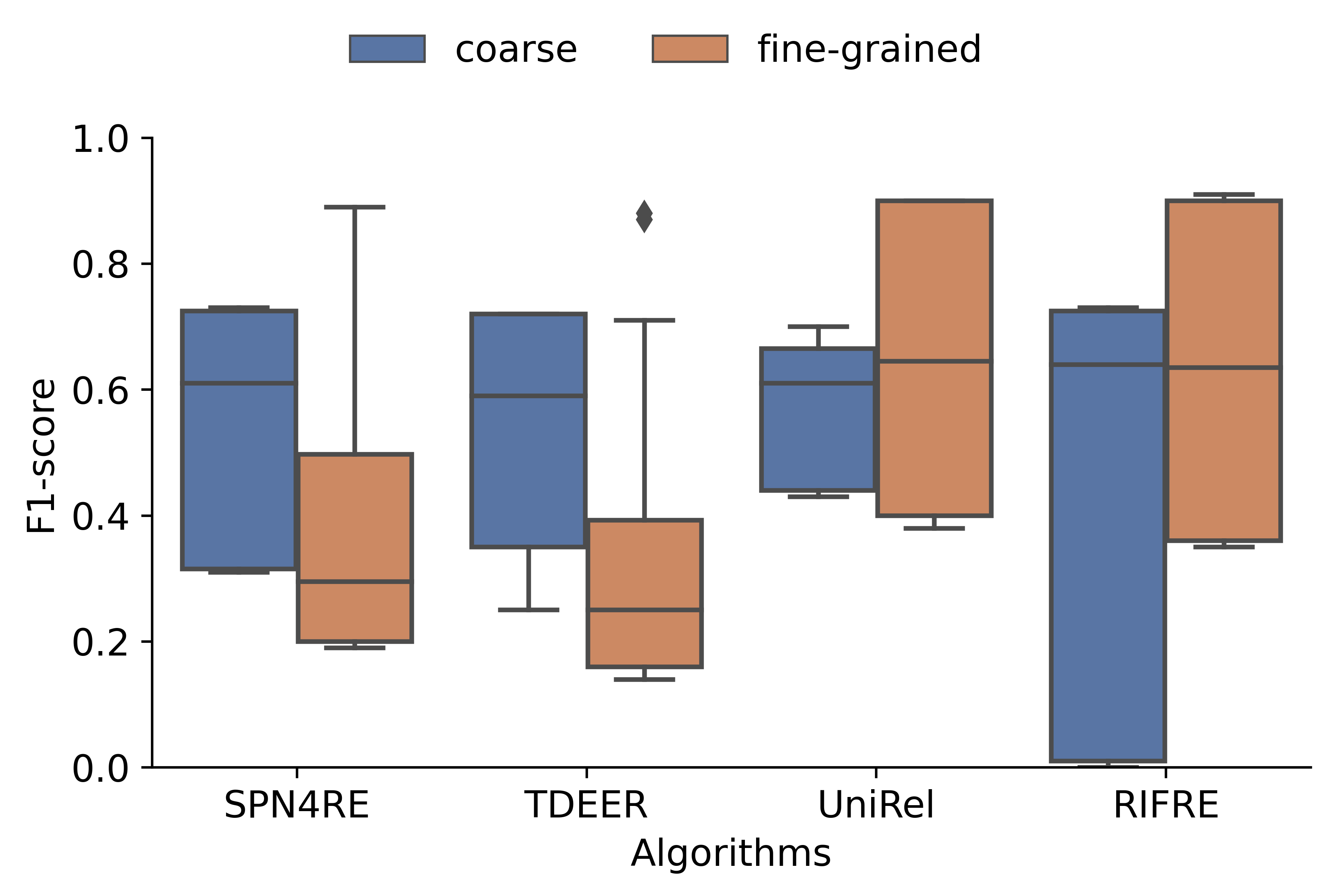}
         \caption{Fine-grained vs. Coarse relation distribution}
         \label{subfig:joint_fin}
     \end{subfigure}
     \hfill
     \begin{subfigure}[b]{\columnwidth}
         \centering
         \includegraphics[width=\columnwidth]{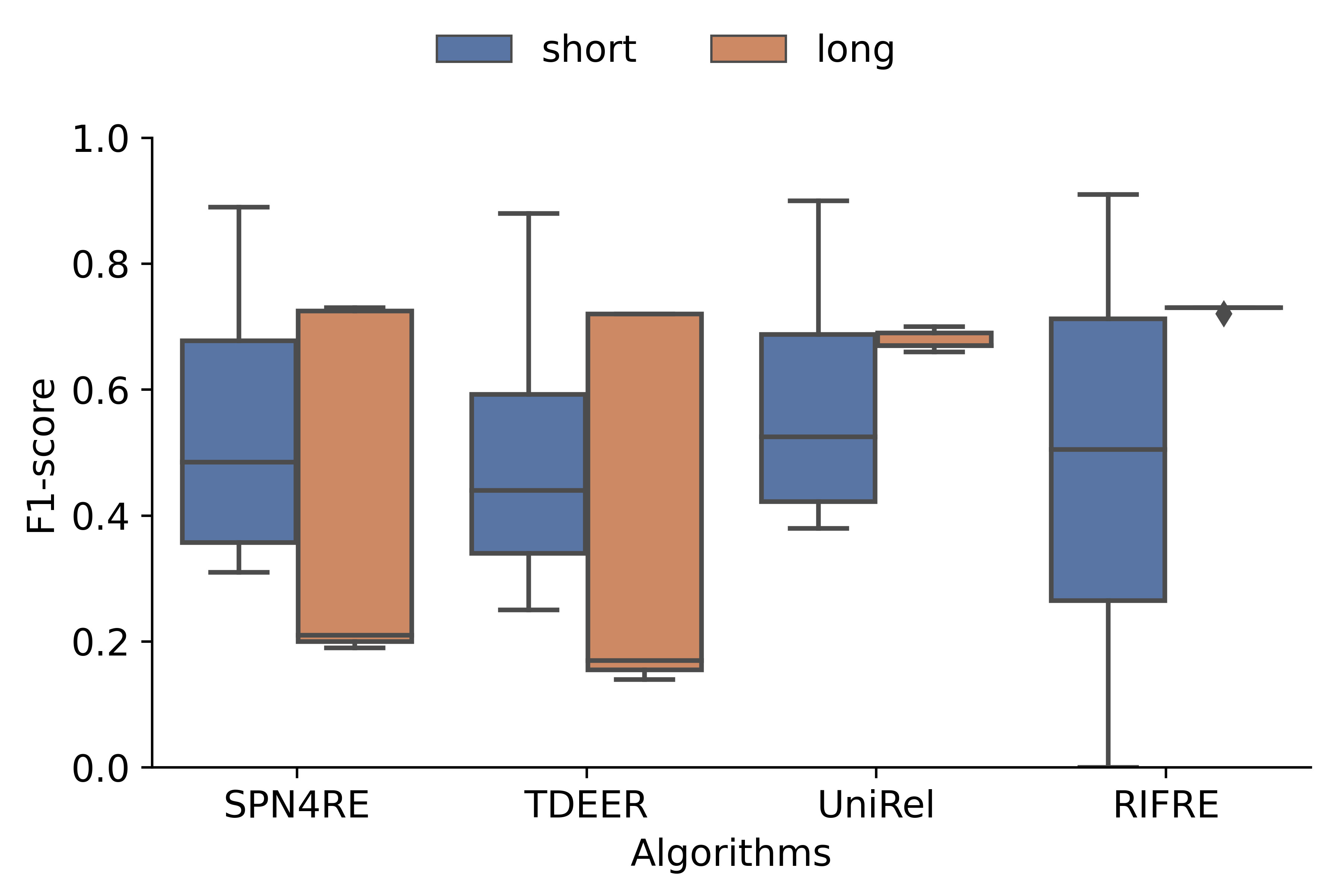}
         \caption{Long vs. Short sampled}
         \label{subfig:joint_len}
     \end{subfigure}
     \caption{Micro F1-score distribution for joint relation extractors}
     \label{fig:jreComplex}
\end{figure}
}

\newcommand{\jreMulti}{
\begin{figure}[h]
     \centering
     \begin{subfigure}[b]{0.45\textwidth}
         \centering
         \includegraphics[width=\textwidth]{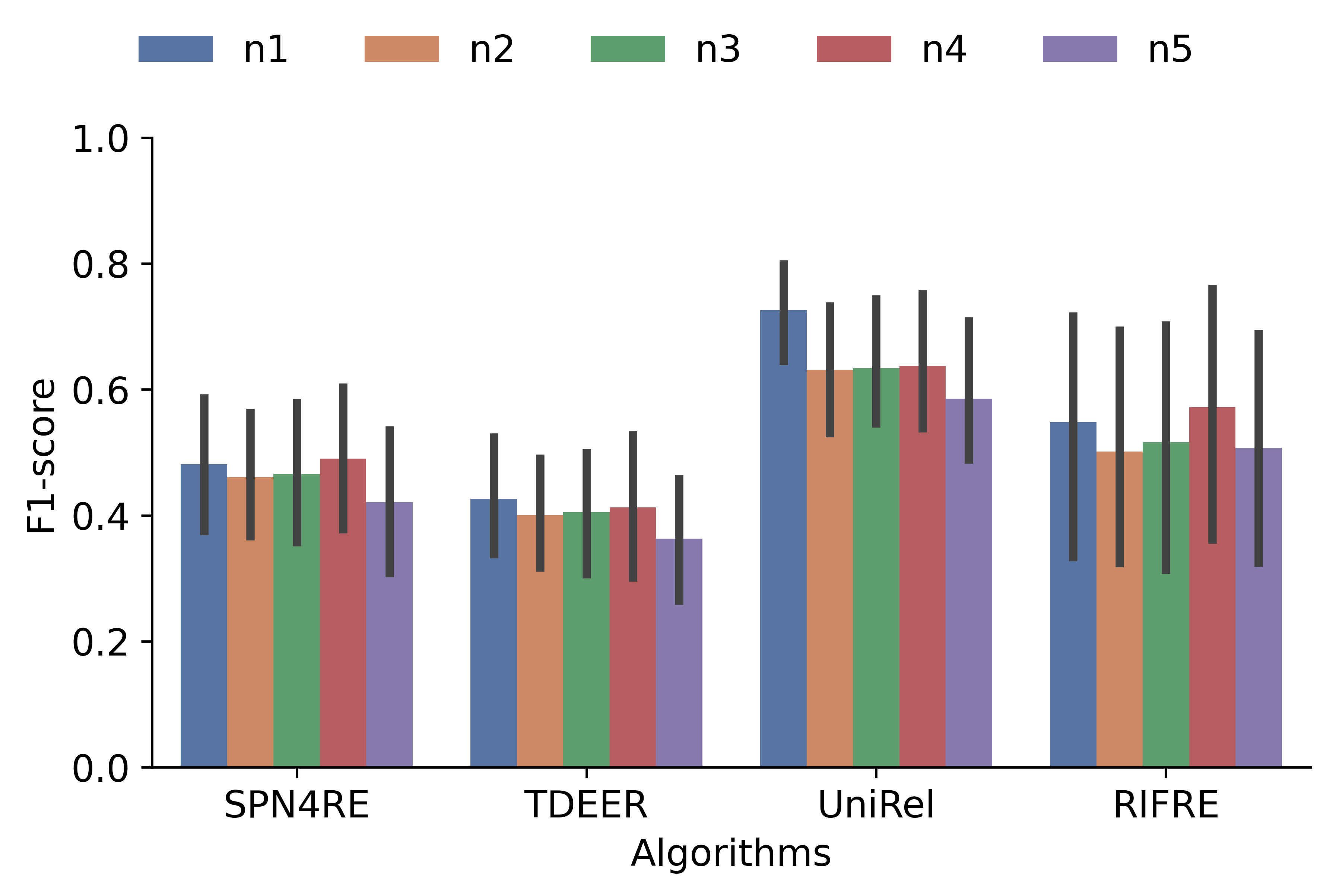}
         \caption{Multiple Relations}
         \label{subfig:jointMulti}
     \end{subfigure}
     \hfill
     \begin{subfigure}[b]{0.45\textwidth}
         \centering
         \includegraphics[width=\textwidth]{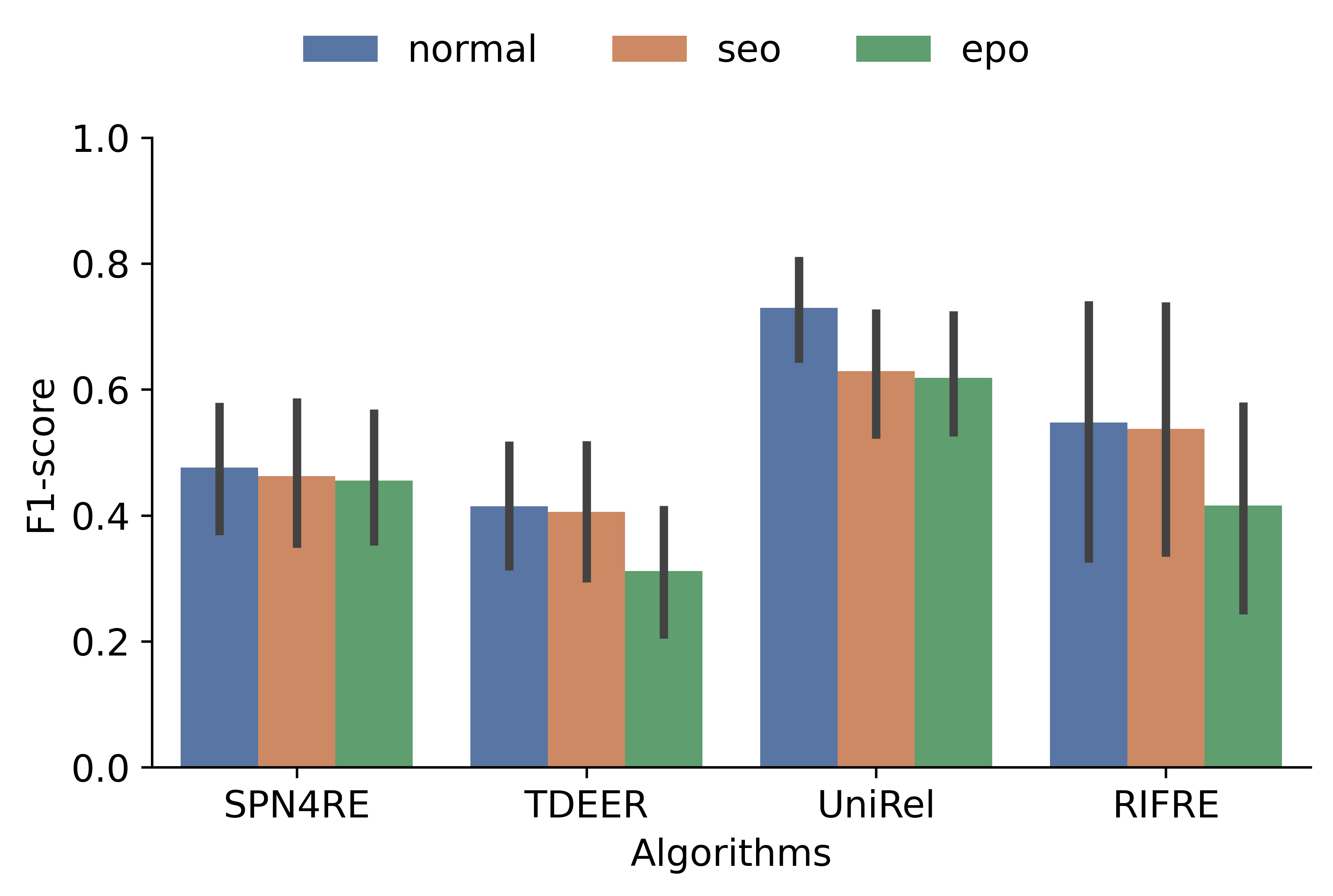}
         \caption{Overlapping Entities}
         \label{subfig:jointOvelap}
     \end{subfigure}
     \caption{Micro F1-score distribution for joint relation extractors}
     \label{fig:jreMultiOver}
\end{figure}
}

\newcommand{\contextsim}{
\begin{figure*}[h]
     \centering
     \begin{subfigure}[b]{0.3\textwidth}
         \centering
         \includegraphics[width=\textwidth]{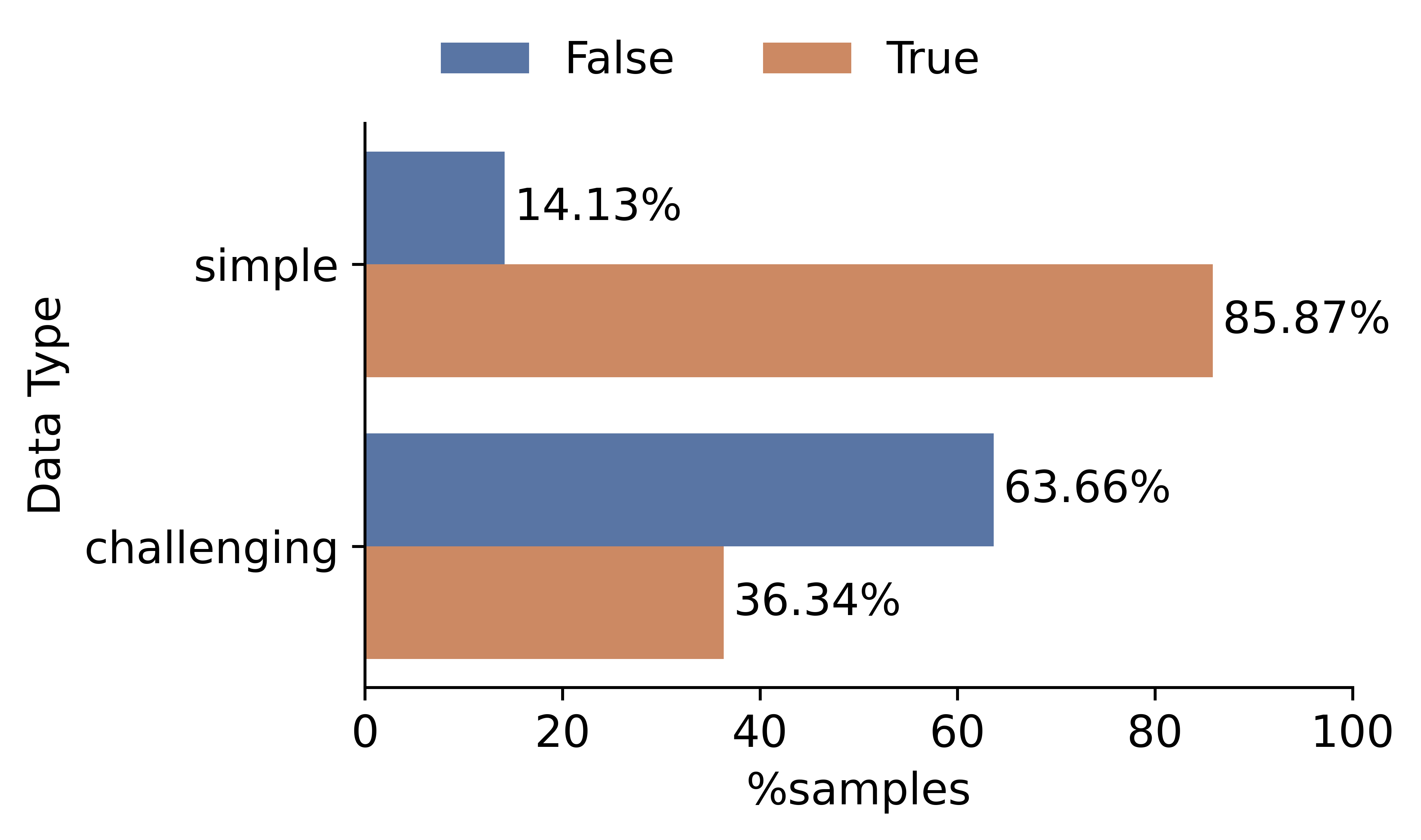}
         \caption{Recurrent Algorithms}
         \label{fig:sim_measure_rec}
     \end{subfigure}
     \hfill
     \begin{subfigure}[b]{0.3\textwidth}
         \centering
         \includegraphics[width=\textwidth]{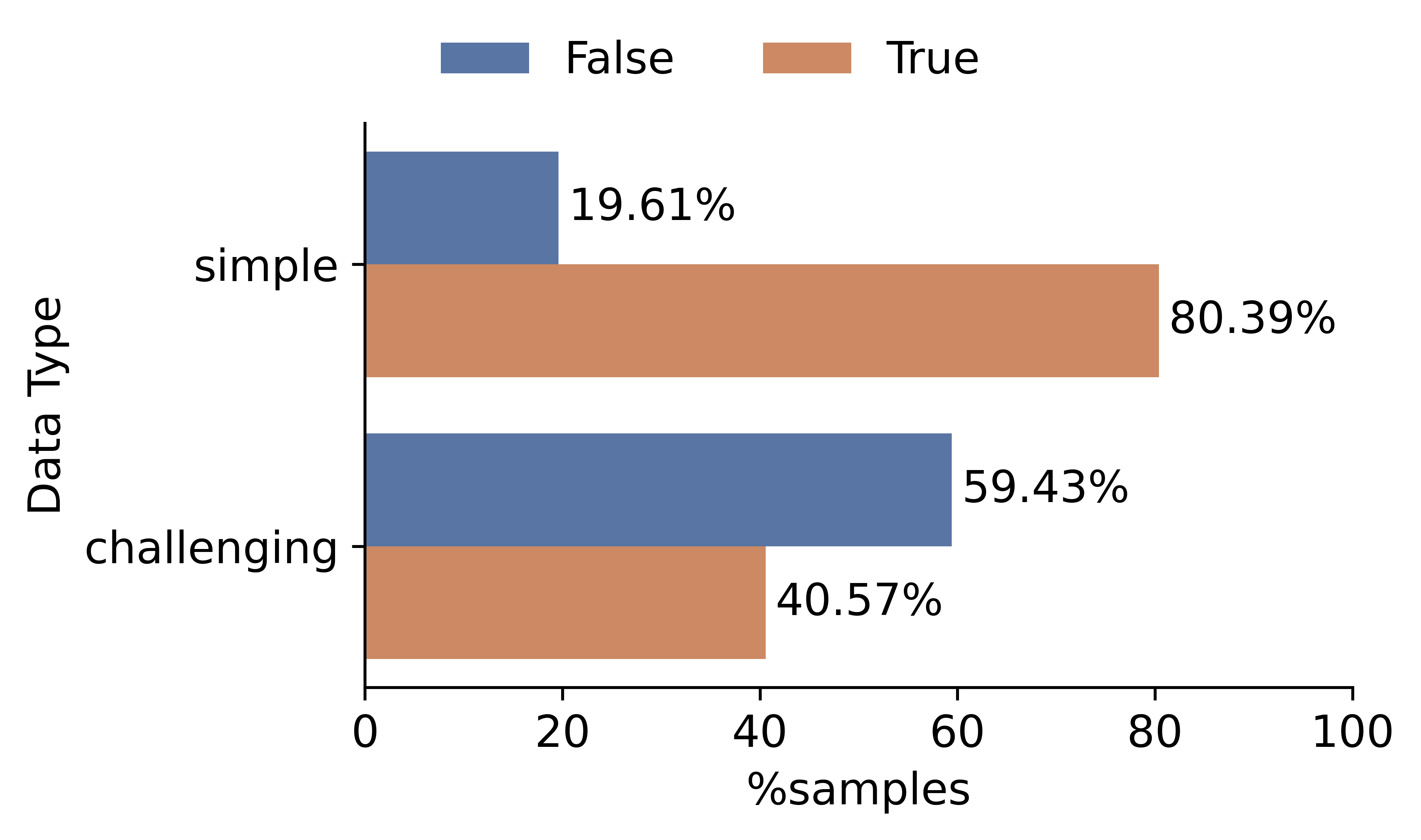}
         \caption{PLM-based Algorithms}
         \label{fig:sim_measure_PLM}
     \end{subfigure}
     \label{fig:jreMulti}
     \hfill
     \begin{subfigure}[b]{0.3\textwidth}
         \centering
         \includegraphics[width=\textwidth]{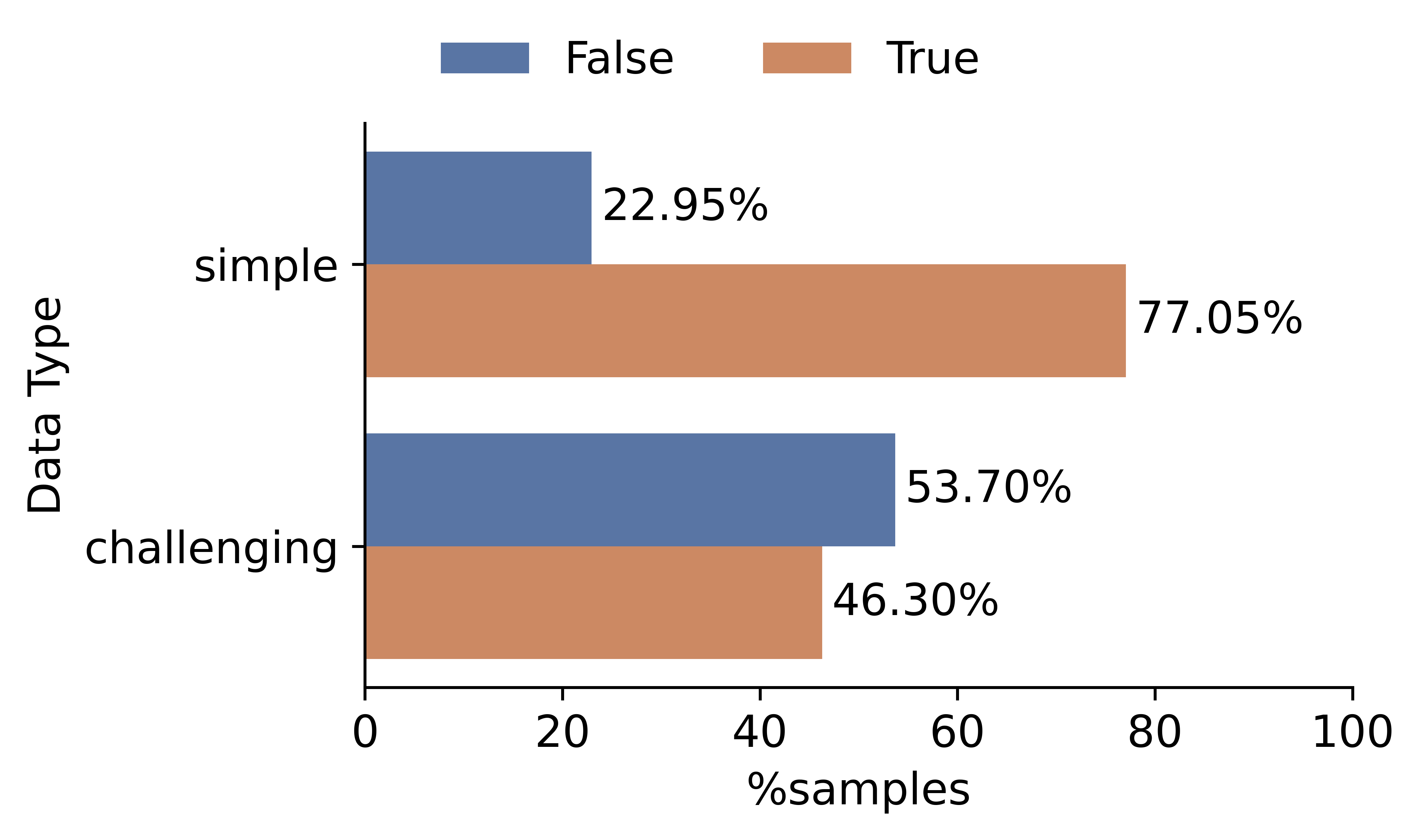}
         \caption{Prompt-based Algorithms}
         \label{fig:sim_measure_Prompt}
     \end{subfigure}
     \caption{Depicts the percentage of samples with or without any nearest neighbor of the same class for supervised algorithms.}
  \label{fig:contextsim}
\end{figure*}
}

\newcommand{\FEWcontextsim}{
\begin{figure*}[h]
     \centering
     \begin{subfigure}[b]{0.3\textwidth}
         \centering
         \includegraphics[width=\textwidth]{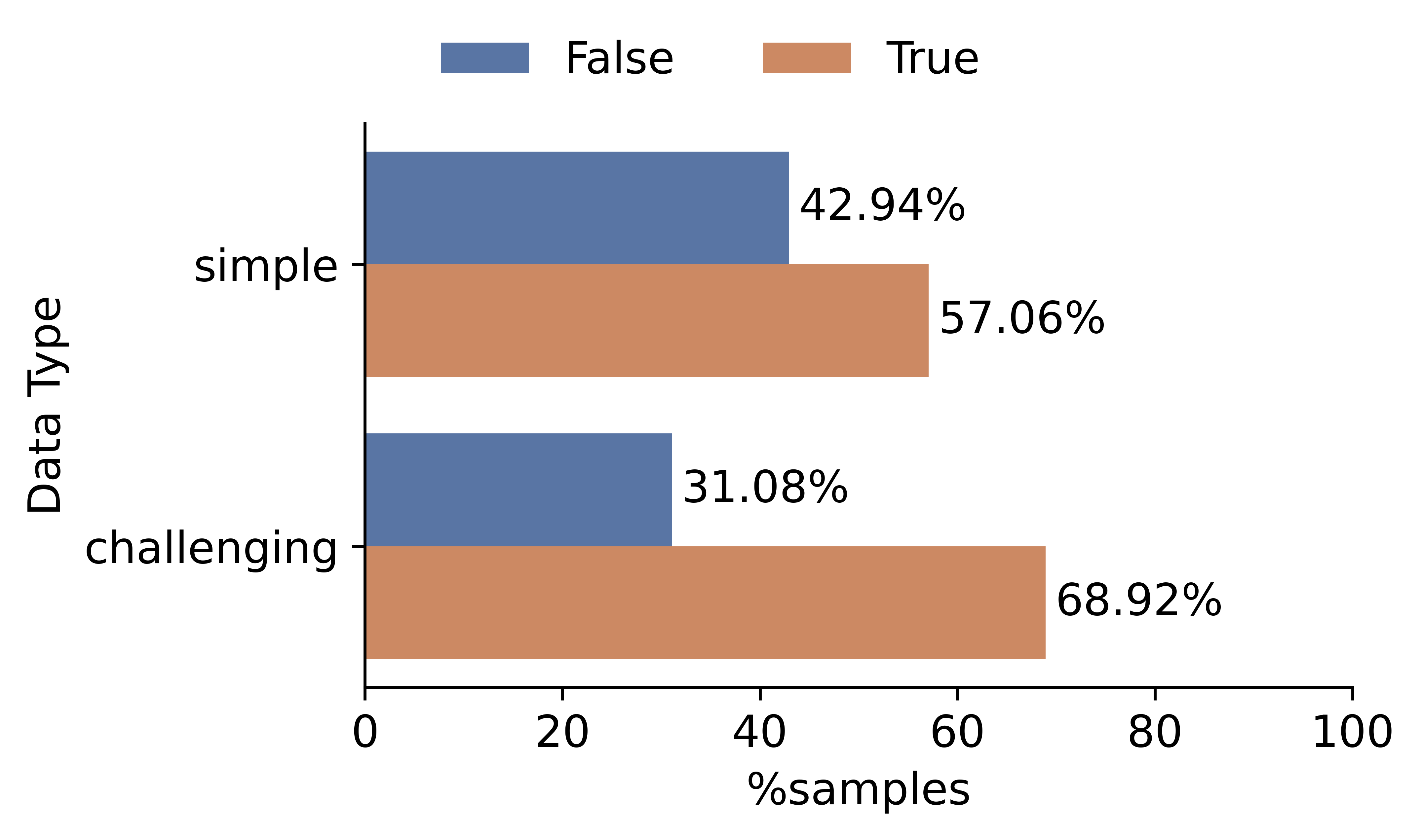}
         \caption{PLM-based Algorithms}
         \label{fig:few_sim_measure_PLM}
     \end{subfigure}
     \hfill
     \begin{subfigure}[b]{0.3\textwidth}
         \centering
         \includegraphics[width=\textwidth]{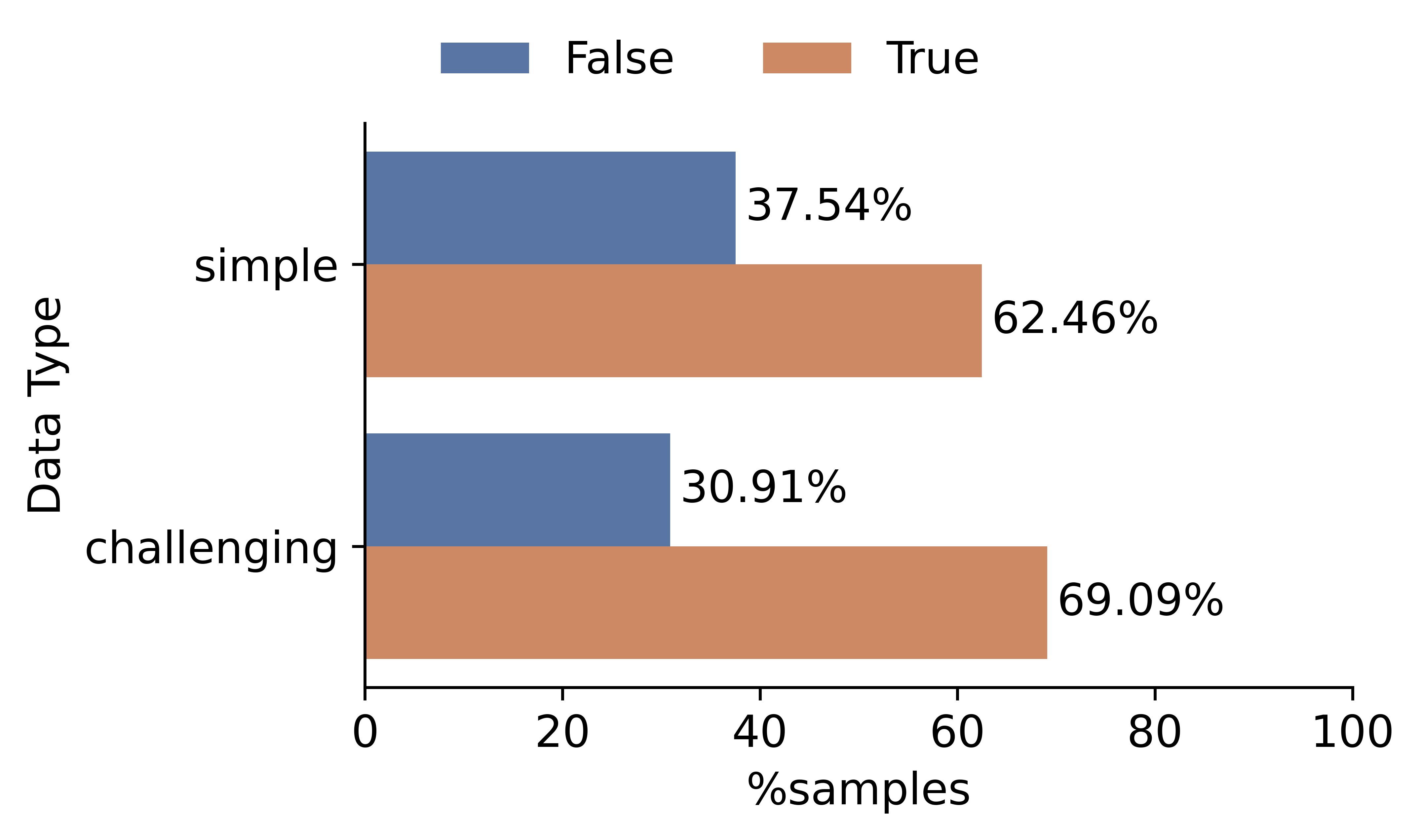}
         \caption{Prompt-based Algorithms}
         \label{fig:few_sim_measure_Prompt}
     \end{subfigure}
     \hfill
     \begin{subfigure}[b]{0.3\textwidth}
         \centering
         \includegraphics[width=\textwidth]{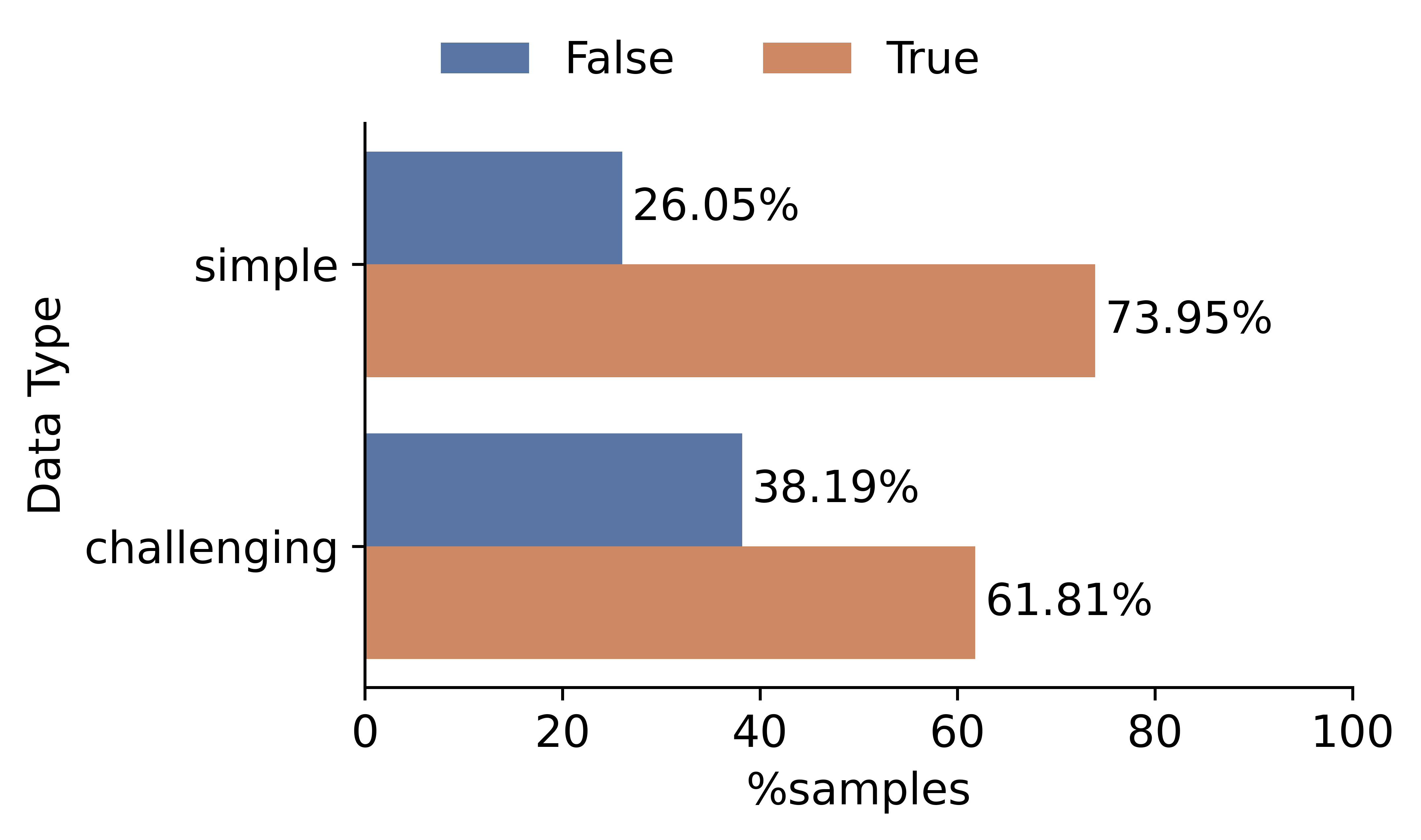}
         \caption{LLM-based Algorithms}
         \label{fig:few_sim_measure_LLM}
     \end{subfigure}
     \caption{Depicts the percentage of samples with or without any nearest neighbor of the same class for few-shot algorithms.}
  \label{fig:FEWcontextsim}
\end{figure*}
}

\newcommand{\missedJoint}{
\begin{figure}[h]
  \centering
  \includegraphics[width=0.75\columnwidth]{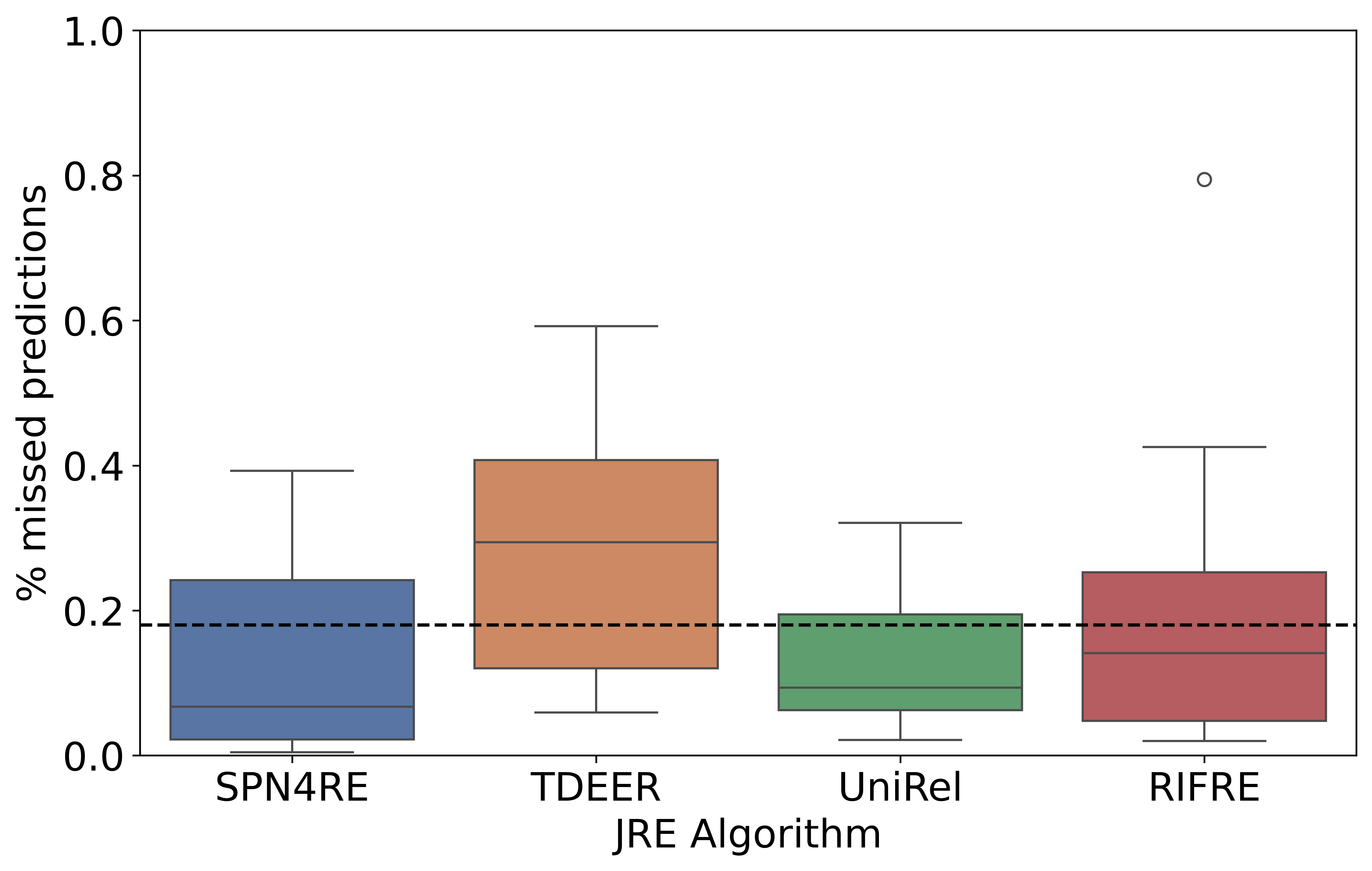}
  \caption{Percentage missed prediction from all test data by joint relation extractors}
  \label{fig:missedJoint}
\end{figure}
}

\newcommand{\LONG}{
\begin{figure}[h]
     \centering
     \begin{subfigure}[b]{\columnwidth}
         \centering
         \includegraphics[width=\columnwidth]{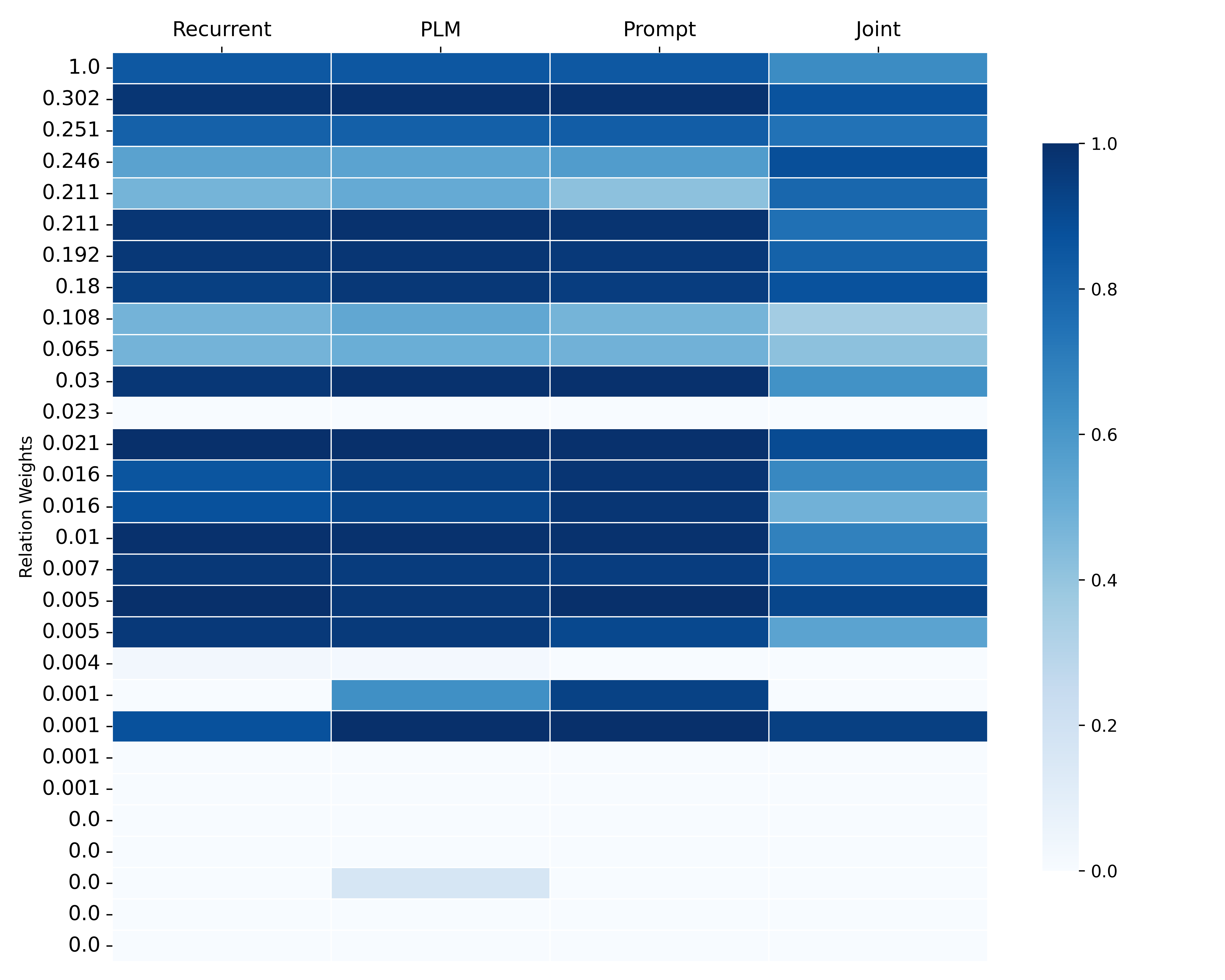}
         \caption{NYT10}
         \label{fig:nytlong}
     \end{subfigure}
     \hfill
     \begin{subfigure}[b]{\columnwidth}
         \centering
         \includegraphics[width=\columnwidth]{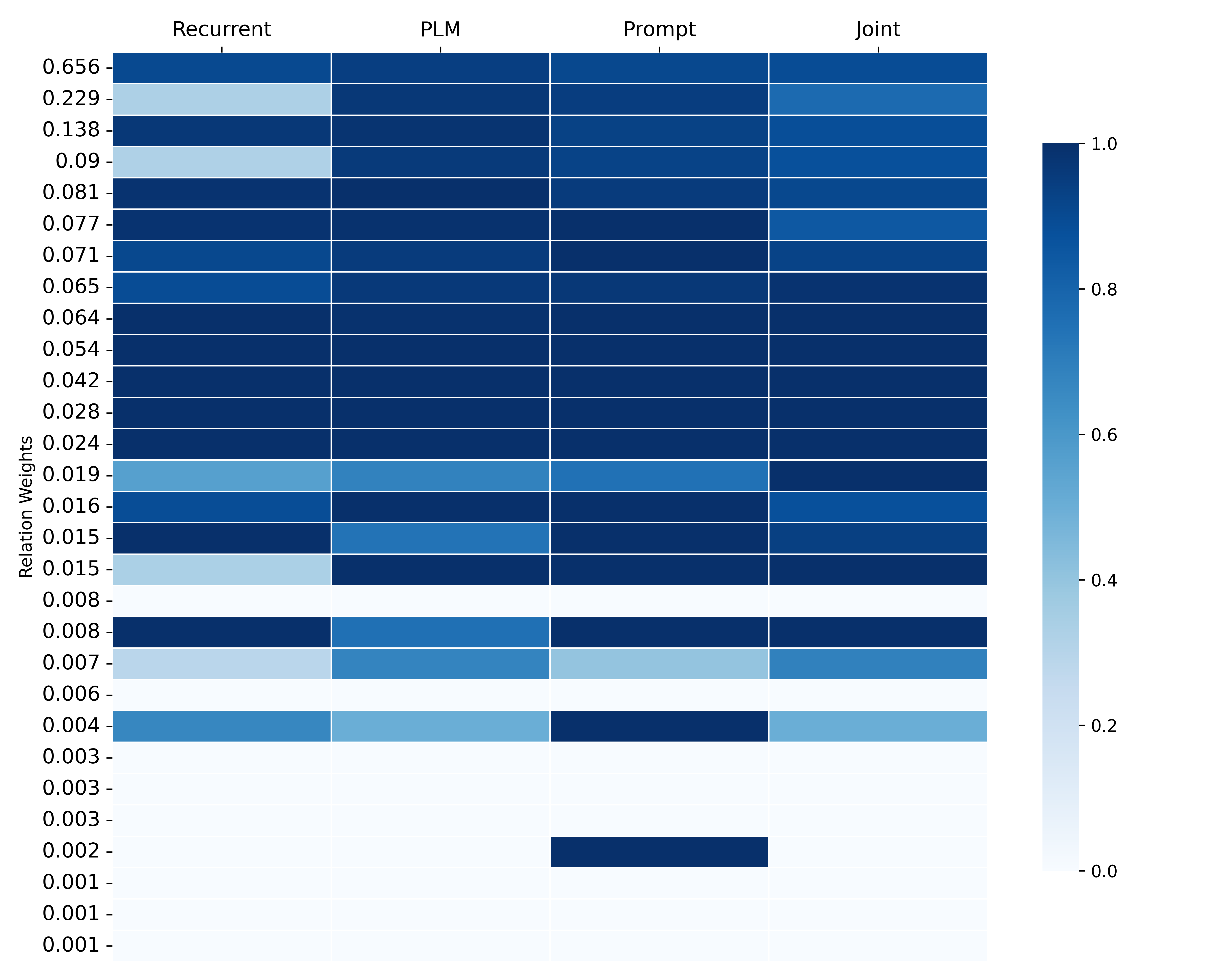}
         \caption{WebNLG (29 relations were randomly sampled due to space constraints.)}
         \label{fig:weblong}
     \end{subfigure}
     \hfill
     \begin{subfigure}[b]{\columnwidth}
         \centering
         \includegraphics[width=\columnwidth]{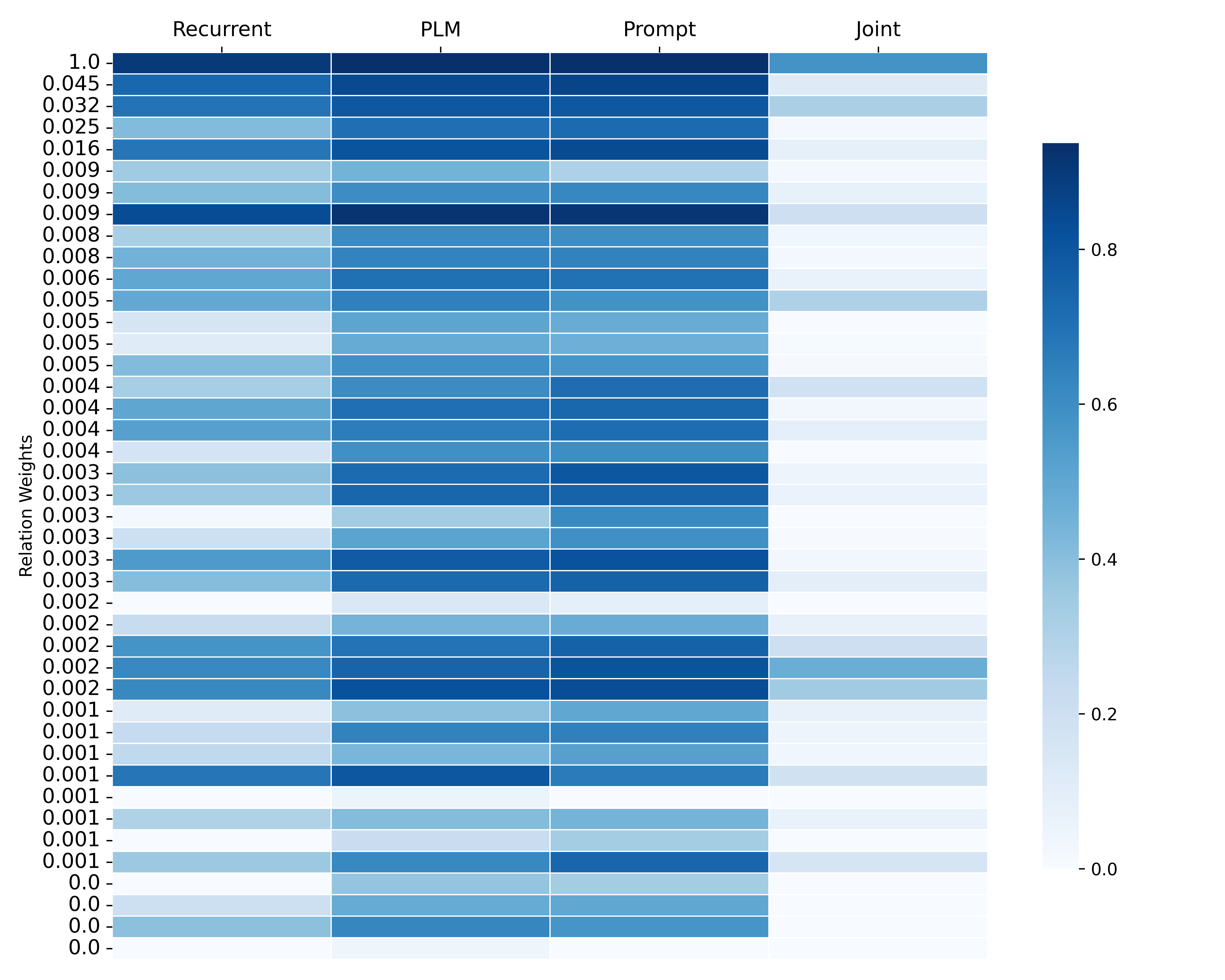}
         \caption{TACRED}
         \label{fig:taclong}
     \end{subfigure}
     \caption{Model performances for relations of the longtail datasets (Micro F1). The ``weights" axis represents the proportion of the relation label in the dataset.}
  \label{fig:long}
\end{figure}
}
\newcommand{\RELTYPE}{
\begin{table}
  \centering
  \caption{Examples of different relation types}\label{tab:datatypes}%
    \begin{tabular}{p{18em}p{9em}}
    \toprule
    \textbf{Sample} & \textbf{Relation} \\
    \midrule
    \multicolumn{2}{p{27em}}{\textbf{Simple Relations}} \\
    1. {\color{blue}Elliot See}, deceased, was a test pilot born in {\color{blue}Dallas} on July 23, 1927. & birthPlace \\
    2. {\color{blue}Renfield} is a fictional character in Bram Stoker's {\color{blue}Dracula}. & characters \\
    
    \midrule
    \multicolumn{2}{p{27em}}{\textbf{Directional Relations}} \\
    1. The index {\color{blue}finger} on my right {\color{blue}hand} has been numb for three days now. & Component-Whole(e1, e2) \\
    2. Private Property - a psychological thriller from {\color{blue}France} starring {\color{blue}Isabelle Huppert}. & per:employee\_of(e2, e1) \\
    \bottomrule
    \end{tabular}%
\end{table}%
}

\newcommand{\DATASETS}{
\begin{table*}[htbp]
  \centering
  \caption{Dataset Characteristics Summary}
    \begin{tabular}{p{8em}lp{1em}p{3.5em}ccp{3.5em}p{3.5em}cp{3.5em}c}
    \toprule
    \textbf{Datasets} & \multicolumn{1}{p{5em}}{\textbf{\#sample}} & \multicolumn{1}{p{5em}}{\textbf{\#rels}} & \multicolumn{8}{p{28em}}{\textbf{Complex Characteristics}} \\
\cmidrule{4-11}    \multicolumn{1}{l}{} &       &       & \textbf{fine- grained} & \multicolumn{1}{p{3.5em}}{\textbf{coarse}} & \multicolumn{1}{p{3.5em}}{\textbf{long- input}} & \textbf{short- input} & \textbf{muliple/ overlap} & \multicolumn{1}{p{3.5em}}{\textbf{one-to- one}} & \textbf{longtail} & \multicolumn{1}{p{3.5em}}{\textbf{uniform}} \\
    \midrule
    SemEval
    \cite{Hendrickx2010-tg} & 10718 & 19    & \multicolumn{1}{c}{} & \multicolumn{1}{p{3.5em}}{$\checkmark$} &       & $\checkmark$   & \multicolumn{1}{c}{} & \multicolumn{1}{p{3.5em}}{$\checkmark$} & $\checkmark$   &  \\
    NYT10 \cite{Riedel2010-yt} & 93598 & 29    & \multicolumn{1}{c}{} &       & \multicolumn{1}{p{3.5em}}{$\checkmark$} & \multicolumn{1}{c}{} & $\checkmark$   &       & $\checkmark$   &  \\
    FewRel \cite{Han2018-bs} & 27328 & 80    & $\checkmark$   &       &       & $\checkmark$   & \multicolumn{1}{c}{} & \multicolumn{1}{p{3.5em}}{$\checkmark$} & \multicolumn{1}{c}{} & \multicolumn{1}{p{3.5em}}{$\checkmark$} \\
    CrossRE \cite{bassignana-plank-2022-crossre} & 19761 & 17    & \multicolumn{1}{c}{} & \multicolumn{1}{p{3.5em}}{$\checkmark$} & \multicolumn{1}{p{3.5em}}{$\checkmark$} & \multicolumn{1}{c}{} & $\checkmark$   &       & $\checkmark$   &  \\
    TACRED \cite{Zhang2017-an} & 106264 & 42    & $\checkmark$   &       & \multicolumn{1}{p{3.5em}}{$\checkmark$} & \multicolumn{1}{c}{} & $\checkmark$   &       & $\checkmark$   &  \\
    RETACRED \cite{stoica2021retacredaddressingshortcomingstacred} & 91467 & 40    & $\checkmark$   &       & \multicolumn{1}{p{3.5em}}{$\checkmark$} & \multicolumn{1}{c}{} & $\checkmark$   &       & $\checkmark$   &  \\
    WebNLG \cite{gardent2017creating} & 14486 & 171   & $\checkmark$   &       &       & $\checkmark$   & $\checkmark$   &       & $\checkmark$   &  \\
    \bottomrule
    \end{tabular}%
  \label{tab:procon}%
\end{table*}%
}

\newcommand{\ALGOS}{
\begin{table}[htbp]
\centering
\caption{Overview of Algorithms and Architectures}
\begin{threeparttable}[t]
\begin{tabular}{@{}p{0.5cm}p{0.9cm}p{1.3cm}p{1.1cm}p{0.8cm}p{0.6cm}p{0.6cm}@{}}
\toprule
Task  & Type      & Algorithm       & Arch.$^1$                & Params. & Super. & Few. \\ 
\midrule
RC    & Recurrent & Att-BLSTM      & BLSTM                & 2M     & $\checkmark$    &   \\ 
&           & PAWARE       & BLSTM                & 15M    & $\checkmark$     &   \\ 
      &           & EntityAtt       & BLSTM                & 8M    & $\checkmark$     &   \\ 
\cmidrule{2-7}
  &   PLM        & R-BERT          & BERT                 & 110M   & $\checkmark$    & $\checkmark$  \\ 
      &           & Roberta\_ base   & RoBERTa (large)      & 355M   & $\checkmark$    &  $\checkmark$ \\ 
      &           & LUKE            & RoBERTa (large)      & 483M   & $\checkmark$    &   \\ 
      &           & ERNIE           & BERT enc+dec$^2$       & 114M   & $\checkmark$    &   \\ 
\cmidrule{2-7}
      & Prompt-PLM & Know-Prompt     & RoBERTa (large)      & 355M   & $\checkmark$    &  $\checkmark$ \\ 
      &            & GenPT           & RoBERTa (large)      & 355M   &     &  $\checkmark$ \\ 
\cmidrule{2-7}
     & Prompt-LLM & GPTRE          & gpt-3.5-turbo        & NA$^3$     &     &  $\checkmark$ \\ 
      &            & Unleash-LLM     & gpt-3.5-turbo        & NA$^3$     &     &  $\checkmark$ \\ 
\midrule
JRE   &            & SPN4RE          & BERT enc+dec$^2$       & 110M   & $\checkmark$    &   \\ 
      &            & TDEER           & BERT encoder         & 110M   & $\checkmark$    &   \\ 
      &            & RIFRE           & BERT enc+dec$^2$       & 109M   & $\checkmark$    &   \\ 
      &            & UniRel          & BERT encoder         & 109M   & $\checkmark$    &   \\ 
\bottomrule
\end{tabular}
    \begin{tablenotes}
    \item[1] Arch.$\rightarrow$Architecture; Params$\rightarrow$Parameters; Sup.$\rightarrow$Supervised Learning; Few.$\rightarrow$Few-shot learning
    \item[2] enc+dec$\rightarrow$encoder+decoder
    \item[3] Number of parameters not disclosed officially
\end{tablenotes}
\end{threeparttable}
\label{tab:algorithms}
\end{table}
}

\newcommand{\RES}{
\begin{table*}[htbp]
  \centering
  \caption{Simplified summary of quantitative findings}
  \begin{threeparttable}[t]
    \begin{tabular}{p{2.835em}llp{2.75em}p{2.75em}p{2.75em}p{2.75em}p{3.25em}p{2.585em}}
    \toprule
    \multirow{2}[4]{*}{\textbf{Task}} & \multicolumn{1}{l}{\multirow{2}[4]{*}{\textbf{Method}}} & \multicolumn{1}{l}{\multirow{2}[4]{*}{\textbf{Type}}} & \multicolumn{6}{p{16.835em}}{\textbf{Complex Characteristics$^1$}} \\
\cmidrule{4-9}    \multicolumn{1}{l}{} &       &       & \textbf{fine- grained} & \textbf{coarse} & \textbf{long- input} & \textbf{short-input} & \textbf{muliple/ overlap} & \textbf{long- tail} \\
    \midrule
    \multirow{3}[2]{*}{RC} & \multicolumn{1}{l}{\multirow{3}[2]{*}{Supervised}} & \multicolumn{1}{p{3.25em}}{Recurrent} & -    & -    & $\checkmark$   & -    & $\times$    & $\times$ \\
    \multicolumn{1}{c}{} &       & \multicolumn{1}{p{3.25em}}{PLM} & $\checkmark$   & -    & $\checkmark$   & -    & $\times$    & $\times$ \\
    \multicolumn{1}{c}{} &       & \multicolumn{1}{p{3.25em}}{Prompt-PLM} & $\checkmark$   & -    & $\checkmark$   & -    & $\times$    & $\times$ \\
    \midrule
    \multirow{3}[2]{*}{RC} & \multicolumn{1}{l}{\multirow{3}[2]{*}{Few-shot}} & \multicolumn{1}{p{3.25em}}{PLM} & -    & $\times$    & $\times$    & -   & -    & \multicolumn{1}{l}{} \\
    \multicolumn{1}{c}{} &       & \multicolumn{1}{p{3.25em}}{Prompt-PLM} & -    & $\times$    & $\times$    & -   & -    & \multicolumn{1}{l}{} \\
    \multicolumn{1}{c}{} &       & \multicolumn{1}{p{3.25em}}{Prompt-LLM} & $\times$    & $\checkmark$   & $\checkmark$   & -    & -    & \multicolumn{1}{l}{} \\
    \midrule
    JRE   &       &       & $\checkmark$   & -    & -    & -    & $\times$    & $\times$ \\
    \bottomrule
    \end{tabular}%
    \begin{tablenotes}
    \item[1] ``$\checkmark$''$\rightarrow$ significant performance gain in this category; ``$\times$''$\rightarrow$ significant performance degradation in this category; ``-''$\rightarrow$ strong inference could not be drawn; ``blank rows'' indicate that the algorithms were not evaluated for the complex category
\end{tablenotes}
\end{threeparttable}
  \label{tab:result}%
\end{table*}%
}

\newcommand{\AMB}{
\begin{table}
  \centering
  \caption{Examples of ambiguity faced by relation extractors}\label{tab:amb}
    \begin{tabular}{p{15em}p{6em}p{5.5em}}
    \toprule
    \textbf{Ambiguous Samples} & \textbf{True} & \textbf{Predicted} \\
    \midrule
    \multicolumn{3}{p{25em}}{\textbf{Context does not provide sufficient information}} \\
    One of {\color{blue}Hasse Ekman's} movies, {\color{blue}"Kungliga Patrasket"} from 1945 about a famous acting family, was partly based on his own upbringing in an acting family. & screen\_writer & director \\
    \multicolumn{3}{p{25em}}{\textbf{Context provides contradictory/multiple information}} \\
    The {\color{blue}"Star Wars Jedi Knight: Mysteries of the Sith"} was developed and published by {\color{blue}LucasArts} as an expansion to "Star Wars Jedi Knight: Dark Forces II." & developer & publisher \\
    \midrule
    \multicolumn{3}{p{25em}}{\textbf{Relation definitions are not distinctive}} \\
    A memorial service will be held in {\color{blue} Annapolis}, {\color{blue} Maryland}, on a date to be announced. & /location/ us\_state/capital & /location/ location/contains \\
    \bottomrule
    \end{tabular}%
\end{table}
}

\newcommand{\RELCORR}{
\begin{table*}[htbp]
  \centering
  \caption{Examples of correlating relations. Depicts the percentage of samples in a relation category misclassified as the correlating relation.}
    \begin{tabular}{ll}
    \toprule
    \textbf{True Relation} & \textbf{Mispredicted Relation} \\
    \midrule
    /business/company\_advisor/companies\_advised & /business/person/company \\
    /film/film\_location/featured\_in\_films & /location/location/contains \\
    /location/br\_state/capital & /location/location/contains \\
    /location/us\_county/county\_seat & /location/location/contains \\
    /location/us\_state/capital & /location/location/contains \\
    EISSN\_number & LCCN\_number \\
    Entity-Destination(e2,e1) & Entity-Destination(e1,e2) \\
    chief & leaderName \\
    elevationAboveTheSeaLevel\_(in\_feet) & elevationAboveTheSeaLevel\_(in\_metres) \\
    fossil & gemstone \\
    mainIngredients & ingredient \\
    notableWork & precededBy \\
    officialLanguage & language \\
    \bottomrule
    \end{tabular}%
  \label{tab:corr_rel}%
\end{table*}%
}


\newcommand{\superDATA}{
\begin{table}[htbp]
  \centering
  \caption{Train and test statistics for supervised relation classification (RC) and joint relation extraction (JRE) experiments.}
    \begin{tabular}{lllll}
    \toprule
    \multirow{2}[4]{*}{\textbf{Datasets}} & \multicolumn{2}{c}{\textbf{RC}} & \multicolumn{2}{c}{\textbf{JRE}} \\
\cmidrule{2-5}          & \textbf{\#train} & \textbf{\#test} & \textbf{\#train} & \textbf{\#test} \\
    \midrule
    CrossRE & 9875  & 9168  & 2122  & 1826 \\
    NYT10 & 87739 & 5859  & 77366 & 4006 \\
    TACRED & 90755 & 15509 & 47523 & 6277 \\
    RETACRED & 78049 & 13418 & 42808 & 5805 \\
    WebNLG & 12894 & 1591  & 5512  & 701 \\
    SemEval & 8001  & 2717  & 7988  & 2714 \\
    FewRel & 20014 & 7309  & 19876 & 7271 \\
    \bottomrule
    \end{tabular}%
  \label{aptab:superData}%
\end{table}%
}

\newcommand{\multiRC}{
\begin{table*}[htbp]
  \centering
  \caption{Test data statistics for supervised relation classification data.}
    \begin{tabular}{llllllllll}
    \toprule
    \multicolumn{1}{c}{\multirow{2}[4]{*}{\textbf{Datasets}}} & \multicolumn{1}{c}{\multirow{2}[4]{*}{\textbf{\#test}}} & \multicolumn{5}{c}{Multiple Relations} & \multicolumn{3}{c}{Overlapping Entities} \\
\cmidrule{3-10}          &       & \textbf{n1} & \textbf{n2} & \textbf{n3} & \textbf{n4} & \textbf{n5} & \textbf{norm} & \textbf{seo} & \textbf{epo} \\
    \midrule
    CrossRE & 9168  & 268   & 534   & 744   & 1044  & 6578  & 418   & 6989  & 1761 \\
    NYT10 & 5859  & 2950  & 1190  & 561   & 956   & 202   & 2976  & 708   & 2175 \\
    tacred & 15509 & 3289  & 2378  & 1902  & 1564  & 6376  & 3319  & 9272  & 2918 \\
    retacred & 13418 & 3154  & 2230  & 1782  & 1268  & 4984  & 3157  & 7818  & 2443 \\
    WebNLG & 1591  & 262   & 346   & 393   & 360   & 230   & 242   & 1247  & 102 \\
    SemEval & 2717  & 2709  & 6     & 0     & 0     & 0     & 2707  & 8     & 0 \\
    FewRel & 7309  & 7231  & 78    & 0     & 0     & 0     & 7239  & 68    & 2 \\
    \bottomrule
    \end{tabular}%
  \label{aptab:RC_multi}%
\end{table*}%
}

\newcommand{\multiJRE}{
\begin{table*}[htbp]
  \centering
  \caption{Test data statistics for supervised joint relation extraction data.}
    \begin{tabular}{llllllllll}
    \toprule
    \multicolumn{1}{c}{\multirow{2}[4]{*}{\textbf{Datasets}}} & \multicolumn{1}{c}{\multirow{2}[4]{*}{\textbf{\#test}}} & \multicolumn{5}{c}{Multiple Relations} & \multicolumn{3}{c}{Overlapping Entities} \\
\cmidrule{3-10}          &       & \multicolumn{1}{c}{\textbf{n1}} & \multicolumn{1}{c}{\textbf{n2}} & \multicolumn{1}{c}{\textbf{n3}} & \multicolumn{1}{c}{\textbf{n4}} & \multicolumn{1}{c}{\textbf{n5}} & \multicolumn{1}{c}{\textbf{norm}} & \multicolumn{1}{c}{\textbf{seo}} & \multicolumn{1}{c}{\textbf{epo}} \\
    \midrule
    CrossRE & 1826  & 268   & 267   & 248   & 261   & 782   & 328   & 1481  & 235 \\
    NYT10 & 4006  & 2950  & 595   & 187   & 239   & 35    & 2963  & 742   & 715 \\
    TACRED & 6277  & 3289  & 1189  & 634   & 391   & 774   & 3260  & 2994  & 380 \\
    RETACRED & 5805  & 3154  & 1115  & 594   & 317   & 625   & 3110  & 2674  & 345 \\
    WebNLG & 701   & 262   & 173   & 131   & 90    & 45    & 242   & 457   & 28 \\
    SemEval & 2714  & 2709  & 3     & 0     & 0     & 0     & 2706  & 6     & 0 \\
    FewRel & 7271  & 7232  & 39    & 0     & 0     & 0     & 7236  & 34    & 1 \\
    \bottomrule
    \end{tabular}%
  \label{aptab:JRE_multi}%
\end{table*}%
}

\newcommand{\fewDATA}{
\begin{table}[htbp]
  \centering
  \caption{Train and test data for few-shot experiments with respect to different shots of data.}
  \begin{threeparttable}[t]
    \begin{tabular}{llll}
    \toprule
    \textbf{Datasets} & \textbf{k} & \textbf{\#train} & \textbf{\#test} \\
    \midrule
    \multirow{5}[2]{*}{CrossRE} & 1     & 34    & 3026 \\
          & 5     & 167   & 3026 \\
          & 10    & 332   & 3026 \\
          & 20    & 654   & 3026 \\
          & 30    & 969   & 3026 \\
    \midrule
    \multirow{5}[2]{*}{NYT10} & 1     & 58    & 1934 \\
          & 5     & 270   & 1934 \\
          & 10    & 524   & 1934 \\
          & 20    & 964   & 1934 \\
          & 30    & 1370  & 1934 \\
    \midrule
    \multirow{5}[2]{*}{TACRED} & 1     & 84    & 5118 \\
          & 5     & 420   & 5118 \\
          & 10    & 834   & 5118 \\
          & 20    & 1644  & 5118 \\
          & 30    & 2428  & 5118 \\
    \midrule
    \multirow{5}[2]{*}{RETACRED} & 1     & 80    & 4428 \\
          & 5     & 400   & 4428 \\
          & 10    & 793   & 4428 \\
          & 20    & 1573  & 4428 \\
          & 30    & 2332  & 4428 \\
    \midrule
    \multirow{5}[2]{*}{WebNLG} & 1     & 280   & 535 \\
          & 5     & 1121  & 535 \\
          & 10    & 1897  & 535 \\
          & 20    & 3007  & 535 \\
          & 30    & 3830  & 535 \\
    \midrule
    \multirow{5}[2]{*}{SemEval$^1$} & 1     & 38    & 897 \\
          & 5     & 182   & 897 \\
          & 10    & 362   & 897 \\
          & 20    & 722   & 897 \\
          & 30    & 1078  & 897 \\
    \midrule
    \multirow{5}[2]{*}{FewRel} & 1     & 160   & 2412 \\
          & 5     & 800   & 2412 \\
          & 10    & 1600  & 2412 \\
          & 20    & 3200  & 2412 \\
          & 30    & 4800  & 2412 \\
    \bottomrule
    \end{tabular}%
    \begin{tablenotes}
    \item[1] SemEval-2018 Task 8 dataset restructured from directional to semantic relations to make them compatible with LLM-based algorithms
\end{tablenotes}
\end{threeparttable}
  \label{aptab:fewData}%
\end{table}%
}

\newcommand{\overlapFEW}{
\begin{table*}[htbp]
  \centering
  \caption{Test data statistics for multiple relations and overlapping entities used for few-shot experiments.}
    \begin{tabular}{clllllllll}
    \toprule
    \multirow{2}[3]{*}{\textbf{Dataset}} & \multicolumn{1}{c}{\multirow{2}[3]{*}{\textbf{seed}}} & \multicolumn{3}{c}{\textbf{Overlapping Entities}} & \multicolumn{5}{c}{\textbf{Multiple Relations}} \\
\cmidrule{3-10}          &       & \textbf{norm} & \textbf{seo} & \textbf{epo} & \textbf{n1} & \textbf{n2} & \textbf{n3} & \textbf{n4} & \textbf{n5} \\
    \midrule
    \multirow{3}[2]{*}{CrossRE} & 13    & 130   & 2339  & 557   & 76    & 171   & 251   & 328   & 2200 \\
          & 42    & 144   & 2284  & 598   & 86    & 187   & 235   & 337   & 2181 \\
          & 100   & 128   & 2311  & 587   & 84    & 169   & 233   & 364   & 2176 \\
    \midrule
    \multirow{3}[2]{*}{NYT10} & 13    & 990   & 228   & 716   & 982   & 391   & 184   & 304   & 73 \\
          & 42    & 1008  & 219   & 707   & 998   & 377   & 189   & 302   & 68 \\
          & 100   & 964   & 214   & 756   & 957   & 389   & 193   & 328   & 67 \\
    \midrule
    \multirow{3}[2]{*}{TACRED} & 13    & 1134  & 3004  & 980   & 1117  & 776   & 626   & 506   & 2093 \\
          & 42    & 1112  & 3035  & 971   & 1103  & 757   & 629   & 517   & 2112 \\
          & 100   & 1086  & 3081  & 951   & 1075  & 824   & 604   & 530   & 2085 \\
    \midrule
    \multirow{3}[2]{*}{RETACRED} & 13    & 1024  & 2609  & 795   & 1035  & 717   & 604   & 415   & 1657 \\
          & 42    & 1005  & 2626  & 797   & 1010  & 712   & 581   & 444   & 1681 \\
          & 100   & 1028  & 2617  & 783   & 1029  & 770   & 620   & 424   & 1585 \\
    \midrule
    \multirow{3}[2]{*}{WebNLG} & 13    & 78    & 420   & 37    & 86    & 113   & 136   & 118   & 82 \\
          & 42    & 85    & 413   & 37    & 91    & 115   & 140   & 117   & 72 \\
          & 100   & 89    & 409   & 37    & 94    & 111   & 130   & 128   & 72 \\
    \midrule
    \multirow{3}[2]{*}{SemEval} & 13    & 892   & 5     & 0     & 895   & 2     & 0     & 0     & 0 \\
          & 42    & 891   & 6     & 0     & 892   & 5     & 0     & 0     & 0 \\
          & 100   & 895   & 2     & 0     & 895   & 2     & 0     & 0     & 0 \\
    \midrule
    \multirow{3}[2]{*}{FewRel} & 13    & 2394  & 18    & 0     & 2391  & 21    & 0     & 0     & 0 \\
          & 42    & 2395  & 16    & 1     & 2391  & 21    & 0     & 0     & 0 \\
          & 100   & 2389  & 22    & 1     & 2385  & 27    & 0     & 0     & 0 \\
    \bottomrule
    \end{tabular}%
  \label{aptab:few_multi}%
\end{table*}%
}

\newcommand{\entSTATS}{
\begin{table}[htbp]
  \centering
  \caption{Percentage entities tagged using Stanford CoreNLP}
  \begin{threeparttable}[t]
    \begin{tabular}{llll}
    \toprule
    \multicolumn{1}{c}{\textbf{Dataset}} & \multicolumn{1}{c}{\textbf{\%subj+obj$^1$}} & \multicolumn{1}{c}{\textbf{\%only\_subj$^2$}} & \multicolumn{1}{c}{\textbf{\%only\_obj$^3$}} \\
    \midrule
    TACRED & 1.00     & 0.0     & 0.0 \\
    RETACRED & 1.00     & 0.0     & 0.0 \\
    WebNLG & 0.81  & 0.09  & 0.07 \\
    FewRel & 0.75  & 0.17  & 0.06 \\
    CrossRE & 0.63  & 0.16  & 0.07 \\
    NYT10 & 0.1   & 0.2   & 0.15 \\
    SemEval & 0.0     & 0.01  & 0.02 \\
    \bottomrule
    \end{tabular}%
    \begin{tablenotes}
        \item[1] annotations available for both subject and object entities
        \item[2] annotations available for only subject entities
        \item[3] annotations available for only object entities
    \end{tablenotes}
    \end{threeparttable}
  \label{aptab:ent_stats}%
\end{table}%
}

\newcommand{\superPERF}{
\begin{table*}[htbp]
  \centering
  \caption{Performance (Micro-F1) of supervised relation classification algorithms. The highest performing algorithm for each dataset has been highlighted.}
  \begin{threeparttable}[t]
    \begin{tabular}{llllllll}
    \toprule
    \textbf{Method} & \textbf{FewRel} & \textbf{NYT10} & \textbf{WebNLG} & \textbf{crossRE} & \textbf{SemEval} & \textbf{TACRED} & \textbf{RETACRED} \\
    \midrule
    Att-BLSTM & 73.7$\pm$0.26 & 79.37$\pm$0.45 & 84.05$\pm$0.31 & 53.86$\pm$1.17 & 80.87$\pm$0.51 & 54.77$^1$ & 71.52$^1$ \\
    PAWARE & 71.4$\pm$0.29 & 78.95$\pm$0.15 & 93.29$\pm$0.62 & 53.62$\pm$0.72 & 77.98$\pm$0.64 & 59.46$\pm$0.57 & 76.65$\pm$0.51 \\
    Entity-Att & 68.14$\pm$0.32 & 79.77$\pm$0.25 & 85.0$\pm$0.49 & 48.62$\pm$1.71 & 76.47$\pm$0.28 & 46.4$\pm$1.62 & 67.11$\pm$0.53 \\
    RBERT & 89.31$\pm$0.34 & \textbf{81.36$\pm$0.18} & 90.52$\pm$1.66 & \textbf{75.06$\pm$0.14} & \textbf{88.71$\pm$0.39} & 69.61$\pm$0.34 & 89.14$\pm$0.2 \\
    Roberta\_base & 89.25$\pm$0.26 & 80.65$\pm$0.04 & 93.66$\pm$0.22 & 72.83$\pm$0.68 & 87.0$\pm$0.45 & \textbf{75.09$\pm$0.68} & \textbf{91.55$\pm$0.2} \\
    LUKE  & \textbf{90.55$\pm$0.08} & 80.41$\pm$0.41 & \textbf{95.61$\pm$0.14} & 74.29$\pm$0.4 & 88.15$\pm$0.48 & 72.02$\pm$0.34 & 90.88$\pm$0.2 \\
    ERNIE & 83.95$\pm$0.66 & 80.91$\pm$0.2 & 91.79$\pm$0.3 & 71.03$\pm$1.27 & 82.77$\pm$0.24 & 65.19$\pm$1.62 & 83.98$\pm$1.27 \\
    KnowPrompt & 88.38$\pm$0.16 & 76.9$\pm$3.33 & 95.32$\pm$0.65 & 74.84$\pm$0.79 & 88.21$\pm$0.52 & 62.06$\pm$22.11 & 90.54$\pm$0.18 \\
    \bottomrule
    \end{tabular}%
            \begin{tablenotes}
\item[1] trained on a single fold
\end{tablenotes}
\end{threeparttable}
  \label{aptab:superPERF}%
\end{table*}%
}

\newcommand{\fewPERF}{
\begin{table*}[htbp]
  \centering
  \caption{Performance (Micro-F1) of few-shot relation classification algorithms.}
  \begin{threeparttable}[t]
    \begin{tabular}{lllllll}
    \toprule
    \multicolumn{1}{c}{\multirow{2}[2]{*}{\textbf{Method}}} & \multicolumn{1}{c}{\multirow{2}[2]{*}{\textbf{Dataset}}} & \multicolumn{5}{c}{\textbf{k}} \\
    \cmidrule{3-7}
          &       & \multicolumn{1}{c}{\textbf{1}} & \multicolumn{1}{c}{\textbf{5}} & \multicolumn{1}{c}{\textbf{10}} & \multicolumn{1}{c}{\textbf{20}} & \multicolumn{1}{c}{\textbf{30}} \\
    \midrule
    \multirow{7}[2]{*}{RBERT} & FewRel & 5.11  & 30.18 & 55.2  & 69.88 & 77.83 \\
          & NYT10 & 2.02  & 12.93 & 23.53 & 40.52 & 48.16 \\
          & WebNLG & 10.28 & 40.19 & 57.26 & 64.11 & 69.97 \\
          & CrossRE & 7.25  & 16.33 & 20.86 & 27.75 & 38.28 \\
          & RETACRED & 2.05  & 8.86  & 18.67 & 34.65 & 42.63 \\
          & SemEval$^2$ & 10.55 & 16.4  & 26.74 & 39.5  & 48.19 \\
          & TACRED & 1.97  & 4.7   & 9.84  & 17.78 & 25.01 \\
    \midrule
    \multirow{7}[2]{*}{Roberta\_base} & FewRel & 1.52  & 3.68  & 11.14 & 36.46 & 50.88 \\
          & NYT10 & 6     & 3.27  & 5.31  & 6.93  & 10.51 \\
          & WebNLG & 0.44  & 4.67  & 18.07 & 42.18 & 50.47 \\
          & CrossRE & 3.26  & 7.82  & 6.72  & 5.42  & 20.9 \\
          & RETACRED & 0.61  & 1.74  & 12.42 & 20.39 & 27.65 \\
          & SemEval$^2$ & 9.13  & 9.59  & 4.7   & 14.58 & 12.61 \\
          & TACRED & 1.26  & 1.79  & 1.95  & 10.56 & 19.61 \\
    \midrule
    \multirow{7}[2]{*}{UnleashLLM$^1$} & FewRel & 35.67 &    -   &    -   &   -    & - \\
          & NYT10 & 50.66 &   -    &   -    &    -   & - \\
          & WebNLG &   -    &  -     &    -   &    -   & - \\
          & CrossRE & 20.04 &     -  &    -   &    -   &  -\\
          & RETACRED & 40.4  &   -    &  -     &   -    & - \\
          & SemEval$^2$ & 50.51 &    -   &   -    &    -   & - \\
          & TACRED & 44.69 &     -  &    -   &     -  & - \\
    \midrule
    \multirow{7}[2]{*}{GPT-RE$^1$} & FewRel & 47.52 & 54.1  & 57.02 & 57.64 & 58.36 \\
          & NYT10 & 74.26 & 79.49 & 77.29 & 76.03 & 74.73 \\
          & WebNLG & 27.79 & 27.78 & 28.9  & 27.47 & 27.89 \\
          & CrossRE & 32.71 & 41.99 & 45.07 & 47.07 & 48.75 \\
          & RETACRED & 51.91 & 69.5  & 71.71 & 74.13 & 74.96 \\
          & SemEval$^2$ & 68.35 & 75.04 & 73.2  & 72.27 & 71.8 \\
          & TACRED & 36.52 & 45.56 & 47.91 & 50.71 & 53.4 \\
    \midrule
    \multirow{7}[2]{*}{GenPT/Bart} & FewRel & 44.87 & 70.29 & 76.73 & 82.44 & 84.56 \\
          & NYT10 & 10.65 & 24.23 & 33.42 & 40.69 & 44.4 \\
          & WebNLG & 49.1  & 74.14 & 78.88 & 85.11 & 87.41 \\
          & CrossRE & 4.58  & 27.34 & 38.81 & 45.92 & 53.18 \\
          & RETACRED & 33.7  & 51.91 & 54.53 & 60.84 & 60.37 \\
          & SemEval$^2$ & 10.31 & 36.79 & 52.41 & 65.04 & 51.01 \\
          & TACRED & 20.54 & 30.3  & 32.73 & 34.79 & 35.03 \\
    \midrule
    \multirow{7}[2]{*}{GenPT/T5} & FewRel & 11.25 & 54.55 & 73.49 & 81.8  & 84.08 \\
          & NYT10 & 6.15  & 20.1  & 35.33 & 44.83 & 47.81 \\
          & WebNLG & 43.74 & 66.6  & 78.32 & 85.36 & 87.6 \\
          & CrossRE & 5.55  & 10.11 & 21.41 & 43.13 & 50.15 \\
          & RETACRED & 35.06 & 45.09 & 53.56 & 57.91 & 60.93 \\
          & SemEval$^2$ & 6.87  & 16.48 & 38.4  & 53.75 & 63.59 \\
          & TACRED & 18.35 & 25.7  & 29.64 & 33.49 & 33.86 \\
    \midrule
    \multirow{7}[2]{*}{GenPT/roberta} & FewRel & 32.17 & 73.66 & 79.41 & 82.82 & 84.63 \\
          & NYT10 & 3.77  & 29.06 & 35.33 & 48.71 & 50.16 \\
          & WebNLG & 53.77 & 71.02 & 79.57 & 86.04 & 87.85 \\
          & CrossRE & 5.31  & 31.39 & 42.59 & 51.75 & 58.8 \\
          & RETACRED & 16.73 & 52.16 & 56.9  & 61.5  & 63.89 \\
          & SemEval$^2$ & 3.88  & 34.33 & 53.57 & 68.91 & 75.6 \\
          & TACRED & 18.86 & 25.34 & 21.41 & 22.18 & 20.01 \\
    \midrule
    \multirow{7}[2]{*}{KnowPrompt} & FewRel & 7.41  & 37.11 & 55.76 & 73.3  & 80.22 \\
          & NYT10 & 0.29  & 1.55  & 10.94 & 24.75 & 32.37 \\
          & WebNLG & 4.05  & 47.6  & 60.25 & 71.96 & 81.62 \\
          & CrossRE & 5.98  & 5.23  & 3.91  & 21.33 & 28.64 \\
          & RETACRED & 0.83  & 9.07  & 23.91 & 37.62 & 47.76 \\
          & SemEval$^2$ & 11.8  & 11.8  & 12    & 23.89 & 41.1 \\
          & TACRED & 0.88  & 5.38  & 17.08 & 27.58 & 30.87 \\
    \bottomrule
    \end{tabular}%
            \begin{tablenotes}
\item[1] excluding samples where GPT predicted out-of-domain labels
\item[2] SemEval-2018 Task 8 dataset restructured from directional to semantic relations to make them compatible with LLM-based algorithms
\end{tablenotes}
\end{threeparttable}
  \label{aptab:fewPERF}%
\end{table*}%
}

\newcommand{\jointPERF}{
\begin{table*}[htbp]
  \centering
  \caption{Performance (Micro-F1) of supervised joint relation extraction algorithms. The highest performing algorithm for each dataset has been highlighted.}
    \begin{tabular}{llllllll}
    \toprule
    \textbf{Method} & \textbf{FewRel} & \textbf{NYT10} & \textbf{WebNLG} & \textbf{crossRE} & \textbf{SemEval} & \textbf{TACRED} & \textbf{RETACRED} \\
    \midrule
    RIFRE & 35.93$\pm$0.53 & \textbf{72.88$\pm$0.5} & \textbf{90.16$\pm$0.41} & 0.72$\pm$0.28 & \textbf{64.27$\pm$0.39} & - & - \\
    SPN4RE & 36.67$\pm$0.12 & 72.66$\pm$0.19 & 88.38$\pm$0.21 & 31.0$\pm$0.79 & 60.56$\pm$0.68 & \textbf{19.81$\pm$0.83} & \textbf{20.66$\pm$1.39} \\
    TDEER & 33.98$\pm$0.62 & 71.86$\pm$0.25 & 70.81$\pm$16.41 & 31.88$\pm$4.04 & 58.79$\pm$1.04 & 15.1$\pm$0.8 & 16.46$\pm$0.84 \\
    UniRel & \textbf{39.53$\pm$0.76} & 67.72$\pm$1.61 & 89.75$\pm$0.27 & \textbf{43.5$\pm$0.3} & 61.56$\pm$0.68 & - & - \\
    \bottomrule
    \end{tabular}%
  \label{aptab:jointPERF}%
\end{table*}%
}

\newcommand{\FewRELPLM}{
\begin{table*}[htbp]
  \centering
  \caption{Examples of mispredicted relations that were completely classified as the mispredicted relation by the PLMs in the few-shot setting}
    \begin{tabular}{ll}
    \toprule
    \multicolumn{1}{c}{\textbf{True Relation}} & \multicolumn{1}{c}{\textbf{Mispredicted Relation}} \\
    \midrule
    /film/film\_location/featured\_in\_films & /people/place\_of\_interment/interred\_here \\
    /location/br\_state/capital & /people/person/place\_lived \\
    /time/event/locations & /people/person/children \\
    EISSN\_number & ground \\
    LCCN\_number & architecturalStyle \\
    academicDiscipline & ground \\
    administrativeCounty & has to its north \\
    areaCode & OCLC\_number \\
    bedCount & has to its northeast \\
    category & editor \\
    chancellor & architecturalStyle \\
    chief & leader \\
    countySeat & ground \\
    class & order \\
    dedicatedTo & has to its north \\
    \bottomrule
    \end{tabular}%
  \label{aptab:fewrel_plm}%
\end{table*}%
}

\newcommand{\FewRELPrompt}{
\begin{table*}[htbp]
  \centering
  \caption{Examples of mispredicted relations that were completely classified as the mispredicted relation by the Prompt-based methods in the few-shot setting}
    \begin{tabular}{ll}
    \toprule
    \multicolumn{1}{c}{\textbf{True Relation}} & \multicolumn{1}{c}{\textbf{Mispredicted Relation}} \\
    \midrule
    /business/company/advisors & /people/place\_of\_interment/interred\_here \\
    /film/film\_location/featured\_in\_films & /film/film/featured\_film\_locations \\
    /location/br\_state/capital & /people/place\_of\_interment/interred\_here \\
    /people/deceased\_person/place\_of\_burial & /people/place\_of\_interment/interred\_here \\
    /people/person/religion & /people/place\_of\_interment/interred\_here \\
    /people/place\_of\_interment/interred\_here & /people/place\_of\_interment/interred\_here \\
    /time/event/locations & /people/place\_of\_interment/interred\_here \\
    EISSN\_number & leader \\
    administrativeCounty & isPartOf \\
    category & neighboringMunicipality \\
    compete in & academicStaffSize \\
    has to its southeast & neighboringMunicipality \\
    headquarter & 1st\_runway\_Number \\
    locationCity & 1st\_runway\_Number \\
    hometown & architect \\
    \bottomrule
    \end{tabular}%
  \label{aptab:fewrel_prompt}%
\end{table*}%
}

\newcommand{\FewRELllm}{
\begin{table*}[htbp]
  \centering
  \caption{Examples of mispredicted relations that were completely classified as the mispredicted relation by the LLMs in the few-shot setting}
    \begin{tabular}{ll}
    \toprule
    \multicolumn{1}{c}{\textbf{True Relation}} & \multicolumn{1}{c}{\textbf{Mispredicted Relation}} \\
    \midrule
    administrativeCounty & part \\
    broadcastedBy & state \\
    category & state \\
    chancellor & title \\
    compete in & part \\
    currentTenants & state \\
    district & part \\
    editor & title \\
    firstPublicationYear & year \\
    foundationPlace & creator \\
    governingBody & city \\
    has to its north & part \\
    has to its northwest & state \\
    governingBody &	city \\
    \bottomrule
    \end{tabular}%
  \label{aptab:fewrel_llm}%
\end{table*}%
}

\newcommand{\GPTRE}{
\begin{table*}[htbp]
  \centering
  \caption{1-shot input prompt examples used for the GPTRE algorithm}
    \begin{tabular}{p{5em}p{40em}}
    \toprule
    \textbf{Type} & \textbf{Prompt} \\
    \midrule
    With NA &  "messages": [{"role": "user", "content": "I'm a knowledgeable person. I will solve the relation extraction (RE) task. Given the context, I'll output the most precise relation between two entities. If there is no relation between them, I will output NONE\textbackslash{}n\textbackslash{}n\textbackslash{}nContext: Nails grow out of deep folds in the skin of the fingers and toes\textbackslash{}nGiven the context, the relation between skin and fingers is COMPONENT AND WHOLE. It is because:\textbackslash{}nThe clues that lead to the relation between \textbackslash{}"skin\textbackslash{}" and \textbackslash{}"fingers\textbackslash{}" being component and whole in the sentence are:\textbackslash{}n\textbackslash{}n1. The phrase \textbackslash{}"in the skin of the fingers and toes\textbackslash{}" suggests that the skin is a larger entity that contains or surrounds the fingers and toes. This indicates that the skin is the component and the fingers are part of the whole.\textbackslash{}n\textbackslash{}n2. The mention of \textbackslash{}"deep folds in the skin\textbackslash{}" implies that the skin has specific features or characteristics that are related to the fingers. This further supports the idea that the skin is the component and the fingers are part of the whole.\textbackslash{}n\textbackslash{}n3. The fact that the sentence specifically mentions the relationship between the skin and the fingers, rather than just mentioning them separately, suggests that there is a close connection between the two and that they are part of a larger whole.\textbackslash{}n\textbackslash{}nContext: The nails of the right hand are easily cut with scissors made for cutting the nails of the left hand\textbackslash{}nGiven the context, the relation between nails and hand is"}]  \\
    \midrule
    Without NA & "messages": [{"role": "user", "content": "I'm a knowledgeable person. I will solve the relation extraction (RE) task. Given the context, I'll output the most precise relation between two entities based on the context\textbackslash{}n\textbackslash{}nContext: you can't walk in peace in the city of dakar anymore , '' said aminata diaw, a history professor here at the university cheikh anta diop , who has studied the culture of charity and begging in senegal .\textbackslash{}nGiven the context, the relation between senegal and dakar is COUNTRY AND CAPITAL. It is because:\textbackslash{}nThe clues that lead to the relation between \textbackslash{}"Senegal\textbackslash{}" and \textbackslash{}"Dakar\textbackslash{}" being country and capital in the sentence are:\textbackslash{}n\textbackslash{}n1. The mention of \textbackslash{}"the city of Dakar\textbackslash{}" implies that Dakar is a city within a larger geographical entity, which is likely a country.\textbackslash{}n2. The reference to Aminata Diaw, a history professor at the University Cheikh Anta Diop, who has studied the culture of charity and begging in Senegal, suggests that Senegal is a country where the University is located.\textbackslash{}n3. The use of the word \textbackslash{}"here\textbackslash{}" in the sentence indicates that the speaker is currently in Senegal, where the University Cheikh Anta Diop is located.\textbackslash{}n4. The fact that Dakar is mentioned in relation to Senegal suggests that Dakar is the capital city of Senegal.\textbackslash{}n\textbackslash{}nContext: what was unique about semb\textbackslash{}u00e8ne was he began to challenge the dominant figure , senghor , '' recalled manthia diawara , a professor of africana studies at new york university who grew up in mali in the 1960s . ''\textbackslash{}nGiven the context, the relation between manthia diawara and mali is"}] \\
    \bottomrule
    \end{tabular}%
  \label{aptab:gptre}%
\end{table*}%
}

\newcommand{\UNLEASH}{
\begin{table*}[htbp]
  \centering
  \caption{1-shot input prompt examples used for the Unleash-LLM algorithm}
    \begin{tabular}{p{50em}}
    \toprule
    \textbf{Prompt} \\
    \midrule
    "messages": [{"role": "user", "content": "Given a context, a pair of head and tail entities in the context, decide the relationship between the head and tail entities from candidate relations: part of, usage, type of, physical, role, related to, origin, compare, named, artifact, general affiliation, topic, win or defeat, temporal, cause effect, social, opposite.\textbackslash{}nContext: Over the past decade , PCNNs have been used in a variety of image processing applications , including : image segmentation , feature generation , face extraction , motion detection , region growing , and noise reduction . The relation between o 'motion detection' and o 'image processing' in the context is part of.\textbackslash{}nContext: Spermidine synthase uses putrescine and S-Adenosylmethioninamine ( decarboxylated S-Adenosyl methionine ) to produce spermidine . The relation between o 'Spermidine synthase' and o 'spermidine' in the context is cause effect.\textbackslash{}nContext: LeRoy Pope Walker of Alabama was made Secretary of War , after being recommended for this post by Clement Claiborne Clay and William Lowndes Yancey ( both of whom declined to accept cabinet positions themselves ) . The relation between person 'LeRoy Pope Walker' and gpe 'Alabama' in the context is physical.\textbackslash{}nContext: China says Taiwan spoils atmosphere for talks . The relation between gpe 'China' and gpe 'Taiwan' in the context is opposite.\textbackslash{}nContext: These albums spawned some of Carey 's most successful singles , including Hero , Without You , All I Want for Christmas Is You , Fantasy , Always Be My Baby , as well as One Sweet Day , which peaked at number one in the U.S. for 16 weeks and became Billboard s Song Of The Decade ( 1990s Decade ) . The relation between work of art 'Always Be My Baby' and person 'Carey' in the context is artifact.\textbackslash{}nContext: Since then there has been a renaissance in Sacred Harp singing , with annual conventions popping up in United States and in a number of European countries recently , including the United Kingdom , Germany , Ireland and Poland , as well as in Australia . The relation between gpe 'United Kingdom' and norp 'European countries' in the context is type of.\textbackslash{}nContext: On 4 October 2009 , George Papandreou , president of the Panhellenic Socialist Movement party and son and grandson of Prime Ministers , 2009 Greek legislative election as the new Prime Minister of Greece , following five years of government under New Democracy leader Kostas Karamanlis , the nephew of long - time Prime Minister and President Konstantinos Karamanlis . The relation between person 'George Papandreou' and org 'Panhellenic Socialist Movement' in the context is role.\textbackslash{}nContext: CycL in computer science and artificial intelligence is an ontology language used by Doug Lenat 's Cyc artificial project . The relation between org 'Cyc artificial project' and o 'CycL' in the context is usage.\textbackslash{}nContext: In 1831 , Michael Faraday made the seminal observation that time - varying magnetic fields could induce electric currents and then , in 1864 , James Clerk Maxwell published his famous paper A Dynamical Theory of the Electromagnetic Field . Maxwell 1864 5 , page 499 ; also David J. Griffiths ( 1999 ) , Introduction to electrodynamics , third Edition , ed . The relation between date '1864 , James' and work of art 'Field' in the context is named.\textbackslash{}nContext: Under the influence of Adrian Maniu , the adolescent Tzara became interested in Symbolism and co-founded the magazine Simbolul with Ion Vinea ( with whom he also wrote experimental poetry ) and painter Marcel Janco . The relation between person 'Tzara' and o 'experimental poetry' in the context is general affiliation.\textbackslash{}nContext: One of the metrics used in NIST ' s annual Document Understanding Conferences , in which research groups submit their systems for both summarization and translation tasks , is the ROUGE metric ( Recall - Oriented Understudy for Gisting Evaluation , In Advances of Neural Information Processing Systems ( NIPS ) , Montreal , Canada , December - 2014 . The relation between o 'ROUGE metric' and org 'NIST ' s annual Document Understanding Conferences' in the context is temporal.\textbackslash{}nContext: Yasser Arafat will meet Shimon Peres in Gaza on Thursday after Palestinians said the right - wing Israeli government had barred the Palestinian leader from flying to the West Bank for talks with the former prime minister . The relation between person 'Yasser Arafat' and person 'Shimon Peres' in the context is social.\textbackslash{}nContext: At the 47th Berlin International Film Festival in 1997 , DiCaprio won the Silver Bear for Best Actor and Luhrmann won the Alfred Bauer Prize . The relation between person 'Luhrmann' and work of art 'Alfred Bauer Prize' in the context is win or defeat.\textbackslash{}nContext: As for the results , the C-HOG and R-HOG block descriptors perform comparably , with the C-HOG descriptors maintaining a slight advantage in the detection miss rate at fixed FALSE positive rate s across both data sets . The relation between org 'C-HOG' and o 'R-HOG' in the context is compare.\textbackslash{}nContext: The period around World War II also saw the publication of the time travel novel Lest Darkness Fall by L. Sprague de Camp , in which an American academic travels to Italy at the time of the Byzantine invasion of the Ostrogoths . The relation between work of art 'Lest Darkness Fall' and norp 'American academic' in the context is topic.\textbackslash{}nContext: that is to end the state of hostility , \textbackslash{}" Thursday 's overseas edition of the People 's Daily quoted Tang as saying . The relation between org 'People 's Daily' and person 'Tang' in the context is related to.\textbackslash{}nContext: The Apollo 7 mission is dramatized in the 1998 miniseries From the Earth to the Moon episode We Have Cleared the Tower , with Mark Harmon as Schirra , John Mese as Eisele , Fredric Lehne as Cunningham , and Max Wright as Wendt . The relation between person 'Eisele' and person 'John Mese' in the context is origin.\textbackslash{}nContext: In 1959 , he resigned from the Indian National Congress and founded the Swatantra Party , which fought against the Congress in the 1962 Indian general election , 1967 Indian general election and 1971 Indian general election elections . The relation between org 'Swatantra Party' and date '1962 Indian general election' in the context is "}] \\
    \bottomrule
    \end{tabular}%
  \label{aptab:unleash}%
\end{table*}%
}
\section{Introduction}\label{sec:introduction}
Relation extraction is a critical step in the task of information extraction that aims to draw high-level inferences from textual data. As the name suggests, relation extraction involves extracting relationships between target nouns, called entities, that bring forth the semantic meaning of the text. It finds usage in various Natural Language Processing (NLP) applications. For example, relation extraction is widely used in question-answering systems to connect question entities to target answer entities \cite{chen-etal-2019-uhop}. Such methods are essential for customer support applications like chatbots and digital personal assistants. Similarly, relation extraction is critical for knowledge base (KB) completion \cite{Trisedya2019-by}. KBs are structured repositories that store data as triplets of entities and relationships (entity1, relation, entity2). KBs find applicability in search engines like Google, Yahoo, and Bing, as they provide an efficient solution for storing large amounts of information to answer search queries. Furthermore, relation extraction can be employed in various other domains to extract knowledge regarding interactions between drugs, proteins, genes, and diseases \cite{wang2018clinical} in biomedical and global industry dynamics \cite{khaldi2022s, kaur2023refind} in business. Thus, owing to its wide-scale applicability, a plethora of research exists in the domain.

Modern-day relation extractors heavily employ deep learning architectures such as BERT \cite{Vashishth2018-lu}, ERNIE \cite{zhang-etal-2019-ernie}, and more recently, GPT-3 \cite{brown2020language} for extracting complex entity and relation representations from the input data. Regardless, it has been observed that modern relation extractors perform well in simple scenarios but poorly on complicated data and relation types. For example, issues such as longer input lengths, multiple relations per sample, overlapping entities, fine-grained relationships, and long-tail data distribution pose challenges that most relation extractors cannot handle efficiently. Although a plethora of traditional surveys that comprehensively summarize the literature exists in this field \cite{bach2007review, Asghar2016-me, Kumar2017-db, Pawar2017-wy, Zhang2017-zm, Smirnova2018-vg, Aydar2020-bd,wang2022deep, zhaocomprehensive, detroja2023survey, wang2021relation, detroja2023survey, peng-etal-2020-learning, alt-etal-2020-tacred, tan-etal-2022-revisiting, alt-etal-2020-probing, han-etal-2020-data, bassignana-plank-2022-mean}, they lack a thorough discussion of the issues faced by the relation extractors and their potential causes. Additionally, the perspective of finding drawbacks in the algorithms rather than concentrating on the data-specific reasons for the challenges is favored in the literature. More importantly, most works focus on very few datasets and algorithm combinations, making it difficult to get a holistic view of the attributes that pose complications for relation extractors \cite{peng-etal-2020-learning, alt-etal-2020-tacred, tan-etal-2022-revisiting, alt-etal-2020-probing, han-etal-2020-data, bassignana-plank-2022-mean}. This also entails that the challenges and future directions proposed by such studies may lack credibility. \textit{Consequently, we perform an in-depth data-centric analysis to put the challenges in perspective and provide a navigable guide through the abundance of research in this field for novice readers to find problem areas that require more attention}.

This paper presents the \textit{first} data-centric performance analysis that investigates a diverse set of relation extraction algorithms to highlight the complex data attributes that adversely affect them. Compared to the existing studies discussed above, this work aims to investigate the root cause of the issues faced by the algorithms by focusing on the attributes of the data. The study summarizes the most important relation extraction paradigms in the literature and highlights the key challenges and future directions supported by an exhaustive set of experiments. Finally, the goal here is not to find shortcomings in the algorithms but rather to determine the data characteristics that make it difficult for neural relation extractors to detect relationships efficiently. This enables future research that addresses these challenges to help create better relation extractors. Thus, the contributions of this work are as follows:
\begin{enumerate}
\item Comprehensive algorithm comparison and sample level analysis of 15 neural relation extractors, including the recent LLM-based algorithms on seven large-scale datasets to highlight reasons behind the performance gap in the field.
\item Extensive discussion on open challenges in relation extraction and future directions to resolve them.
\item A software repository of datasets and re-implemented LLM-based algorithms shared for practitioners in industry and academia for reproducibility\footnote{\url{https://aaig.ece.ufl.edu/projects/relation-extraction}}.
\end{enumerate}

The remainder of the paper is structured as follows. Section \ref{sec:background} presents an overview of the field by discussing a general pipeline for relation extraction and the research done in this field. Section \ref{sec:method} gives an overview of the datasets, algorithms, and methodology used to conduct the performance analysis. Also, it details findings from a general performance assessment of the algorithms. Section \ref{sec:analysis} discusses the data-centric performance analysis, which aims to answer where and why modern relation extractors fail. Finally, Section \ref{sec:future} discusses the research gaps and future directions to further the field of relation extraction. We conclude the paper with some closing remarks in Section \ref{sec:conclusion}.
\section{Background \& Related Work}\label{sec:background}
Relation extraction is the task of extracting relationships between entities \cite{Nadeau2007-fq, yadav-bethard-2018-survey} from textual data. For example, the sentence \emph{``Eno Raud was the son of the writer Mart Raud."} depicts the relation \textbf{son} between the entities \textbf{Eno Raud} and \textbf{Mart Raud}. Such relationships provide rich contextual knowledge critical to language understanding. Two types of relations exist in the literature, and a few examples of each category are shown in Table \ref{tab:datatypes}. First, \emph{simple relations} describe the semantic relationship between the two entities. Second, \emph{directional relations} define the semantic relationship and the relation's directionality. For example, the relation \emph{Component-Whole(e1, e2)} represents a relationship between two entities where \emph{entity1} is a component of \emph{entity2}. The relation labeling is done using combinations of automatic and manual labeling strategies. Distant supervision, a technique that automatically labels relations by linking entities in the text to entities present in KBs, has been a widely used labeling strategy in this field and is discussed in section \ref{subsec:distant_background}.
\RELTYPE

The research in relation extraction has been divided into three sub-domains. The first area deals with classifying relationships between pre-defined entities, also known as \emph{relation classification}. The second area deals with extracting entities and relations between them as relation triplets \emph{(entity1, relation, entity2)}, known as \emph{joint relation extraction}. Both algorithms work in a supervised manner wherein the former matches relations with a predefined set of labels, and the latter matches relation triplets to a list of ground truth triplets. It is worth mentioning that a third set known as \emph{distant supervision} algorithms also exists in the literature \cite{Smirnova2018-vg}. Although these algorithms work on the relation classification paradigm, research in this domain has focused solely on dealing with large automatically labeled datasets and the wrong labeling problem that comes with it. The problem arises when creating large datasets; relations and entities are incorrectly labeled by making false links with KBs \cite{Lin2016-al, Ji2017-cm, Feng2018-de}. However, most SOTA algorithms today are based on the former two categories. Figure \ref{fig:taxo} presents a taxonomy of the relation extraction algorithms in the literature. The following sections present a literature overview of the three domains.
\taxo

\subsection{Relation Classification} \label{subsec:rc_background}
Relation classification techniques aim to extract semantic relationships between pre-defined entities present in text. Initial research in relation classification began with feature-based and kernel-based algorithms that used low-level NLP features such as part-of-speech (POS) and dependency parsers or developed specialized kernels for the task. These algorithms employed machine learning technologies \cite{Asghar2016-me} to classify relationships between pairs of target entities. With the advent of neural networks, research started incorporating deep learning architectures for relation classification. However, the literature has followed a uniform pipeline regardless of technology. A typical relation extractor classifies relationships by extracting feature representations of the input data. Figure \ref{fig:pipe} shows a traditional relation classification pipeline used in the literature.
\pipe

The input to a relation classifier consists of relation labels and text samples labeled with entity mentions. The relation labels are converted to numeric integers or verbalized before being used as input by the classifier. The entities in the text sample are annotated through manual labeling or using named entity recognizers such as Spacy\footnote{\url{https://spacy.io/}} and Stanford CoreNLP\footnote{\url{https://stanfordnlp.github.io/CoreNLP/}} \cite{manning2014stanford}. The annotation of the type of entity with its position in the text is also prevalent in the literature. The entity-annotated text sample acts as an input to the classifier. Most machine learning relation extractors extract low-level features from the text sample, such as POS tags, WordNet hypernyms, dependency parsers, and words between entities \cite{Pawar2017-wy}. These features are used as input to the classifier. On the other hand, deep learning relation extractors employ neural network architectures for feature extraction. Neural networks such as convolution neural networks (CNNs) and recurrent neural networks (RNNs) are exceptional in learning complex internal representations from input data. Additionally, long short-term memory networks (LSTMs) that capture long-range dependencies and large language models, such as BERT and RoBERTa, that extract contextual representations of the text sample are employed for extracting meaningful text and entity representations. Subsequently, the extracted features are used to classify text samples into various relational categories using a classification layer. The classification module usually consists of support vector machine (SVM) \cite{bach2007review} for machine learning algorithms and simple multilayer perceptrons (MLP) for deep learning algorithms. The classifier generates output scores for all the relation classes. The final relation label is assigned to the top-scoring class. 

Research in relation classification ranges from machine learning and statistical approaches to more recent neural networks.  Although machine learning algorithms could extract semantic relationships from the data, they required considerable feature engineering and manual labor to create handcrafted rules \cite{Kumar2017-db}. On the other hand, neural networks could extract complex internal representations from the input text, providing a noise-prone alternative to external NLP features employed by previous methods. Thus, the research focus in this field has predominantly shifted to using neural networks for relation classification. As discussed above, language model-based neural approaches for relation classification extracted complex internal representations of the input data using neural network architectures and used a separate classification head to predict the relationships. Subsequently, a new paradigm was introduced with the advancement of generative language models, especially GPT-3, making it possible to employ the language model as a predictor. Various prompting strategies were used to probe the knowledge present within the language model. The following section discusses the research endeavors in traditional and prompt-based categories.

\subsubsection{Traditional Approaches}
Extracting relationships between two entities requires an efficient understanding of relevant information that can be present anywhere in the input sample. Recurrent architectures, such as the LSTM, proved very useful due to their capability of capturing long-range dependencies. Thus, all initial relation classifiers employed LSTM architectures with a fully connected softmax layer for classification. Algorithms such as Attn-BLSTM \cite{Zhou2016-xt} and CRNN-Att \cite{Raj2017-ee} were among the first to provide enhanced solutions to relation extraction by incorporating different strategies of attention mechanisms \cite{Vaswani2017-ol} enabling them to extract essential information present anywhere in the input sample. Yet, they did not account for the position and presence of the entities in the training process. Subsequent research focused on using entity-specific information, such as position features, to aid the classification process \cite{Zhang2017-an, sym11060785}. However, the recurrent architectures struggled with relation classification due to their shorter context window making them prone to forgetting past states. Thus, with the introduction of the transformer architecture \cite{Vaswani2017-ol}, pre-trained language models (PLMs) were heavily employed for this task. 

Popular PLMs such as BERT and RoBERTa \cite{liu2019robertarobustlyoptimizedbert} have performed exceptionally well in various NLP tasks. They are trained on large corpora of data using masked language modeling (MLM) styled training and can be easily fine-tuned on smaller task-specific datasets, providing significant performance gains. Initially, PLM-based algorithms for relation classification aimed to understand contextual information with minimum knowledge of the entities and gradually progressed to more advanced forms of additional knowledge. For example, RBERT \cite{Wu2019-oi} and RoBERTa-large \cite{Zhou2021-da} used contextualized token representations from pre-trained BERT and RoBERTa models and incorporated knowledge about the global position of entities by introducing special tokens at the input level. Subsequently, algorithms started incorporating more advanced forms of knowledge by mapping entities to their KB counterparts \cite{zhang-etal-2019-ernie, peters-etal-2019-knowledge, wang-etal-2021-k, 10.1162/tacl_a_00360}. The introduction of PLM-based algorithms significantly advanced the research in the field due to the capabilities of the networks to extract and connect contextual representations of the entities and the relations. However, such relation classifiers suffered due to the objective gap between the pre-training and fine-tuning strategies. Most PLM-based relation classifiers used MLM to pre-train the language model, while a classification objective was used for fine-tuning. Prompt-based algorithms were introduced to alleviate the reliance on separate classification modules and use the language models as a classifier.

\subsubsection{Prompt-based Approaches}
Prompt-based techniques aim to model the relation classification problem as an auto-regressive text generation task, enabling language models to be predictors rather than feature extractors \cite{le-scao-rush-2021-many}. This ability is achieved using cloze-style (fill-in-the-blanks) prompts or statements to probe the language model to perform various downstream tasks. A typical prompt consists of a template and verbalizations of the label space. The prompt is coupled with the input sentence and made to answer a cloze-style statement. This technique is now widely employed in the NLP literature and has shown great improvements in zero and few-shot learning where training data is scarce. However, adapting the prompting strategy to relation classification has been a challenging task due to the diversity of the large label space. Not only is it difficult to create distinct verbalizations of multiple relations manually, but it is also difficult to fit the large label space in prompts, especially for expensive LLMs such as the GPT variants. Thus, prompt-based approaches for relation classification aimed to alleviate these issues by creating custom prompt-tuning strategies by employing PLMs like BERT, RoBERTa, T5, and BART \cite{han-etal-2020-data, chen2022knowprompt, han-etal-2022-generative}. 

Finally, with the introduction of GPT-3, most research in relation classification has shifted to the use of chat-based LLMs. These algorithms use manually curated prompt templates designed to ask the LLM cloze-style questions. The prompt is oftentimes augmented with selected training demonstrations to carry out few-shot learning \cite{xu-etal-2023-unleash, wadhwa-etal-2023-revisiting}. For few-shot learning, various demonstrations are added to the prompt to guide the LLM to better learn the task specifics. This is called the in-context learning (ICL) framework, and research in this subdomain has concentrated on creating better demonstration retrieval strategies for relation classification \cite{wan2023gpt}. Finally, recent algorithms have further tried to advance the field by combining PLMs and LLMs to create better few-shot extractors \cite{ma2023large}.

\subsection{Distant Supervision}\label{subsec:distant_background}
The distant supervision paradigm was developed to cater to the need for larger annotated datasets by creating them without human intervention. This was accomplished by aligning training corpora with KBs such as Freebase\footnote{\url{https://en.wikipedia.org/wiki/Freebase_(database)}}, DBPedia\footnote{\url{https://www.dbpedia.org/}} and Wikidata\footnote{\url{https://www.wikidata.org/wiki/Wikidata:Main_Page}}. Mintz et al. \cite{Mintz2009-ed} were the first to introduce this paradigm in the context of relation extraction. According to their research, if two entities \emph{(e1,e2)} participate in a relation \emph{X} present in a KB \emph{D}, then all sentences from a text corpus \emph{C} containing \emph{(e1,e2)} will participate in the relation \emph{X}. This way, large amounts of data could be automatically annotated using relations from KBs. Although this technique attracted much attention, it was built on a stringent assumption that \textbf{all} sentences containing \emph{(e1,e2)} must belong to a given relation. This assumption was prone to falter, leading to the wrong labeling of the datasets.

Most research in distant supervision revolved around alleviating the problem of noisy labels by finding efficient ways of selecting sentences associated with specific relations. A prominent work done by Riedel et al. \cite{Riedel2010-yt} relaxed the above assumption and restated the distant supervision paradigm as follows: If two entities \emph{(e1,e2)} belonging to a text corpus \emph{C} participate in a relation \emph{X} present in a KB \emph{D} then out of all the sentences containing \emph{(e1,e2)} from \emph{C},  \textbf{at least one} sentence will participate in the relation \emph{X}. This new assumption significantly helped reduce the number of noisy labels; however, it worked on the assumption that an entity pair could not participate in multiple relations. This assumption faltered when it came to large, distantly supervised datasets. Thus, the multi-instance multi-label (MIML) paradigm for relation extraction was introduced to enable the joint modeling of entities and relations and to combat the problem of overlapping relations\cite{Hoffmann2011-qj, Surdeanu2012-sl}. The MIML paradigm emphasized the need and benefits of extracting relations and entities in a single pipeline. Consequently, the joint relation extraction paradigm was developed.

\subsection{Joint Relation Extraction Algorithms} \label{subsec:joint_background}

 In the literature discussed above, the knowledge of entities was obtained before the relation classification process. However, separate entity extraction strategies can lead to issues like induction of noise, discrepancies in feature space, and loss of critical information about the entities. Therefore, joint relation extraction algorithms were developed to unify the entity and relation extraction pipelines. Initial work in joint modeling employed handcrafted features such as integer linear programming and card-pyramid parsing \cite{Pawar2017-wy, Miwa2016-mm}. The models learned entity representations and used them to extract relations. Model parameters were updated jointly using both entity and relation labels. Subsequently, novel techniques were introduced, such as employing neural networks to better learn global features with little use of syntactic grammar \cite{Zhang2017-os}. These methods worked on a table-filling framework and incrementally extracted entities and relations with joint parameter updates. However, the entity and relation extraction pipelines were still separate and created redundant information \cite{Zheng2017-kk}. The triplet extraction paradigm was introduced to unify the entity and relation extraction pipelines further. It extracted relation triplets in the form of \emph{(entity1, relation, entity2)} in one go rather than extracting entities and relations consecutively. Novel tagging mechanisms were introduced to extract relation triplets from text \cite{Zheng2017-kk, Wei2019-wh}. These algorithms were designed as a sequence labeling task and needed humans to design complex tagging schemes \cite{10103602}. 

One of the primary challenges for the early joint relation extractors was the problem of multiple relations and overlapping entities. The assumption that two entities participate in a single relation in a given sentence is rarely satisfied in real-world data. Relationships can occur in various forms in natural text. For example, the sentence \emph{``Dominican Republic signer and songwriter Juan Luis Guerra led the Latin Grammy Awards nominations followed by Puerto Rico's Ricky Martin"} depicts the relation \textbf{city of birth} between the entities \textbf{Dominican Republic} and \textbf{Juan Luis Guerra}. However, the same relationship exists between \textbf{Puerto Rico} and \textbf{Ricky Martin}. A sentence can exhibit multiple relations by sharing the same or different entities \cite{Zeng2018-cy}. Figure \ref{fig:overlap} shows a few more examples of this complex scenario. Entity pair overlap (EPO) and single entity overlap (SEO) are two instances of this problem where multiple relations are shared between both or a single entity. Thus, subsequent research in this field was done to develop relation extractors robust to this issue. Some methods to solve this problem included creating better representation schema for instances with overlap \cite{Nayak2019-mi}, incorporating relational knowledge as apriori to avoid extracting irrelevant entities \cite{Zhou2021-da}, decomposition-based methods \cite{li-etal-2021-tdeer, Zhao2021-kv} and using the seq2seq paradigm to circumvent the need for triplet extraction in a particular order \cite{10103602}. 
\overlap

The triplet extraction paradigm allowed for seamless information sharing of entity and relation types, which helped tackle the problem of entity overlap. However, such models suffered from exposure bias due to the discrepancies in the context provided during training and inference \cite{wang-etal-2020-tplinker}. Thus, subsequent solutions aimed at further unifying the extraction process by introducing various strategies such as token-pair linking and entity-relation interaction modeling \cite{wang-etal-2020-tplinker, tang-etal-2022-unirel}.

This section highlighted the plethora of research endeavors in relation extraction literature. The efforts can be categorized into classification and extraction strategies that aim to classify relationships from a pre-defined label set or extract entity and relation triplets, respectively. The algorithms introduced have progressively advanced over time, each employing a significantly more advanced deep learning backbone than the previous one. The most recent one was the introduction of relation extraction algorithms using OpenAI's GPT-3 architecture, which has shown notable performance in low-resource scenarios. These research efforts have imparted significant gains to the field of relation extraction. However, there is still a considerable performance gap in the field where relation extractors have not reached their full potential. This study hypothesizes that the disparity comes from input data characteristics that pose complicated use cases for relation extractors to tackle. Thus, the study highlights some of these complex data characteristics by looking at a diverse group of relation extraction algorithms and critically examining their effectiveness in extracting relationships from complex textual data. To this end, the next section discusses the methodology used for the data-centric performance analysis and some preliminary insights from it.
\section{Methodology} \label{sec:method}
This study aims to gain insights into the decision-making process of neural algorithms and highlight challenging data characteristics that potentially inhibit the performance of relation extractors. Thus, exhaustive experiments encompassing 15 relation extraction algorithms and seven datasets were conducted. This section details the methodology used to conduct the analysis and the insights gained from it.

\subsection{Datasets \& Complex Data Characteristics}
Seven SOTA datasets from the relation extraction literature were used for this study. The datasets were chosen based on their complex data characteristics. These characteristics deal with the challenging aspects of real-world data that adversely affect the performance of relation extractors. Therefore, it is critical to analyze the performance of the algorithms under such complicated scenarios. To this end, this analysis categorizes the datasets according to the data complexities they exhibit. 
\DATASETS

Each selected dataset represents one or more characteristics that challenge the accurate extraction of relations by neural relation extractors. These characteristics include fine-grained relations, long-text samples, multiple relations associated with a sample and overlapping entities, and long-tail data distribution. The datasets were categorized based on these categories. Furthermore, their counter-categories were also created as a baseline for comparison. The details of these categorizations are presented below.
 \begin{itemize} 
 \setlength\itemsep{1mm}
     \item \textbf{Fine-grained vs. Coarse relations.} Fine-grained relation distribution deals with the presence of numerous relations in a dataset. Such a distribution leads to the relation labels having similar meanings and are usually difficult to distinguish from each other. For example, the WebNLG dataset has more than 100 relation labels. The label space includes relationships such as ``chief'' and ``leaderName'', which can easily be confused together without a sound understanding of the associated context. Efficient handling of fine-grained datasets is critical for extracting relations from real-world data where numerous relationships can be present, and fine-grained distinction is necessary for discriminating between similar relation types. Thus, datasets with 40 or more relation labels were categorized as \emph{fine-grained} relation datasets and datasets with less than 40 relations as \emph{coarse} relation datasets. 
     
     \item \textbf{Long vs. short input.} The number of words or length of the input text sample has been a source of disparity in all NLP tasks. Both long and short inputs can provide complex use cases that might impede the performance of relation extraction algorithms. On the one hand, the short input text is challenging to tackle due to its succinct nature and insufficient knowledge. For example, it is difficult to interpret the relation \emph{entity-origin} between entities \emph{beer} and \emph{barley} from the sentence \emph{``space beer is made from barley grown in space''}. On the other hand, long input lengths might cause issues due to conflicting contextual information and the need for a larger context window. Efficient handling of short-input samples can prove to be beneficial when extracting relations from social media platforms like Twitter, where the text lengths are restricted. Similarly, extracting relations from long-input text can be critical to summarizing lengthy text documents abundant in biomedical, finance, and academic fields. Thus, datasets with an average sample length greater than 35 tokens were categorized as \emph{long-input} datasets. Similarly, datasets with an average sample length of less than and equal to 35 tokens were categorized as \emph{short} input datasets. 
     
     \item \textbf{Multiple relations and overlapping entities vs. one-to-one associations.} As discussed in Section \ref{subsec:joint_background}, text samples incorporating entities that share multiple relation labels are a big concern in the relation extraction literature. The presence of these complex attributes leads to the possibility of multiple correct predictions for a text sample. Efficient extraction of relations shared among the same entities can make algorithms robust to intricate data from business and finance domains, which are abundant with one-to-many associations. Thus, the datasets were categorized into \emph{multiple/overlap} and \emph{one-to-one} categories based on the presence or absence of this complex characteristic.
     
     \item \textbf{Long-tail vs. uniform distribution.} The long-tail data distribution deals with the severely skewed distribution of representative input data for different relations in a dataset. For example, the WebNLG dataset consists of relation classes with only one sample. Such distribution has been known to adversely impact the performance of relation extractors. Efficient handling of long-tail distribution is critical for low-resource applications in NLP, such as clinical information extraction and personality prediction. Thus, the datasets were categorized into \emph{long-tail} and \emph{uniform} categories based on the distribution of the data.  

 \end{itemize}

The categorization of each selected dataset can be found in Table \ref{tab:procon}. These datasets provide a wide range of variability of complicated characteristics that a neural relation extractor must mitigate and are worth investigating. To this end, the following section details the selected algorithms used for this analysis.

\subsection{Algorithms}
SOTA relation classification and extraction algorithms were implemented from the literature to conduct the data-centric performance analysis presented in this study. The algorithms chosen represent the major paradigms present in the domain of neural relation extraction. Thus, they provide a comprehensive view of the literature in this field. The following sections give brief details of the methodology used by these algorithms.

\subsubsection{Relation Classification Algorithms}
The relation classification experiments were conducted using algorithms ranging from recurrent architectures to modern LLMs. Although LLMs provide significant performance gains for relation classification, it is essential to analyze recurrent algorithms to get a holistic view of the field. Furthermore, existing research only showcases the performance of these algorithms on a select few datasets. Hence, a comprehensive analysis of the recurrent algorithms for their generalizability and robustness to complex data characteristics might lead to novel and crucial insights into the field. Thus, this work utilizes three recurrent algorithms. 
\begin{itemize}
    \item \textbf{Att-BLSTM} \cite{Zhou2016-xt}, a single-layer bidirectional long short-term memory (BLSTM) network with an attention mechanism to extract the most important information present in the input sample. 
    \item \textbf{PAWARE} \cite{Zhang2017-an} utilizes entity information by augmenting each token with its relative position to the entities. The algorithm uses a position-aware attention mechanism on the output of a 2-layer LSTM network to create a fine-grained representation of the input. 
    \item \textbf{Entity-BLSTM} \cite{sym11060785}, uses a single-layer BLSTM network with self- and entity-aware attention mechanisms. It incorporates entity-type information using latent topic clustering along with the position features used by PAWARE.
\end{itemize}

\noindent A crucial area with significant research has been the use of PLMs for relation classification. These algorithms aim to fine-tune SOTA PLMs by introducing various learning strategies. The high-performing algorithms that have been implemented for this work are discussed below.
\begin{itemize}
    \item \textbf{RBERT} \cite{Wu2019-oi} uses contextualized token representations and sentence encodings from a pre-trained BERT model. It incorporates knowledge about the global position of entities by introducing special tokens at the input level.
    \item \textbf{Roberta\_base} \cite{Zhou2021-da} presents a RoBERTa-based baseline for relation classification. It introduces various methods of adding entity-related information to the input sentence, such as entity mention and type markers.
    \item \textbf{LUKE} \cite{Yamada2020-lu} treats entities as separate tokens and uses their contextualized representation with the input tokens. The model used a RoBERTa-base transformer and applied custom pre-training strategies based on entity masking to achieve the relation extraction objective.
    \item \textbf{ERNIE} \cite{zhang-etal-2019-ernie} incorporates information about entities by mapping them to their KB counterparts. It uses a dual encoder strategy in which the first encoder (BERT) creates holistic input representations, and the second decoder injects external knowledge from the KBs.
\end{itemize}

\noindent With the popularity of decoder-based algorithms, relation classification was transformed into a language generation task using various prompt-based algorithms. These algorithms target both supervised and few-shot learning paradigms. Two SOTA prompt-tuning algorithms have been explored for this study.
\begin{itemize}
    \item \textbf{KnowPrompt} \cite{chen2022knowprompt} incorporates knowledge contained in relation labels by using virtual answer words. The representation of these words is optimized using a RoBERTa model by calibrating them with respect to the context words.
    \item \textbf{GenPT} \cite{han-etal-2022-generative} reformulates relation classification as a text-infilling task compared to the masked language modeling style used by other prompt tuning methods to eliminate rigid prompt restrictions. The algorithm uses an entity-guided decoding strategy to align the generated sequences with the predefined labels, making the prediction process more effective. The algorithm experiments with Bart \cite{lewis2019bart}, T5 \cite{raffel2020exploring}, and RoBERTa-base PLMs, with RoBERTa achieving the highest performance.
\end{itemize}

\noindent Finally, two LLM-based algorithms were utilized for this study. These algorithms target the few-shot learning paradigm and employ chat-based LLMs. It must be noted that the SOTA algorithms in the relation classification literature are still constrained to the use of GPT-based models. Thus, both the selected algorithms use GPT-3.5 architectures.
\begin{itemize}
    \item \textbf{GPT-RE} \cite{wan-etal-2023-gpt} utilizes sentence similarity (simCSE \cite{gao-etal-2021-simcse}) with a KNN-based demonstration retrieval strategy to perform in-context learning with a GPT-based architecture. The study also introduces the use of relation representations from PLMs, such as BERT, that are finetuned for the relation extraction task to extract relevant demonstrations. However, due to the unavailability of the source code, only the sentence similarity paradigm was used in this study. 
    \item \textbf{UnleashLLM} \cite{xu-etal-2023-unleash} exploits GPT-3.5 for relation classification by creating prompts with task-related instructions with the complete label list. The entity types are also embedded within the demonstrations. The study also proposed a data augmentation strategy, which has not been implemented for this study due to resource constraints.
\end{itemize}

\subsubsection{Joint Relation Extraction Algorithms}
To analyze the performance of the joint relation extraction paradigm, four SOTA encoder and decoder-based algorithms were utilized for this study. It was found that the few studies that used prompting and LLMs for this task did not have publicly available source code and sufficient instructions for reproducibility. Thus, prompting-based joint relation extractors have not been investigated in this study. The algorithms in this category have been discussed below.
\begin{itemize}
    \item \textbf{SPN4RE} \cite{10103602}, an improved decoder-based algorithm that eliminates the need for ordered triplet extraction using a non-autoregressive decoder. The decoder is bidirectional and is composed of stacked identical transformer layers. The algorithm also introduces a bipartite loss function invariant to any permutation of predictions.
    \item \textbf{TDEER} \cite{li-etal-2021-tdeer} is a decomposition-based algorithm that uses a translating layer to map the generated sequence of labels to the final output, considering the dependencies between entities and relations. A binary classifier predicts all entities' start and end positions, followed by a multi-label classifier to extract multiple relations associated with an input sample. Additionally, it incorporates a negative sampling strategy to address the accumulation of errors at various stages. 
    \item \textbf{RIFRE} \cite{Zhao2021-kv} uses a decomposition strategy, which is enhanced by making use of a graph neural network (GNN) and adding knowledge of the relations before extraction.
    \item \textbf{UniRel} \cite{tang-etal-2022-unirel} uses a BERT-base encoder to encode contextual representations of entities, relations, and the input sentence. It models entity-entity and entity-relation interactions using the self-attention mechanism of a transformer.
\end{itemize}
\ALGOS

  Table \ref{tab:algorithms} details a summary of the selected algorithms. It is worth mentioning that distant supervision algorithms were not included in the performance comparison. Due to their task-specific nature, it was found that these algorithms were extremely difficult to generalize to other datasets. Most algorithms require entity tags to link entities to different KBs. Unfortunately, this information was not readily available for most of the datasets used in this performance analysis. Therefore, these models were omitted from this analysis. The next section discusses the experimental methodology for training and testing the selected algorithms.

\subsection{Experimental Methodology} \label{method}
This performance analysis aims to answer where, how, and why neural models fail to extract relations from textual data. For this, 15 neural relation extraction algorithms were selected to extract relations from the above-mentioned datasets. The algorithms were re-implemented using open-source codes available on GitHub\footnote{\url{https://github.com/}}. Algorithms such as PAWARE, Roberta\_base, and UnleashLLM required entity type information, which was only available for TACRED and RETACRED datasets. Thus, the NER tool from Stanford CoreNLP\footnote{\url{https://stanfordnlp.github.io/CoreNLP/}} \cite{manning2014stanford} was used to annotate the other datasets with the entity type information. However, even this endeavor could not categorize all entities with their corresponding types, and many entities were annotated with the unknown (`O') entity type. The statistics of this entity annotation process are available in Table \ref{aptab:ent_stats} in the Appendix. Also, the annotated datasets have been released with the source code. Finally, adjustments were made to the algorithms to accommodate for the incomplete entity-type information. Detailed implementation details of the affected algorithms can be found in Section \ref{app:method} in the Appendix. The subsequent methodology can be divided into two categories: 1) Supervised and 2) Few-shot learning strategies.

\subsubsection{Supervised Strategy}
For supervised learning, the relation classification algorithms were categorized into \textbf{Recurrent}, \textbf{PLM-based}, and \textbf{Prompt-PLM} approaches, and all joint extractors were evaluated individually. These algorithms were trained in the supervised setting, where the complete training and development sets were utilized during fine-tuning of these models. The training was done using 5-fold cross-validation with hold-out testing. Most datasets were used as-is, but FewRel was restructured to make it compatible with the supervised pipeline used for this analysis. Specifically, 100 instances from each class were sampled for the training set and 200 for the validation and test set \cite{zhang-etal-2019-ernie}. Also, for joint relation extractions, the duplicate samples in the datasets were grouped together, and their corresponding relations were restructured in the form of entity and relation triplets. 

For evaluation, micro-averaged F1 scores without the contribution of the negative class were used to analyze both relation classification and joint relation extraction algorithms based on the trend in the literature. Also, the ``exact'' match criterion was used to match the predicted triplets with the ground truth triplets \cite{taille-etal-2020-lets} for the joint extractors. The predicted triplets were matched using both head and tail tokens of the entities and the relation between them.

\subsubsection{Few-shot Strategy}
Similarly, more recent algorithms were utilized for few-shot learning and categorized into prompt and LLM-based approaches. The few-shot strategy deals with training algorithms using a small subset of the original training data in a low-resource setting. They have been tagged as \textbf{Prompt-PLM} and \textbf{Prompt-LLM}. Both PLM and LLM-based algorithms utilizing prompts were trained using this strategy. The GenPT algorithm was trained using Bart, T5, and RoBERTa. However, only the RoBERTa models were used for the data-centric analysis as they consistently performed well on most datasets. To create a basis for comparison with the traditional approaches, algorithms RBERT and RoBERTa\_base (some of the best algorithms from the supervised category) were also trained using this setting and were categorized as \textbf{PLM-based} algorithms. 

For this study, 1, 5, 10, 20, and 30-shot datasets were created by random sampling from each class. The sampling used three seed values (13, 42, and 100). Thus, experiments were conducted on the three iterations of the datasets. The SemEval-2010 Task 8 dataset was restructured to make it compatible with the LLM-based algorithms. As per the existing literature, the directionality of the relations was removed from the relation labels by grouping samples based on the base simple relations. The statistics of this new dataset can be found in Table \ref{aptab:fewData} in the Appendix. Finally, due to the high cost of running GPT-based algorithms, smaller subsets of the original test sets were also sampled using the same seed values. Similar to the supervised algorithms, the micro-F1 score was used to evaluate the performances of the algorithms for different shots of data. The algorithms were analyzed without incorporating the contribution of the negative class.

\noindent Details of the distribution of the algorithms with respect to the training strategies can be found in Table \ref{tab:algorithms}. The next section highlights the findings from a general evaluation of the algorithms on the selected datasets.

\subsection{Results}\label{sec:results}

The trained algorithms were made to predict relationships from the test samples of the selected datasets. Figures \ref{fig:sup_all}, \ref{fig:joint_all} and \ref{fig:few_all} depict the average performance and standard deviation for each algorithm dataset pair across five cross-validated folds (supervised relation classification and joint relation extraction) and three seed iterations (few-shot learning). Detailed performance scores for the figures can be found in Tables \ref{aptab:superPERF}, \ref{aptab:jointPERF} and \ref{aptab:fewPERF} in the Appendix. This section highlights some findings from analyzing the performance of the algorithms to show the validation of the conducted experiments and highlight any intersections or deviations from trends in the literature. To this end, a critical evaluation of the relation classification and joint extraction algorithms has been discussed below.

\subsubsection{Algorithms under fully-supervised setting}

First, it can be inferred from Figure \ref{fig:sup_all} that both PLM-based and Prompt-PLM algorithms were better at extracting relationships in the supervised setting than the recurrent algorithms. This finding matches the general trend in the literature where PLMs have shown significant performance gains at accurately extracting relationships from text. A probable reason for this observation can be due to the ability of the PLMs to understand the contextual information present in a sentence and come up with a comprehensive representation. However, it is worth mentioning that the performance of the recurrent architectures was not significantly behind that of the PLMs. This finding highlights the performance gap in the field, where even with highly advanced PLMs, not much performance gain was observed in the overall task of relation classification.

Next, the joint relation extractors were investigated based on the performances depicted in Figure \ref{fig:joint_all}. It was found that all joint relation extractors were extremely brittle and did not consistently perform across the datasets selected for this study. This finding can be observed by the high variation of performances, with most extractors performing the best on the WebNLG dataset and worst on the TACRED dataset. The existing literature on joint RE has only evaluated the algorithms on a few datasets. The observations from the experiments conducted in this study showcase that a significant performance gap exists in this domain, and it is essential to make joint extractors more generalizable.

\superAll
\jointAll

\subsubsection{Algorithms under low-resource setting}

Next, in low resource settings, it can be inferred from Figure \ref{fig:few_all} that the Prompt-PLM algorithms consistently performed well across the k-shots in the traditional relation classification category. Two major inferences can be drawn from this finding. First, traditional PLM-based algorithms like RBERT and Roberta\_base are inefficient at extracting relationships from textual data in low-resource scenarios, as seen by the low comparative performances of these algorithms. Second, converting the relation extraction objective to a language generation objective has significant benefits that best come into play without considerable training data. 

Finally, the performances of the LLM-based algorithms were found to be highly volatile and depended on the properties of the target datasets. Although the algorithms UnleashLLM and GPT-RE showed performance gains for some dataset combinations, especially in the 1-shot setting, making these algorithms compatible with the task was difficult. First, the performance scores depicted in Figure \ref{fig:few_all} for the LLM-based algorithms had to be calculated from a smaller subset of the test data because of the tendency of the LLMs to generate out-of-domain predictions that were not a part of the original label space. Such samples had to be discarded from the performance calculation. This issue is evident by the extremely low performance on the WebNLG dataset, which has a complicated and large label space. Second, it was impossible to fit the full label space into the input prompt, especially for the larger dataset. Thus, the UnleashLLM algorithm could not be evaluated in more than 1-shot setting. This observation highlights the inefficiency of the LLM-based algorithms for relation classification, where it is common to have complex and large label spaces.
\fewAll

\hfill \break
\noindent Based on the findings discussed above, it can be inferred that there exist niche issues that hamper the performance of most relation extraction algorithms, even in the presence of SOTA architectures. Being the counterpart to algorithms, this naturally elucidates the necessity for a data-centric performance assessment to find possible attributes in the data that may hinder the performance of the relation extractors. The following section pursues this direction.
\section{A Data-centric Performance Comparison}\label{sec:analysis}
This work investigates the attributes in the data that adversely affect the relation extraction capabilities of neural algorithms. To facilitate this objective, two types of analysis were conducted on the selected algorithm and datasets discussed in the previous section. First, a quantitative analysis was performed to critically evaluate the performance of the selected algorithms under various complex data characteristics. Second, a qualitative error analysis was conducted to reach the root cause of the incorrect predictions observed in the algorithms.

\subsection{Quantitative Analysis}
This analysis was conducted to find connections between model performances and complex data characteristics. For this, the performance of the algorithms was observed for the complex and counter-complex categories discussed in Section \ref{sec:method}. Thus, the algorithms were analyzed under \textbf{coarse vs. fine-grained relation distribution}, \textbf{short vs. long input}, \textbf{one-to-one vs. multiple-relations/overlap}, and \textbf{uniform vs. long-tail} categories. Figures \ref{fig:superComplex}, \ref{fig:fewComplex}, \ref{fig:superOverlap} and \ref{fig:fewOverlap} show aggregate plots composed by averaging the performance scores of different algorithm-dataset combinations over five cross-validated folds (supervised learning) and three seed iterations (few-shot learning). Similarly, Figures \ref{fig:jreComplex} and \ref{fig:jreMultiOver} depict the average performances of joint relation extraction algorithms across five cross-validated training runs. The following sections discuss the findings from evaluating both relation classification and joint relation extractors with respect to the complex attributes.

\subsubsection{Fine-grained vs. coarse relation distribution}

Studies suggest that the performance of relation classification algorithms suffers when many relation types are present in a dataset (fine-grained relation distribution). Figure \ref{subfig:sup_fine} depicts the performance of relation classification algorithms in the presence of fine-grained and coarse relation distributions under a fully supervised setting. Contrary to popular belief, it was found that PLM and Prompt-PLM algorithms were robust at tackling data with fine-grained relations. Most such algorithms are based on transformer architectures that are efficient at semantic understanding from large corpora of data. Thus, a probable reason for the efficiency of these algorithms can be attributed to this capability, wherein the presence of fine-grained relations leads to a larger label space, making it possible to learn better distinctions in the data. On the other hand, the performance of the recurrent architecture was comparable across the two categories, with a high deviation from the mean when tackling fine-grained relations. This finding indicates that recurrent architectures were sensitive to other data characteristics when handling fine-grained relation distributions.

Similarly, distinct findings were observed with few-shot learning. First, it can be inferred from Figure \ref{subfig:few_fine} that Prompt-LLM algorithms showed superiority in tackling coarse relation distribution compared to PLM and Prompt-PLM algorithms. This finding further corroborates that the latter algorithms require a large sample space to learn relation distinctions accurately. On the other hand, the performance gain of the LLMs can be attributed to the vast amount of knowledge already present within these models, making them efficient at classification when the label space is sparse. However, the lower overall performance of the Prompt-LLMs in the fine-grained category suggests that LLMs tend to get confused when diverse relations are possible. Additionally, PLM and Prompt-PLM algorithms performed better at tackling fine-grained relations due to the above-mentioned reasons. However, the large standard deviations experienced by these algorithms highlight that in low-resource settings, these algorithms are not robust at tackling fine-grained relations.

Finally, the joint relation extractors' performance was analyzed with respect to fine-grained and coarse datasets. It can be inferred from Figure \ref{subfig:joint_fin} that algorithms UniRel and RIFRE performed better at tackling fine-grained relations than SPN4RE and TDEER. A probable reason for this finding stems from the use of prior relation label knowledge incorporated in the training process by the former algorithms. \textit{Thus, it can be inferred that the prior knowledge of the relation labels can help relation extractors when diverse relationships are present in a dataset.}

\superComplex
\fewComplex
\jreComplex

\subsubsection{Short vs. long sample lengths.}
The long length of the input samples is believed to be a complex use case for most relation extractors. However, based on the results in Figure \ref{subfig:sup_len}, it can be observed that Recurrent, PLM, and Prompt-PLM algorithms were sufficiently adept at handling this complex scenario with only a negligible decrease in performance from the short-input to long-input categories. Thus, it can be inferred that in the presence of sufficient data, the individual sample length is not a contributing factor to the performance variations experienced by neural relation classifiers.

On the contrary, it was found that in the few-shot learning setting, Prompt-LLM relation classifiers performed significantly better than PLM and Prompt-PLM algorithms at predicting relationships from long input samples. The performance boost of the LLMs can be attributed to their longer context windows (more than 16k tokens for GPT-3.5-turbo) and efficiency at tackling low-resource scenarios. Finally, the performance degradation experienced by PLM and prompt-PLM algorithms when handling long input compared to shorter samples leads to interesting insights. It can be inferred that, in general, such algorithms are inefficient at handling long input data. The performance observed for the supervised setting in the long input category could result from learning similar distributions from the training data rather than the contextual understanding of the input sample. Thus, without the strength of the training data, PLM and prompt-PLM algorithms fail to tackle longer input samples.

Finally, it was found that joint relation extractors performed comparably in the short and long-input categories, indicating that the length of the input sample is not a contributing factor to the performance degradation of these algorithms.

\subsubsection{Muliple relations vs. one-to-one relations.}
A fine-grained analysis was conducted to investigate the performance of the models by isolating different categories of multiple relations and overlapping entities. For the analysis of multiple relations, the data was divided into samples with only one relation ($n=1$), two relations ($n=2$), three relations ($n\geq3$), four relations ($n\geq4$), and five or more relations ($n\geq5$). Similarly, for overlapping entities, the split included samples with no overlap (\emph{norm}), entity pair overlap (\emph{epo}), and single entity overlap (\emph{seo}), respectively. Tables \ref{aptab:RC_multi}, \ref{aptab:few_multi} and \ref{aptab:JRE_multi} in the Appendix show the split statistics of the test data for datasets that suffer from this problem under supervised, few-shot and joint relation extraction domains.

First, the performance of the relation classification algorithms was investigated under the above-mentioned categories of multiple relations and overlapping entities. It can be observed from Figure \ref{fig:superOverlap} that recurrent architectures were most susceptible to performance degradation as the number of relations increased per sample. Both PLM and Prompt-PLM algorithms performed better in this scenario, with only slight degradation in the complex categories. However, all algorithms had reduced performance when handling samples in \emph{epo} and \emph{seo} categories. Thus, these findings highlight the susceptibility of the relation classification algorithms to instances where entities share multiple relations in a text sample. Multiple relations present a confusing use case as multiple correct predictions can be applicable, and the relation classification paradigm is inept at tackling this use case.

Next, in the few-shot setting, the performance of Prompt-LLM algorithms took a hit when multiple relations and overlapping entities were associated with the text samples in the 1-shot setting. This can be inferred from Figure \ref{fig:fewComplex}. However, this trend became less apparent as more data was introduced in the training/demonstration process. Similarly, the performance of the PLM and prompt-PLM algorithms remained consistent in the complex categories in the few-shot setting. A probable reason for this consistency can be attributed to the fewer training samples and demonstrations used in this setting. The presence of a sparse training space might force the algorithms to understand the semantics of the input while making inferences, leading them to make connections between entities and relationships. These findings highlight the potential of using relation classification algorithms in low-resource settings to tackle multiple relations and overlapping entities.
\superOverlap
\fewOverlap
\jreMulti

Finally, the complex characteristics of multiple relations and overlapping entities were investigated with respect to the performance of the joint relation extractors. It can be observed from Figure \ref{fig:jreMultiOver} that the performance of all extractors deteriorated as the number of relations increased in the given sample, with significant performance drops when more than 5 relations were present. Similarly, the performance of the relation extractors was significantly worse when tackling samples with \emph{epo} overlap. These two observations are interconnected as more relations would positively correlate with both entities sharing different relationships. The presence of multiple relations and overlapping entities should be tackled by the prediction of multiple triplets by the joint algorithms. However, the observations indicate that the algorithms were inefficient at doing so. A possible reason could be the discrepancy between the predicted and the ground truth triplets, which do not match under the `exact match' setting.

\subsubsection{Longtail vs. uniform distribution.}
Figure \ref{fig:longtail} presents a pictorial representation of the long-tailed distribution by mapping the total number of samples in each relational class for the datasets categorized as long-tail in Table \ref{tab:procon}. It was found that all datasets other than FewRel experienced mild to adverse degrees of long-tail distribution, as can be seen by the constant and sharp decline in the number of samples. This trend points to the non-uniformity in the data distribution, where only a few relation labels are heavily populated. Thus, this section details the findings from the analysis to investigate the performance of the supervised learning algorithms on different relation labels from the long-tail datasets. The few-shot learning models were not evaluated for this setting.
\longtail

To facilitate the experiments, normalized weights were calculated for each relation label using min-max normalization with respect to the relational class consisting of the maximum and minimum samples. Figures \ref{fig:nytlong}-\ref{fig:taclong} in the Appendix show the heat maps of the F1 scores achieved by the relation classification algorithms for the relations of some of the long-tail datasets. It can be inferred from the figures that all relation classification and joint relation extraction algorithms were inefficient at handling relations with low representation in the datasets. This trend is evident from the dark-hued top rows corresponding to high F1 scores and the light-hued lowest rows correlating to low F1 scores on the heat maps. The problem stems from the inability of the models to learn an accurate representation of classes that do not have enough samples during training. Thus, these findings re-instantiate the prevalence of the long-tail distribution problem in the domain of neural relation extraction.

\hfill \break
\noindent The findings presented above highlight the strengths and weaknesses of relation classification and joint relation extraction algorithms. It showcases the distinct behavior of relation extractors with different data characteristics. By doing so, the analysis points to the challenging data characteristics that complicate the extraction of relations from textual data. Key observations from the analysis are recorded below, along with a simplified summary of the findings presented in Table \ref{tab:result}.
\begin{itemize}
    \item It can be inferred that PLM and Prompt-based algorithms learn from the distribution of the training set and require significant data for fine-tuning. This trend was apparent from their superior performance in tackling fine-grained datasets and long input lengths in the supervised setting. However, the low performance of the PLM and Prompt-based algorithms when handling long input in the few-shot setting brings forth the drawback of these algorithms by indicating the tendency of the algorithms to learn from the training distribution rather than accurately understand long samples. 
    \item LLM-based algorithms were efficient at tackling coarse relationships and long input lengths due to their vast knowledge and larger context windows, respectively. 
    \item It was found that the issue of multiple relation/overlapping entities and long tail data distribution were still significant problems for most relation classification and joint extractors in the supervised setting. The findings highlighted the possibility of using few-shot algorithms to tackle these issues. 
\end{itemize}
The next section presents results from the error analysis that investigated the reasons for the mispredictions observed by these algorithms.
\RES

\subsection{Qualitative Error Analysis}

The findings presented in the previous section mention some of the issues that neural relation extractors face today. It brought forth critical issues like the difficulty in extracting certain relation types and robustness to complicated relation extraction scenarios that are yet to be answered in the literature. Next, this section presents a qualitative error analysis that investigates the root cause of the performance gap in this field. By doing so, the analysis facilitates the exploration of data points that are difficult to classify for neural relation extractors and the reasons behind these misclassifications.

All relation classification and joint relation extractors were employed in the supervised setting for this analysis. On the other hand, only the 1-shot experiments were used for few-shot learning, as the UnleashLLM algorithm was only compatible with 1-shot experiments. The test data and corresponding predictions from the models were combined. The RETACRED dataset was not used in this analysis, as it has the same text samples as TACRED. The final test samples were then segregated into simple and challenging classes according to the ease or difficulty of prediction. A voting-based scheme was devised to achieve this segregation. Therefore, the pipeline created by Alt et al. \cite{alt-etal-2020-tacred} based on erudite \cite{wu-etal-2019-errudite} was followed, and each data sample was scored based on the number of models correctly classifying it. To this end, the input samples were segregated as follows:
\begin{itemize}
    \item \textbf{Simple} category consists of all samples correctly classified by more than 75\% of the models.
    \item \textbf{Challenging} category consists of all samples incorrectly classified by at least 75\% of the models.
\end{itemize}

First, an exploratory data analysis was conducted to investigate the data characteristics that prevent the extraction of relationships. The influence of linguistic markers and readability constructs was investigated as a starting point. Various low-level features were extracted from the raw text samples. Some examples include 1) linguistic features such as the number of subjects, verbs, and objects in the text sample, 2) entity attributes such as part-of-speech (POS), named entity recognition (NER), and dependency trees, and 3) nine readability measures from the textstat\footnote{\url{https://pypi.org/project/textstat/}} library. Extracting relations from text samples with lower readability scores and distinct linguistic properties was expected to be a challenge for neural relation extractors. However, no correlations were found between the linguistic measures and the model predictions for simple and challenging samples. Thus, further investigations were carried out to find the root cause of the errors.

The performance analysis highlighted the superiority of the PLM-based models when used in a supervised setting for relation classification. However, under certain conditions, the PLMs' tendency to memorize the training distribution diffused their focus from the context of the input sample. Thus, this section investigated the context of the text samples to find possible reasons for this behavior. On manually inspecting the samples in the simple and challenging categories, it was found that most samples in the challenging class were difficult to understand, even for a human annotator. Two scenarios were found responsible for this ``ambiguity'': 1) when the context provides insufficient or contradictory information, thereby confusing the model, and 2) when the relation definition is too fine-grained, and the model confuses between multiple relation labels. Some examples of the two cases of ambiguity are presented in Table \ref{tab:amb}. The discussion presented below further corroborates these findings.
\AMB

\subsubsection{Investigating contextual ambiguity}
First, a context similarity measure was borrowed from the error analysis tool Azimuth \cite{gauthier-melancon-etal-2022-azimuth}. The aim was to investigate the influence of contradictory context on relation extraction. For this, sentence transformers \cite{reimers-2019-sentence-bert} that could encode up to 512 tokens were used to generate contextual embeddings of all training and test samples. Then, Facebook's Faiss library \cite{johnson2019billion} was used to find the closest neighbors of the test samples in the training set. The goal was to analyze the nearest neighbors of the test samples and investigate whether they belonged to the same relational class. It was assumed that the relationship associated with a text sample would have some correlation with its context. The presence of neighbors of other classes would mean that the test sample is more similar to samples depicting other relations. Thus, it can be inferred that such samples have ambiguous contextual information that might confuse the algorithms in predicting incorrect relationships.

Figure \ref{fig:contextsim} shows the percentage of samples with or without at least one nearest neighbor of the same class among the five closest neighbors for recurrent, PLM, and Prompt-based algorithms. It can be inferred that, for all three algorithm types, more than 50-60\% of the samples from the challenging class did not have a single nearest neighbor from the same relational class in the training set. Similarly, almost ~80\% samples from the simple class had at least one nearest neighbor of the same class in the training set. These observations suggest that the samples in the challenging class were more similar to samples from other relational classes. Thus, these results indicate a strong presence of ambiguous context in the misclassified text samples.
\contextsim

Next, the susceptibility of the few-shot learning algorithms to contextual ambiguity was investigated. A major hurdle when using the context measure discussed above for the few-shot experiments was the unstable nature of the training data used by the LLM-based algorithms. For example, the GPT-RE algorithm used a demonstration retrieval strategy to select the most relevant training samples to be used as demonstrations. Thus, the demonstrations for this algorithm differ from the randomly selected samples used by the other algorithms. To resolve this issue, the nearest neighbors of the test samples were found from the whole training dataset, as in the previous section. It was hypothesized that even though all the training samples were not used for training or as demonstrations, the presence of distinct nearest neighbors would still indicate that the context of a text sample is ambiguous and hints at the presence of relations different from the ground truth labels.

Figure \ref{fig:FEWcontextsim} shows the nearest neighbor statistics of the few-shot experiments. The figure shows that the issue of contextual ambiguity is not prevalent in the few-shot setting. This is apparent from the high percentage of samples with nearest neighbors of the same class in both challenging and simple categories. Thus, it can be inferred that in the absence of abundant training data, the algorithms are less likely to make incorrect predictions based on similar samples from other classes in the training data.
\FEWcontextsim

\subsubsection{Investigating relation ambiguity}
Another ambiguous scenario stems from the poor and ill-structured definitions of relation labels. These correlating relations have similar meanings and are easily confused by the models. Table \ref{tab:corr_rel} shows a few relations and their correlating counterparts that models had difficulty deciphering. It must be noted that each relation was confused with relations belonging to the same family with similar meanings. For example, all samples with the relation ``location/us\_state/capital'' were confused with ``location/location/contains''. The two relations are similar in meaning and are derived from the same root ``location" of the knowledge base used for their creation. Such relations provide multiple correct prediction possibilities and are challenging for relation classification algorithms.
\RELCORR

Similarly, relational ambiguity was investigated by analyzing the top incorrect predictions made by the algorithms in the few-shot setting. Tables \ref{aptab:fewrel_llm}, \ref{aptab:fewrel_plm}, and \ref{aptab:fewrel_prompt} show the relations where all samples were misclassified as the mispredicted relation. Contrary to the results shown in Table \ref{tab:corr_rel}, it can be observed that the mispredicted relation by PLM, Prompt, and LLM-based models do not belong to similar relation classes. For example, all samples with the relation ``location/br\_state/capital'' were misclassified as ``people/person/place\_lived'', which has a completely different semantic meaning. Thus, it can be inferred that relation ambiguity is not a detrimental factor for the missed predictions made by the few-shot algorithms. 

\subsubsection{Joint Relation Extraction}

The issues discussed above relate to ambiguity in the input samples and the relation labels. Such scenarios confuse the algorithms by presenting multiple options for the correct predictions. At this point, one might argue that joint relation extractors were created to handle such ambiguous cases, as they work on predicting multiple relation triplets for each input sample. Thus, these algorithms should be the obvious solution to this problem. Although this is true, on analyzing the predictions obtained from the joint extractors, it was found that only $5\%$ of the samples marked as challenging by the supervised algorithms were correctly classified by the joint extractors. The negligible improvement, coupled with far more samples in the challenging category, suggests the inefficiency of joint relation extractors. The poor performance can be attributed to two major issues. First, it was found that, in most cases, the relation extractor could produce coherent entities and relation triplets. However, a significant discrepancy between ground-truth labels and the extracted triplets led to an increase in the false negative predictions made by the models, affecting their overall performance. Second, a major source of disparity comes from the problem of missed predictions by the joint relation extractors. Figure \ref{fig:missedJoint} depicts the distribution of samples that each JRE algorithm could not predict across all datasets. It can be inferred that most algorithms could not produce valid triplets for $\sim20\%$ percent of the test data. Ultimately, this issue led to a significant increase in false negative classifications and severely affected the performance of these algorithms.
\missedJoint

\hfill \break
\noindent The analysis presented in this section highlights critical issues and data characteristics unfavorable for efficient relation extraction from textual data. It was found that many niche issues hinder the performance of neural relation extractors. Ambiguous context and correlating relations were found to be one of the major issues the relation classification algorithms faced. Additionally, joint relation extraction algorithms did not prove to be a viable solution to this problem due to the sensitivity of the algorithms to ground truth data and the proclivity to missed predictions. Thus, few-shot algorithms were investigated, and it was found that the algorithms did not face the same issues of contextual and relational ambiguity as their supervised counterparts. Thus, it can be inferred that the field could benefit from a comprehensive few-shot learning-based analysis to find the root cause of the performance gap for these algorithms.
\section{Challenges \& Future Directions}\label{sec:future}
Relation extraction is a significantly advanced domain with many SOTA algorithms employing advanced deep-learning architectures. However, from the analysis discussed in this paper, it became apparent that there are still opportunities to enhance and improve the extraction process further. Thus, this section presents some crucial challenges that, if addressed, can help create better relation extractors. 

\subsection{Ambiguity}
The foremost challenge to relation extraction is the aspect of contextual ambiguity in the input data. This issue was apparent from the large percentage of ambiguous samples in the challenging class for supervised relation classifiers. The presence of ambiguous samples with diverging contextual information confuses the models, as the possibility of more than one correct relation exists. Furthermore, it was observed that relations with similar/correlating relation definitions posed a significant challenge for supervised relation extractors. Much like the previous case, these complicated relations led to the possibility of more than one correct relation per sample. Thus, alleviating these ambiguous scenarios would greatly help to make relation extractors more efficient. 

On one hand, joint relation extraction presents a natural way of alleviating scenarios arising from ambiguous contexts (detailed discussion in the following section). 
On the other hand, a probable direction for making relation classifiers robust to this problem could be by employing some input enhancement techniques.  Specifically, in cases where the context does not provide sufficient information, data augmentation techniques such as paraphrasing \cite{sharma2022systematic} or neural text generation \cite{10.1145/3486622.3493957} could be employed to expand the context of the text sample and possibly add context and entity-relevant information that can make the relational meaning more apparent. Similarly, for tackling relational ambiguity, more research in the direction of Prompt-PLM methods, which incorporate the verbalizations of the label space in the input prompt, should be conducted. Solving this complex scenario can have many benefits, as it can improve neural networks at fine-grained language understanding. 

\subsection{Joint Relation Extraction}
The analysis elucidates the inefficiency of joint relation extractors in handling most relation extraction scenarios. First, it was observed that the performance of the joint models was extremely brittle, raising questions about their generalizability. Second, they were found to have low and comparable performance with the transformer-based supervised algorithms in handling the issue of multiple relations and overlapping entities. However, despite the shortcomings, joint relation extraction remains a popular paradigm due to its integrated entity and relation extraction pipeline. An efficient joint relation extractor can eliminate the need for a separate named entity recognizer and can also aid in creating better information extraction systems.

Existing research in this field focuses on testing joint algorithms on a few datasets with similar characteristics. This practice has led to a narrow view of these algorithms, where they perform exceptionally well on the said datasets. According to the findings presented in this paper, one of the significant issues with joint extractors proved to be the false negative problem. This problem stems from the insufficiency of ground-truth labels, so even semantically correct predictions are deemed misclassification. One way to solve this problem would be to create datasets with comprehensive entity and relation annotations. Since this can be a cumbersome task due to the manual efforts involved, using LLMs as relation triplet annotators could be explored as a future direction to create more comprehensive datasets \cite{pmlr-v225-goel23a}. Second, a more exhaustive solution would be to reformulate the evaluation strategies for joint extractors. A probable avenue could be the introduction of evaluation metrics that do not solely rely on the ground truth labels but evaluate qualitative aspects such as semantic and contextual accuracy of the predicted triplets \cite{jiang2024genres}. Finally, it was found that research on joint relation extraction using LLMs has been scarce in the literature. Thus, conducting exhaustive research in this direction is crucial to making joint relation extractors robust and generalizable.

\subsection{Long-tail Data Distribution}
The problem of long-tail relations that arise from datasets with imbalanced classes proved to be a significant disruptor for supervised relation extractors. This problem is a critical bottleneck in the relation extraction pipeline, as supervised and joint extractors cannot handle relation classes with negligible data samples. A probable solution to this problem could be to harness the generative nature of LLMs and use them as synthetic data generators to balance out the long tail of the distribution \cite{zhou2024pga}. If the creation of novel data is not feasible, then the potential of LLM-based relation extractors should be explored to tackle this complex use case. The low resource inference capabilities of such models afforded by pre-training on internet-scale datasets make them a natural solution to this problem. The alleviation of issues arising from the long-tail distribution can prove to be critical in domains where the availability of labeled data is scarce, such as biomedical and personality prediction.

\subsection{Fine-grained Relation Distribution}
The findings in this study highlight the inefficiency of the LLMs in extracting relations from fine-grained datasets with few-shot learning. Relation extraction on fine-grained datasets that involve numerous labels is a challenging task as the possibility of similar relation types increases as more relations are introduced. However, these datasets pose an even more challenging use case for LLMs. It is impractical to incorporate the verbalization of the full label space in the prompt due to the restricted token limit. Furthermore, it is difficult to add demonstrations of each relation type in the prompt as the number of relations increases. Thus, future work is needed to improve LLMs at tackling fine-grained relation distributions. A possible direction could be the development of better-prompting strategies that can provide concise knowledge of the label space. For example, KBs contain essential knowledge regarding numerous entity and relation types; incorporating such knowledge sources in place of label verbalizations and demonstrations might help provide a comprehensive image of the label space and aid the extraction process \cite{pan2024unifying}. Research in developing robust fine-grained relation extractors would significantly impact relationship extraction from complex domains such as finance and business.

\section{Conclusion}\label{sec:conclusion}
This paper presents a comprehensive performance and error analysis of some of the most prominent neural relation extractors. The presented analysis addresses the research gap in the literature by investigating data-centric attributes that adversely affect the process of extracting relationships from text. The analysis concludes that contextual and relational ambiguity are the most significant reasons for the misclassifications observed by supervised relation extractors. Also, it highlights the inefficiency of the joint extractors and brings forth various challenges faced by the algorithms. Finally, through the investigation of the most recent prompt and LLM-based algorithms, the study aimed to find possible solutions for the existing problems while also highlighting some challenges with the newer algorithms. Future research to tackle the challenges and opportunities listed in this paper would significantly help build better relation extractors. The next step would be to develop better relation extractors in low-resource scenarios that are robust to ambiguous use cases and fine-grained relationships.

\bibliographystyle{unsrt}
\bibliography{ref}

\appendix
\section{Additional Details}
\subsection{Datasets}

The statistics of the datasets used for the supervised relation classification and joint relation extraction experiments can be found in Table \ref{aptab:superData}. Similar statistics for the few-shot experiments are in Table \ref{aptab:fewData}.
\superDATA
\fewDATA

Tables \ref{aptab:RC_multi}, \ref{aptab:few_multi} and \ref{aptab:JRE_multi} depict the number of samples belonging to various categories of multiple relations and overlapping entities for supervised and few-shot relation classification and joint relation extraction.

\multiRC
\multiJRE
\overlapFEW

The statistics of the entity type tagging endeavor can be found in Table \ref{aptab:ent_stats}

\entSTATS

\subsection{Algorithms} \label{app:method}
The re-implementation details for algorithms where significant changes were made have been discussed below. The rest of the algorithms were implemented based on their original source code.

\begin{itemize}
    \item \textbf{PAWARE}: The algorithms replace entity mentions with entity type information for all train and test samples. However, since accurate entity type information was not present for all entities of the datasets used in this study, this replacement was bypassed, and original entity mentions were used.
    \item \textbf{GPTRE}: The original research uses the ``text-davinci-003'' model. However, due to the deprecation of this model at this time, the study uses the ``gpt-3.5-turbo'' model from OpenAI. Only the sentence similarity-based demonstration retrieval strategy has been implemented for this work. This model is termed the ``GPT-RE\_SimCSE*+Reasoning'' model in the original paper. Due to the large datasets used in this study, adding the verbalization of all labels in the prompt was impossible. Thus, prompts were used without the verbalizations. These prompts were extracted from the original GPTRE GitHub repository provided by the authors. Example prompts used can be found in Table \ref{aptab:gptre}.
    \item \textbf{UnleashLLM}: Since the algorithm verbalizes the full label space in the input prompt, the WebNLG dataset could not be incorporated with this algorithm. Table \ref{aptab:unleash} shows example prompts used for this study.
    \item \textbf{UniRel}: This algorithm uses a one-word verbalization strategy for the input label space. The authors provide the same for the NYT10 and WebNLG datasets. Custom verbalizations were created for the rest of the datasets. However, due to the complexity of the label space for TACRED and RETACRED datasets, it was not possible to accommodate these datasets with the one-word verbalization strategy. The verbalizations used for the rest of the datasets can be found in Table
\end{itemize}

\GPTRE
\UNLEASH

\section{Results}
\superPERF
\fewPERF
\jointPERF

\LONG
\FewRELllm
\FewRELPLM
\FewRELPrompt

\end{document}